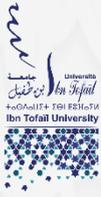
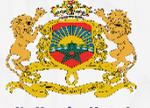

# NATIONAL SCHOOL OF APPLIED SCIENCES

DOCTORAL PROGRAM: SCIENCE AND ENGINEERING

THESIS

## AI-Driven Frameworks for Enhancing Data Quality in Big Data Ecosystems: Error Detection, Correction, and Metadata Integration

By

Widad Elouataoui

DECEMBER 2023

# Abstract

The extensive adoption of big data has inaugurated a new epoch of data-informed decision-making, revolutionizing various industries and domains. However, the effectiveness of these data-driven decisions is intrinsically tied to the quality of the underlying data, as poor data quality leads to inaccurate analyses and misleading conclusions. In addition, the massive volume, velocity, and variety of data sources pose significant challenges to data quality management, making the problem of big data quality more critical. This pressing concern has garnered recognition from both the scientific and practitioner communities, leading to a surge in research and practical solutions. Nonetheless, the problem of big data quality remains in its nascent stage, and current approaches often lack comprehensiveness and genericity. Indeed, there is limited coverage in addressing data quality dimensions as most systems tend to primarily focus on a limited set of metrics, overlooking the other data quality aspects.

Moreover, current approaches are more context-specific and lack a generic method, which limits their applicability to different domains. Furthermore, we noticed that there is a severe need for more intelligent and automated approaches that leverage the potential of artificial intelligence (AI) to perform advanced and sophisticated data quality corrections. To address these critical gaps in the field of big data quality, this Ph.D. thesis introduces an innovative set of interconnected, end-to-end frameworks designed to enhance big data quality by addressing the foundational axes of the field. For this, we first address the big data quality assessment by defining new quality metrics and introducing a novel concept of weighted data quality scoring to achieve greater precision in evaluating data quality. In a second axis, we address quality anomaly detection by proposing a generic and comprehensive framework that allows for detecting different types of quality anomalies based on an AI-driven model. In a third and final axis, we correct the detected quality anomalies using an innovative framework that predicts and replaces erroneous values with correct ones, based on a predictive model leading to an effective enhancement of big data quality. Furthermore, we go beyond the confines of data to address metadata by proposing an integration framework to enhance metadata quality within big data ecosystems. The proposed frameworks are implemented in different datasets and rigorously demonstrated to showcase high efficiency in improving big data quality. Finally, conclusions are made, and future work directions are highlighted.

# Acknowledgment

Alhamdulilah, all praises to Allah for giving me the ability, strength, and opportunity to undertake this study and complete this thesis.

First and foremost, I would like to express my deep gratitude to my supervisors, Pr. Youssef Gahi and Pr. Saida El Mendili for their exceptional supervision, invaluable advice, abundant kindness, and high availability throughout the entirety of this academic journey. Despite their multiple occupations, they consistently made themselves available to lend a sympathetic ear, offer motivation, and provide guidance. Special thanks go to Pr. Youssef Gahi, a conscientious advisor whose expertise and support were invaluable assets in making this dissertation possible and turning it into a pleasurable experience.

I would also like to thank Dr. Imane El Alaoui for her efforts, valuable insights, and support that have certainly contributed to advancing this academic research.

A special thanks goes out to my family members and friends for their moral and emotional boost. My warmest thanks go to my parents for their unconditional love, help, time, and support. Without you, I would not be what I have become today. I love you.

I am incredibly thankful to Ibn Tofail University for providing me with a beautiful working environment that facilitated my research. I am also profoundly grateful to all the dedicated professors whose expertise and commitment to teaching have contributed significantly to my knowledge and academic growth.

I would also like to thank the jury members for accepting to judge this thesis.

Finally, I want to convey my gratitude to all those who have directly or indirectly supported me in this noble field of scientific research.

# Table of Contents

# List of Tables

# List of Figures

# Chapter 1: Introduction

## 1. Background

In recent years, the advent of big data has revolutionized the way organizations collect, process, and analyze information. The exponential growth of digital data from various sources, such as social media, IoT devices, sensors, and transactional systems, has led to an unprecedented abundance of data. As a result, big data has become a driving force behind critical decision-making processes, strategic planning, and business innovations across diverse industries. The wide use of big data stems from its immense potential to unlock valuable insights and patterns that were previously hidden in the vast sea of information. Thus, thanks to big data, organizations can gain deeper insights into customer behaviors, market trends, operational efficiency, and predictive analytics. These insights empower companies to make data-driven decisions, enhance customer experiences, optimize processes, and identify new business opportunities. Despite the tremendous benefits of big data, this technological phenomenon also presents a distinctive set of challenges, with data quality being a prominent concern. Indeed, the particular nature of big data, characterized by its enormous volume, diverse variety, high velocity, and varying veracity, creates inherent complexities that can significantly affect the quality of the collected and processed data. Indeed, the vast amount of information results in errors, duplicate entries, and incomplete data, while the diverse sources and formats lead to inconsistencies and inaccuracies.

Additionally, including unstructured data in big data further complicates data quality management, as these data types lack predefined structures and necessitate specialized processing techniques. Moreover, the dynamic nature of big data raises doubts about its timeliness and reliability, which can potentially impact the accuracy of the decision-making process and result in disastrous outcomes. With all these facts, data quality is the driving force that unleashes the full potential of data and requires dedicated focus to address the particular challenges it presents effectively. Thus, high-quality data is the foundation for effective data analytics and informed decisions. With healthy data, companies can have reliable insight into their business, make solid strategic plans, better understand their customers, and even achieve financial gains.

On the other hand, poor data quality can lead to inaccurate analyses, misleading conclusions, and erroneous business strategies. Moreover, the potential risks of making decisions based on flawed data can have serious consequences, such as bad decisions, financial losses, compliance, and legal issues. Thus, it becomes evident that data quality is not just a technical concern. It is the master key that unlocks the actual value of big data.

## 2. Motivations

Exploring the era of big data has heightened our understanding of its immense significance, leading us to consider it the new gold rush of organizations worldwide. Nevertheless, despite the evolution of big data analytics technologies and tools, the critical matter of data quality remains an unresolved challenge. Indeed, aware of the axiom "Garbage in, Garbage out," we believe that poor-quality data can only lead to biased outcomes regardless of the advancements and capabilities of the employed tools. This big significance of data quality motivates us to delve further into the big data quality era to explore the underlying obstacles making it an ongoing concern and make significant contributions. Although much progress has been made in exploiting big data's potential, considerable gaps exist in effectively addressing data quality concerns comprehensively and in a sophisticated manner. While data quality assessment, detection, and correction are crucial components of the data quality chain, current practices often fall short in considering all aspects of data quality with the depth and precision required. The traditional data quality approaches primarily focus on a limited set of metrics, overlooking the other quality aspects that collectively contribute to data quality. As a result, data quality evaluation may lack the necessary granularity and context to accurately identify potential anomalies and inconsistencies.

Moreover, in the context of anomaly detection, existing approaches tend to focus solely on identifying outliers within the data and do not consider the other aspects of data quality. These data quality anomalies can be complex to detect because they don't show the usual patterns of abnormal behavior. Therefore, there is a strong need for more advanced and practical approaches to find and fix data quality anomalies, especially when dealing with Big Data.

Furthermore, the existing methodologies for anomaly correction in big data environments are still at a nascent stage. Indeed, existing research studies tackling data quality anomalies have been context-specific, lacking a general approach. As a result, their applicability to different situations and industries is limited.

Moreover, these studies often only addressed one specific aspect, like outliers or duplications, failing to capture all potential data quality anomalies that may affect the reliability and accuracy of big data. Thus, in the realm of big data quality, we have identified three main weaknesses: Firstly, there is a lack of comprehensiveness, as only a few data quality dimensions are considered. Secondly, the generality of approaches is limited as they are often context-specific. Lastly, there is a severe need for more sophisticated and automated methodologies that leverage the power of machine learning to perform advanced and effective data quality corrections. This thesis seeks to propose novel frameworks and methods to bridge these critical gaps and enhance the big data quality assessment, anomaly detection, and anomaly correction capabilities in big data. By incorporating comprehensive data quality metrics into the assessment process and exploring innovative approaches to quality anomaly detection and correction, we aim to advance big data analytics and allow organizations to make more informed decisions based on high-quality data.

## 3. Objectives

This thesis aims to thoroughly address the quality of Big Data through three main axes of data quality management:

1. **Data Quality Assessment:** This involves measuring and evaluating the different quality aspects of the dataset to gain insights into its level of quality. This allows us to quantify the extent to which the dataset is good or bad regarding quality assessment.
2. **Data Quality Anomaly Detection:** This consists of detecting the quality anomalies that exist within the dataset, which degrade the level of quality. This involves identifying data points or patterns that deviate significantly from the expected behavior.
3. **Data Quality Anomaly Correction:** After detecting quality anomalies, the necessary corrections are made to address the identified issues. This phase involves refining and improving the data by rectifying errors, resolving inconsistencies, and enhancing the overall quality of the dataset.

For each axe, we aim to address the three main weaknesses mentioned in the previous section: Limited coverage of data quality dimensions, High context dependency, and the lack of automated and machine learning-driven approaches. Thus, by effectively managing data quality through these three key axes, big data quality is improved to a better level, allowing more reliable informed decision-making.

## 4. Contributions

This thesis aims to address the three aforementioned key axes of data quality management by making the following contributions:

1. We define a comprehensive Big Data Quality Assessment Framework based on 12 quality metrics: Timeliness, Completeness, Volatility, Conformity, Uniqueness, Consistency, Relevancy, Ease of manipulation, Security, Readability, Integrity and Accessibility. Moreover, we introduce the weighted data quality assessment concept to enhance the performed measures' accuracy.
2. We define a novel Data Quality Anomaly Detection Framework based on a machine learning model that allows identifying potential generic data quality anomalies for Big Data related to six quality dimensions: Accuracy, Completeness, Consistency, Conformity, Readability, and Uniqueness.
3. As a foundational step for detecting quality anomalies, we suggest an End-to-End Big Data Entity Resolution Framework, surpassing current techniques and delivering optimal outcomes by applying a semi-supervised learning methodology. This proposed framework employs a real-time learning technique, effectively handling declining model performance while preserving high accuracy in real-time operations.
4. We define and assess a novel parameter called the "Anomaly Quality Score." This parameter quantifies the extent of deviation from the norm and the low quality of the identified anomalies across all quality dimensions and within the whole dataset.
5. We propose an advanced framework for Big Data Quality Anomaly Correction that adopts an intelligent and sophisticated methodology based on a predictive model, allowing the correct anomalies related to six critical dimensions of data quality: Accuracy, Completeness, Conformity, Uniqueness, Consistency, and Readability. The proposed framework is not restricted to a specific field and offers a generic approach to correcting big data quality anomalies.

## 5. Thesis Outline

The remainder of this thesis is organized as follows. In Chapter 2, we highlight the immense importance of data quality for big data by exploring the primary information about the big data era and data quality. We survey existing approaches that have tackled data quality in quality assessment, anomaly detection, and anomaly correction. In Chapter 3, we present our proposed big data quality assessment framework. Chapter 4 describes the new big data quality anomaly detection framework after presenting our submitted Deduplication Framework based on Online Continuous Learning. Chapter 5 offers a new big data quality anomaly correction framework that improves data quality by correcting the identified anomalies. Finally, in Chapter 6, we conclude our thesis report and provide future research directions.

## 6. List of publications

**Journals:**

1. W. Elouataoui, E. Saida and Y. Gahi, "Quality Anomaly Detection Using Predictive Techniques: An Extensive Big Data Quality Framework for Reliable Data Analysis," in *IEEE Access*, vol. 11, pp. 103306-103318, 2023, doi: 10.1109/ACCESS.2023.3317354.
2. W. Elouataoui, I. E. Alaoui, S. E. Mendili, and Y. Gahi, 'An End-to-End Big Data Deduplication Framework based on Online Continuous Learning,' International Journal of Advanced Computer Science and Applications (IJACSA), vol. 13, no. 9, Art. no. 9, Dec. 2022, doi: 10.14569/IJACSA.2022.0130933.
3. W. Elouataoui, I. El Alaoui, S. El Mendili, and Y. Gahi, 'An Advanced Big Data Quality Framework Based on Weighted Metrics,' Big Data and Cognitive Computing, vol. 6, no. 4, Art. no. 4, Dec. 2022, doi: 10.3390/bdcc6040153.
4. W. Elouataoui, E. Saida and Y. Gahi, " An Automated Big Data Quality Anomaly Correction Framework using Predictive Analysis" (To Appear).

**Chapters:**

5. W. Elouataoui, I.E. Alaoui, and Y. Gahi, "Data Quality in the Era of Big Data: A Global Review," in Y. Baddi, Y. Gahi, Y. Maleh, M. Alazab, and L. Tawalbeh (Eds.), Big Data Intelligence for Smart Applications, vol. 994, Studies in Computational Intelligence, Springer, Cham, 2022, pp. 1. doi: https://doi.org/10.1007/978-3-030-87954-9_1

**International Conferences:**

6. W. Elouataoui, I. El Alaoui, and Y. Gahi, "Metadata Quality Dimensions for Big Data Use Cases," in Proceedings of the 2nd International Conference on Big Data, Modelling and Machine Learning - BML, ISBN 978-989-758-559-3, SciTePress, pp. 488-495, 2022. doi: 10.5220/0010737400003101

7. W. Elouataoui, I. El Alaoui, and Y. Gahi, "Metadata Quality in the Era of Big Data and Unstructured Content," in Advances in Information, Communication and Cybersecurity, ICI2C 2021, vol. 357, Lecture Notes in Networks and Systems, Springer, Cham, 2022, pp. 145-157. doi: https://doi.org/10.1007/978-3-030-91738-8_11

8. W. Elouataoui, E. Saida, and Y. Gahi, "Online Continual Learning Data Quality Framework for Entity Resolution," in the 10th International Conference on Future Internet of Things and Cloud (FICloud), ISBN 979-8-3503-1635-3/23/$31.00 IEEE, doi: 10.1109/FiCloud58648.2023.00039.

9. W. Elouataoui, E. Saida and Y. Gahi, "Big Data Quality Anomaly Scoring Framework Using Artificial Intelligence " in Proceedings of the 7th IEEE 2023 Congress on Information Science and Technology. (To Appear)

# Chapter 2: The Big Importance of Data Quality for Big Data

## 1. Introduction

The digital transformation of companies has led to data becoming the cornerstone for organizations to differentiate themselves and gain a sustainable competitive advantage. However, effectively leveraging this data-driven strategy requires robust data collection, storage, and large-scale processing infrastructure. As the number of data-generating devices grows, the volume of data produced also rises exponentially. According to a study by IDC, the amount of data generated is expected to reach 175 zettabytes (175 billion terabytes) by 2025 [1]. Consequently, the emergence of Big Data has provided a new data ecosystem capable of supporting, processing, and managing this explosive growth of information. In the following sections, we will delve into the Big Data Era, beginning with defining Big Data and its key characteristics, then exploring the big data value chain and the specific tools and methodologies employed in this era. Subsequently, we will delve into the realm of data quality, examining its dimensions, the challenges posed by data quality in the big data era, and elucidating the immense importance of data quality for Big Data.

## 2. Big Data Era

### 2.1. What is Big Data?

Big Data involves collecting, storing, processing, examining, and extracting knowledge from a rapidly expanding and diverse dataset [2]. Indeed, big data comprises data from multiple sources, representing a fusion of structured and unstructured data that encompasses a variety of data formats, typically characterized by noise and imprecision. Consequently, within the realm of big data, the data acquired is often of low quality, rendering it unsuitable for direct utilization and, thus, needs to be cleaned and transformed into useful information. It is essential to place Big Data into its historical perspective to understand Big Data best. While the concept of big data is relatively recent, the roots of large datasets can be traced back to the 1960s and '70s, when the data landscape was in its early stages with the emergence of data centers and the development of relational database systems. During this initial phase, data management heavily relied on techniques for storing, extracting, and optimizing structured data in Relational Database Management Systems (RDBMS). In the early 2000s, the proliferation of the Internet and web applications brought about an exponential increase in data generation. New unstructured data from IP-specific search and interaction logs have appeared alongside the structured data stored in relational databases. These unstructured data sources, including click rates, IP-specific location data, and search logs, provided organizations with insights into the behaviors and needs of internet users.

Consequently, organizations faced the challenge of finding new approaches and storage solutions to analyze these diverse data types effectively. Social media data's advent and subsequent expansion further intensified the demand for tools, technologies, and analytical techniques capable of extracting valuable information from unstructured data. During this stage, the term "Big Data" began to gain prominence. Later, a new phase in the evolution of Big Data started, driven by the wide adoption of mobile technology and devices, resulting in an enormous amount of data generated from these sources. In 2011, the number of mobile devices and tablets surpassed that of laptops and PCs for the first time, leading to new data analysis requirements, such as real-time processing and location-aware analysis. Later, the proliferation of Internet of Things (IoT) devices further amplified the generation of big data, as these interconnected devices continuously produced vast streams of data from various sources, including sensors, wearables, and connected gadgets. In 2022, the worldwide market for big data and business analytics reached US$294.16 billion, with projections indicating growth to US$662.63 billion by 2028. A timeline outlining the evolution of Big Data is illustrated in Figure 1.

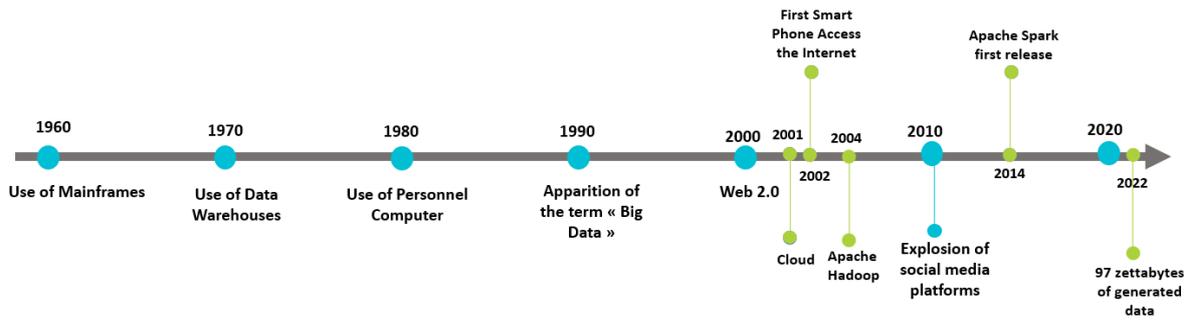

*Figure 1: Big Data Evolution Timeline*

## 2.2. Big Data Characteristics

Admittedly, the expression 'big data' is associated with an ever-growing and substantial volume of data, yet it's essential to understand that big data extends beyond data volume. Big Data is characterized by numerous characteristics called Big Data V's. These characteristics were initially formulated by Doug Laney in 2001, defining big data with three V's: Volume, Velocity, and Variety. Subsequently, SAS (Statistical Analysis System) introduced two additional features: Variability and complexity. Oracle later extended the definition to encompass four V's: Volume, Velocity, Variety, and Value. In 2014, Kirk Born from Data Science Central expanded the list to 10 V's, which included Volume, Variety, Velocity, Veracity, Validity, Value, Variability, Venue, Vocabulary, and Vagueness [3]. Over time, the literature proposed more V's, like Vulnerability, Virality, and Verbosity. The following table presents the most commonly encountered Big Data V's.

*Table 1: Big Data Characteristics*

| Characteristic | Elucidation | Meaning |
|---|---|---|
| **Volume** | What is the size of the data? | A significant feature of big data. It refers to the vast amount of generated data measured in Exabytes, Zettabytes, or even Yottabytes. |
| **Variety** | How heterogeneous are the data? | Big data encompasses a wide range of data sources (e.g., social media, IoT, mobile devices) that come in different formats, such as |
| | | geolocation data, sensor data, and social media data. |
| **Velocity** | How fast is data generated? | It refers to the rate of data generation: It denotes the speed at which data is being produced, analyzed, and processed. |
| **Veracity** | How much could data be trusted? | This characteristic emphasizes the accuracy and truthfulness of the generated results. Ensuring trustworthy data is crucial for making informed business decisions. |
| **Value** | Is the generated data valuable? | It refers to the potential business benefits that can be derived from processing the data. |
| **Variability** | What is the rate of change of data meaning? | This aspect sheds light on the dynamic nature of data, indicating how often the meaning of the data changes over time. |
| **Visualization** | How is it challenging to develop a meaningful visualization of data? | It refers to presenting extracted information in a clear and understandable format. |
| **Validity** | Is data suitable and accurate for the intended use? | It pertains to the correctness of the input data and its suitability for its intended purpose. |
| **Volatility** | How long could data be kept and considered valuable? | It reflects the duration of data's usefulness and relevance before it becomes outdated. |
| **Viability** | How relevant is the data? | This process consists of "uncovering the latent, hidden relationships among these variables." to "confirm a particular variable's relevance before investing in creating a fully-featured model." [4] |

| Vocabulary | What terminology is used to describe data? | It refers to the specific language and terms used to characterize the content and structure of the data. |
|---|---|---|
| Vagueness | How much data is available? | This pertains to the ambiguity surrounding big data, including its nature, content, availability of tools, and platform selection. |
| Venue | What kind of platforms and systems are used to manage data? | It concerns the different types of platforms used to handle data, including personnel systems and private & public clouds. |
| Viscosity | How difficult is it to work with data? | It refers to the difficulty of using or integrating the data efficiently. |

The unique characteristics of big data, including its immense Volume, diverse Variety, high Velocity, and the critical aspect of Veracity, present significant challenges in processing and handling such vast amounts of information. Additionally, the ever-changing Variability of data meaning and the difficulty in developing meaningful Visualizations add complexity to the data management process. Moreover, data volatility requires constant evaluation of its relevance over time. Therefore, ensuring the Validity of such data is of utmost importance to extract reliable insights for informed decision-making. To overcome these challenges and unlock the full potential of big data, data goes through a transformative process known as the Big Data Value Chain, transforming raw data into valuable insights. In the next section, we will delve into the Big Data Value Chain, exploring each step and understanding its significance in deriving meaningful insights from the vast world of big data.

## 2.3. Big Data Value Chain

The data value chain has emerged as a response to the digital transformation, as a framework that outlines data flow within data-centric enterprises and facilitates the extraction of data's inherent value. However, when faced with the escalating challenges of big data, the traditional data value chain has shown significant limitations, necessitating the introduction of the big data value chain (BDVC). The BDVC serves as an extension of the data value chain, specifically designed to accommodate the unique characteristics of big data and enable the transformation of raw data into valuable insights. Various models of the BDVC have been proposed in academic literature [5] [6] [7]. This section presents the most comprehensive and up-to-date BDVC, as documented in [8], consisting of seven stages depicted in Figure 2.

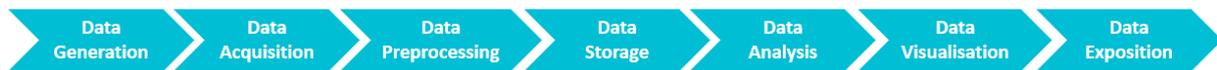

Figure 2 : Big Data Value Chain (BDVC)

1- **Data Generation:** The initial phase involves data generation, where information is produced. The data generated can take various forms, such as structured, semi-structured, or unstructured, depending on the source type (human-generated, process-mediated, or machine-generated) [16]. Following data generation, the subsequent step involves collecting and recording data via an acquisition process.

2- **Data Acquisition:** This stage involves aggregating data from diverse origins and preserving it in its original form before any alterations. Given that the collected data may originate from multiple sources and be of various types, it is essential to identify the integrity and diversity of these sources, as each data type necessitates distinct processing approaches.

3- **Data Pre-processing:** The preprocessing phase within BDVC is of utmost importance, as it serves the purpose of refining the data collected in the previous stage by eliminating inconsistencies and inaccurate values, thereby transforming the data into a valuable and practical format. One of the most prevalent challenges during this phase is dealing with noisy and poor-quality data. This phase significantly enhances data quality through its various subphases, including data transformation, data integration, data cleaning, and data reduction [29]:

- *Data transformation* involves converting the data's structure into a more suitable format to increase its value.
- *Data integration:* The process of combining diverse data types from multiple sources into a cohesive dataset that is easily accessible.
- *Data cleaning:* Ensuring accuracy and eliminating anomalies by eradicating corrupted and duplicated values.

- *Data reduction:* This process aims to represent the original data in the most simplified form possible to reduce storage requirements. It encompasses various techniques, such as data noise reduction, data compression, and other dimensionality reduction methods [29].

4- **Data Storage:** Following the preprocessing phase, the data is stored to preserve a clean and refined version for future utilization. Data storage is critical in the Big Data Value Chain (BDVC), as it houses vast amounts of data collected from various sources. The data storage phase is crucial in ensuring the data's availability, accessibility, and reliability for subsequent analysis and decision-making processes.

5- **Data Analysis:** The data analysis stage represents a pivotal phase, examining and manipulating previously cleaned data to unveil unknown correlations and patterns, ultimately converting data into meaningful knowledge. Various big data analysis techniques, such as machine learning, deep learning, and data mining, are employed during this phase.

6- **Data Visualization:** During this phase, the analysis outcomes are translated into a visually accessible format, utilizing graphical elements like graphs, maps, and dashboards. This approach enables managers to explore data more efficiently and provides valuable support for decision-making.

7- **Data Exposition:** The Big Data Value Chain's final step involves disseminating and exposing the valuable insights derived from each phase. These extracted insights can be utilized to offer specialized services, shared as open data with the public, or used internally to enhance organizational performance.

The data value chain has played a pivotal role in unlocking the potential of data-centric enterprises by guiding the data flow and extracting its intrinsic value. However, the effectiveness of the big data value chain heavily relies on implementing the appropriate tools and methodologies. Indeed, the emergence of big data has brought new challenges and requirements related to data storage and processing techniques, as traditional tools have shown serious weaknesses in handling big data. In the next section, we will review some of the most common tools and methodologies specifically designed to address the unique requirements of big data, exploring their capabilities to store and analyze the vast amount of raw data.

## 2.4. Big Data Tools

To harness the potential of Big Data and derive valuable insights, specialized techniques are employed to store, process, and analyze these massive datasets efficiently. Indeed, in the realm of Big Data, data's vast and complex nature requires specific methodologies and tools adapted to its previously presented characteristics. Thus, because of Big Data V's, traditional data tools, primarily designed to handle smaller and d data, are unsuitable for effectively managing and analyzing Big Data. This section will explore the methodologies and tools used in Big Data analysis. These methodologies and tools encompass various data processing, storage, and analysis techniques and provide the necessary infrastructure to handle the challenges related to Big Data. While many technologies and methodologies are available for Big Data analysis, we present the most common and widely used Big Data tools, as illustrated in Figure 3.

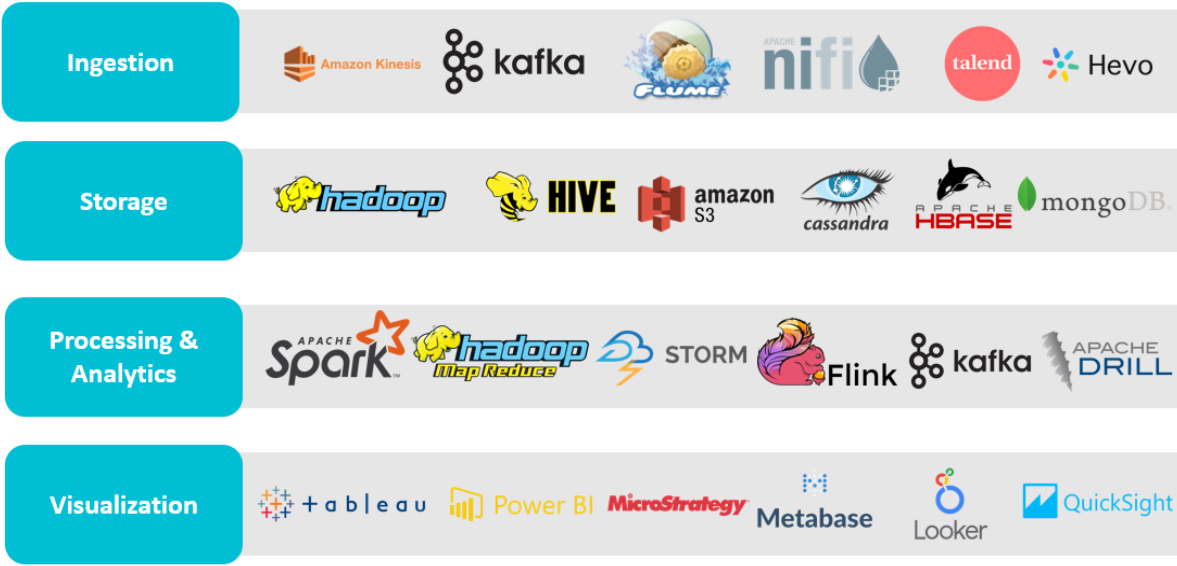

*Figure 3 : Big Data Tools*

### 2.4.1. Ingestion:

**Apache Nifi:** Apache Nifi is an open-source data integration tool that facilitates the automation of data flow between various systems, enabling seamless data movement, transformation, and processing.

**Apache Flume:** Apache Flume is a distributed, reliable, and available system designed for efficiently collecting, aggregating, and moving large amounts of streaming data from various sources to a centralized data store.

**Apache Kafka:** Apache Kafka is a distributed event streaming platform that enables the ingestion, storage, and processing of high volumes of real-time data streams, making it suitable for building real-time data pipelines and applications.

**Amazon Kinesis:** Amazon Kinesis is a managed streaming platform provided by Amazon Web Services (AWS) that allows users to collect, process, and analyze real-time streaming data at scale, enabling applications to respond to data in real time.

**Talend:** Talend is a widespread open-source data integration and ETL (Extract, Transform, Load) tool that helps organizations with data integration, data transformation, and data governance tasks, providing a comprehensive platform for managing data workflows

**Hevo:** Hevo is a big data ingestion platform that automates collecting, transforming, and loading data from various sources into data warehouses and other destinations, simplifying and accelerating the data integration process.

### 2.4.2. Storage:

**Hadoop-HDFS**: Hadoop Distributed File System (HDFS) is a core component of the Hadoop ecosystem. It is a distributed, fault-tolerant file system designed to store and manage vast amounts of data across multiple nodes in a Hadoop cluster.

**Hive:** Hive is a big data storage tool and data warehouse infrastructure built on top of Apache Hadoop. It provides a high-level, declarative query language called HiveQL (similar to SQL) that allows users to query and analyze large datasets stored in distributed storage systems like Hadoop HDFS.

**Cloud Storage Platforms:** Among the popular cloud storage platforms for big data, we can cite Amazon S3 (Simple Storage Service), Microsoft Azure Blob Storage, Google Cloud Storage, IBM Cloud Object Storage, etc.

**Cassandra:** Cassandra is a distributed NoSQL database designed for handling large amounts of data across multiple commodity servers. It offers high availability, fault tolerance, and linear scalability, making it suitable for managing and querying large-scale, decentralized data.

**HBase:** HBase is an open-source, distributed, and scalable NoSQL database that runs on top of the Hadoop Distributed File System (HDFS). It is optimized for real-time read and write operations on large datasets and is commonly used for applications requiring low-latency access to big data.

**MongoDB:** MongoDB is a popular open-source NoSQL database that uses a document-oriented model to store data in a flexible, schema-less format. It is designed for scalability, high performance, and ease of development, making it suitable for various applications and workloads.

### 2.4.3.    Processing & Analytics:

**Apache Drill:** Distributed SQL query engine designed to perform interactive queries using SQL syntax on various big data sources.

**Flink:** Stream processing framework offering batch and real-time data processing capabilities, providing high throughput, event-time processing, and state management.

**Storm:** Real-time stream processing system focused on ingesting, processing, and distributing streaming data, often used for real-time analytics and event processing.

**Spark:** Apache Spark is a fast and distributed data processing engine. It provides in-memory data processing capabilities, making it efficient for iterative algorithms and real-time data processing tasks.

**Hadoop-MapReduce:** Hadoop MapReduce is a programming model and processing framework for distributed data processing in Hadoop. It divides tasks into map and reduce functions for parallel processing.

**Apache Kafka:** Apache Kafka is a distributed streaming platform that enables real-time data streaming and processing. It is used for handling high-throughput, fault-tolerant, and scalable data streams.

### 2.4.4. Visualization:

**Tableau:** Tableau is a powerful data visualization tool that allows users to create interactive and shareable dashboards, charts, and reports from various data sources, helping to uncover insights and patterns in data.

**Power BI:** Power BI is a business analytics tool by Microsoft that enables users to visualize data, create interactive reports, and share insights. It integrates with various Microsoft products and other data sources.

**MicroStrategy:** MicroStrategy is a business intelligence platform that offers data visualization, self-service analytics, and mobile apps for data exploration, enabling organizations to make data-driven decisions.

**Metabase:** Metabase is an open-source business intelligence tool that simplifying data visualization and exploration. It allows users to create dashboards, query databases, and generate insights from their data.

**Amazon QuickSight:** Amazon QuickSight is an AWS cloud-powered business intelligence service enabling users to create interactive dashboards and reports from various sources, facilitating data exploration and analysis.

**Looker:** Looker is a data exploration and business intelligence platform that allows organizations to create and share insights through interactive dashboards and reports, connecting to multiple data sources for analysis.

The evolution of big data storage platforms and methodologies has significantly enhanced our ability to efficiently store and process massive volumes of information. Cloud storage platforms and distributed storage systems, such as Hadoop HDFS and NoSQL databases, have revolutionized how we handle big data. These tools have undoubtedly enabled efficient storage and processing of big data; however, it does not guarantee the inherent quality of the managed data. Indeed, big data often encompasses data from diverse sources, in various formats, and at high velocities, which can lead to issues related to different aspects of data quality. Data quality is essential to derive meaningful insights and make well-informed decisions from big data. For this purpose, in the next section, we will delve into the critical aspect of data quality in the context of big data. We will explore the dimensions of data quality, the challenges of ensuring high-quality data, and why data quality is paramount in the big data era.

## 3. Data Quality for Big Data

### 3.1. Data Quality Dimensions

Before tackling the challenges surrounding data quality in big data, gaining a comprehensive understanding of what data quality entails is essential. Data quality refers to the degree to which data meets the requirements and expectations of its intended use. Data quality could be defined in terms of its various aspects, such as data accuracy, data completeness, or data consistency, also known as Data quality dimensions (DQDs). Thus, data quality dimensions are "a set of data quality attributes that represent a single aspect or construct of data quality" [9]. Data quality becomes increasingly critical in big data, where large volumes of diverse and dynamic information must be managed and processed. Undoubtedly, the concept of data quality predates the advent of big data and initially found application in the context of conventional data stored within relational databases.

Nevertheless, big data has introduced new challenges that conventional data quality methodologies cannot address. This shift from data quality to big data quality is intimately tied to including diverse data formats, data origins, and application realms within the big data domain [10]. With the rapid evolution and growing need for big data, the concept of data quality has expanded significantly, leading to numerous data quality dimensions. Over time, researchers and practitioners have identified various aspects that contribute to the overall quality of data in the context of big data. While the traditional data quality dimensions remain crucial, such as accuracy, completeness, and timeliness, new dimensions have been introduced to address the unique challenges presented by big data. Today, more than 50 data quality dimensions are suggested in the literature, reflecting the diversity and complexity of data sources, formats, and use cases in the big data landscape. Table 2 defines the most common and essential data quality dimensions encompassing the critical aspects of data quality.

*Table 2 : Data Quality Dimensions*

| Dimension | Meaning |
|---|---|
| **Accuracy** | The degree to which data is reliable and reflects real-world values |
| **Completeness** | The extent to which necessary data is present and no information is missing. |
| **Uniqueness** | The absence of duplicate records referring to the same world entity within a dataset. |
| **Consistency** | The coherence and uniformity of data across different sources or within the same dataset. |
| **Timeliness** | Refers to how recent and up-to-date the data is. |
| **Volatility** | Refers to the rate of change or the frequency at which data values or attributes are updated, modified, or replaced over time |
| **Conformity** | Refers to the alignment of data with predefined rules, standards, and data types |
| **Validity** | The extent to which data adhere to its intended use or purpose. |
| **Readability** | The extent to which data is presented clearly and understandably. |
| **Integrity** | Refers to the accuracy and trustworthiness of data over its lifecycle and ensures that data values have not been altered. |
| **Relevancy** | Assesses the pertinence and significance of data to its specific business context or use case. |
| **Accessibility** | The ease with which data can be accessed and retrieved when needed ensures that data is available. |
| **Security** | It focuses on safeguarding data against unauthorized access, ensuring data privacy, confidentiality, and protection from potential threats. |
| **Ease of Manipulation** | The degree to which data can be manipulated and processed efficiently, ensuring that data is amenable to various data operations and analyses. |

| **Understandability** | The clarity and simplicity with which data can be interpreted and understood by users ensures that data is presented in a manner that facilitates easy comprehension. |
|---|---|
| **Data specification** | The data quality dimension involves using metadata and data documentation to specify data's attributes, structure, and semantics, ensuring that data is well-described and documented. |
| **Granularity** | The level of detail or granularity at which data is recorded and stored ensures that data is captured appropriately for its intended use. |
| **Interoperability** | The data quality dimension assesses the ability of data to be exchanged and used between different systems and applications, ensuring that data can be seamlessly integrated and shared. |
| **Concise** | Refers to the brevity and conciseness of data, ensuring that data is represented compactly and straightforwardly. |
| **Interpretability** | The ease with which data can be interpreted and understood in the context of its analysis and use ensures that data is transparent and interpretable to derive meaningful insights. |

Even with numerous data quality dimensions available, their relevance and importance may vary depending on the specific intended use of the data. Thus, different industries and use cases may prioritize exact dimensions over others based on the nature of their data and the criticality of the quality dimension for their operations. It is also essential to recognize that data quality dimensions are often closely interrelated and dependent on each other. Therefore, simply focusing on one dimension without considering its correlation with different dimensions may not be a practical approach to support overall data quality. Thus, the effectiveness of data quality efforts relies on a holistic and comprehensive understanding of how different dimensions interact and influence one another. Having explored the concept of data quality and its main dimensions, we can now delve into the challenges of big data quality and how big data characteristics can influence the dimensions of data quality.

## 3.2. Big Data Quality Challenges

Big data characteristics cannot be left behind when discussing big data quality. Indeed, enhancing data quality for big data is a challenging thesis because of the particular characteristics of big data. The following table highlights the data quality challenges raised by each aspect of big data.

*Table 3 : Big Data Quality Challenges*

| Characteristic | Impacts on Data Quality |
|---|---|
| Volume | The sheer volume of data presents challenges in efficiently processing and managing vast amounts of noisy and inconsistent data. |
| Variety | Data diversity results in a mix of data types, inconsistent formats, and incompatible structures and meanings, adding complexity to data management. |
| Velocity | High-velocity data demands high-performance processing tools and can create timing challenges, especially for time-sensitive tasks. |
| Veracity | Data may contain irregularities or incorrect values, affecting data reliability and truthfulness. |
| Value | Extracting valuable insights from data isn't always straightforward and can challenge organizations and data managers. |
| Variability | Data with high variability introduces challenges regarding data freshness and timeliness. |
| Visualization | Visualizing extensive datasets in static or dynamic forms can be a complex task. In [11], Agrawal et al. have presented the challenges and opportunities of big data visualization. |
| Validity | Data validity should be confirmed before processing, as invalid data are unusable and can't support decision-making. |

| Volatility | Outdated data is irrelevant and can introduce bias into data analysis. |
|---|---|
| Viability | Big data should remain live and active indefinitely, adaptable for development, and capable of generating additional data when necessary. |
| Vocabulary | Data readability and availability are closely tied to the vocabulary and terminology used. |
| Vagueness | Unclear and incomprehensible data hinder effective utilization. |
| Venue | Not all big data sources are reliable; some, such as social media, may offer inaccurate information. |
| Viscosity | The complexity of big data can directly impact data usability and effectiveness. |

In light of the numerous data quality challenges posed by the emergence of big data, ensuring high-quality data becomes crucial to fully leverage the potential of big data and its capabilities for data analytics and informed decision-making. In the next section, we will highlight the impact of poor-quality data and explore why data quality is essential for big data.

## 3.3. Why Data Quality is Crucial for Big Data?

In today's world, ensuring data quality is no longer just an advantage but a requirement for successful big data initiatives. Indeed, data quality plays a critical role in the effectiveness and reliability of big data analytics, and organizations must prioritize it to derive meaningful and accurate insights. In this section, we will shed light on the immense importance of data quality for big data and highlight five key reasons why it is crucial for successful analytics:

1. **Accurate Decision-Making**: Data quality ensures that the information used for decision-making is reliable and trustworthy. When dealing with large volumes of data, inaccuracies, inconsistencies, or errors can lead to erroneous analysis and, therefore, to wrong decision-making. Recent research [12] by marketing analytics platform Adverity found that 63% of chief marketing officers (CMOs) make decisions based on data, but 41% of marketing data analysts are "struggling to trust their data."

2. **Enhanced Customer Experience:** Big data analytics provides an opportunity to gain valuable insights into customer behavior, preferences, and sentiment [13]. For example, an e-commerce company used customer data analytics to personalize product recommendations and marketing campaigns. However, when inaccurate or incomplete customer data was used, customers received irrelevant offers, leading to dissatisfaction and decreased engagement.
3. **Data Compliance:** The relationship between data quality and compliance is closely connected. Indeed, regulations such as the General Data Protection Regulation (GDPR) impose that organizations rectify any inaccuracies or deficiencies in personal data. Thus, data quality standards become imperative for businesses to ensure the precision of their information.
4. **Data Integration and Collaboration**: In many organizations, data is generated and stored across multiple systems, departments, and sources. In such scenarios, data quality becomes crucial for successful integration and collaboration. For example, mismatches in data field names, missing or incomplete data, or incompatible data types can arise during integration, especially when data sources are inconsistent in data structure and format, leading to more biased data quality.
5. **Cost Efficiency:** Big data initiatives often require significant investments in infrastructure, tools, and skilled personnel [14]. Poor data quality can result in wasted resources and increased costs. For example, inaccurate data may lead to unnecessary or incorrect analysis, which consumes time and effort. Poor data quality can lead to missed opportunities or flawed insights, resulting in inefficient resource allocation. A report [15] by IBM estimated that poor data quality costs the US economy approximately $3.1 trillion annually.

By prioritizing data quality, organizations can unlock the true potential of big data, gain a competitive advantage, and drive innovation and growth. Recognizing the significant importance of data quality, professionals and researchers have dedicated increased attention to this area. Extensive studies in the literature have been conducted to explore and understand the intricacies of data quality, which will be further examined in the next section.

# 4. Beyond Data Quality: Metadata Quality in the Era of Big Data

Effective utilization of data analytics involves the presence of a well-structured metadata. This latter, often described as "data about data"[16], is indispensable as it is the repository of crucial information required for processing and making efficient use of data, encompassing elements like content, descriptions, and context. As for data, the emergence of big data has raised new challenges related to metadata management and quality. The accumulated metadata typically exhibits deficiencies in terms of consistency and quality. Furthermore, it frequently shows irregularities such as empty fields, duplicate entries, or conflicting values. As data, poor metadata quality bias data analytics and lead to skewed outcomes. Thus, it is vital to enhance the metadata quality to ensure data analysis's effectiveness and accuracy.

## 4.1. Metadata Quality Dimensions

Metadata quality was initially defined based on seven quality dimensions [17]. Subsequently, additional metadata quality attributes were proposed within the academic domain to encompass supplementary aspects of metadata quality, including shareability, extensibility, and version management [18] [19]. In the subsequent sections, we introduce the most common metadata quality dimensions, which we categorize into four fundamental aspects: Usability, Reliability, Availability, and Regularity, as shown in Figure 4.

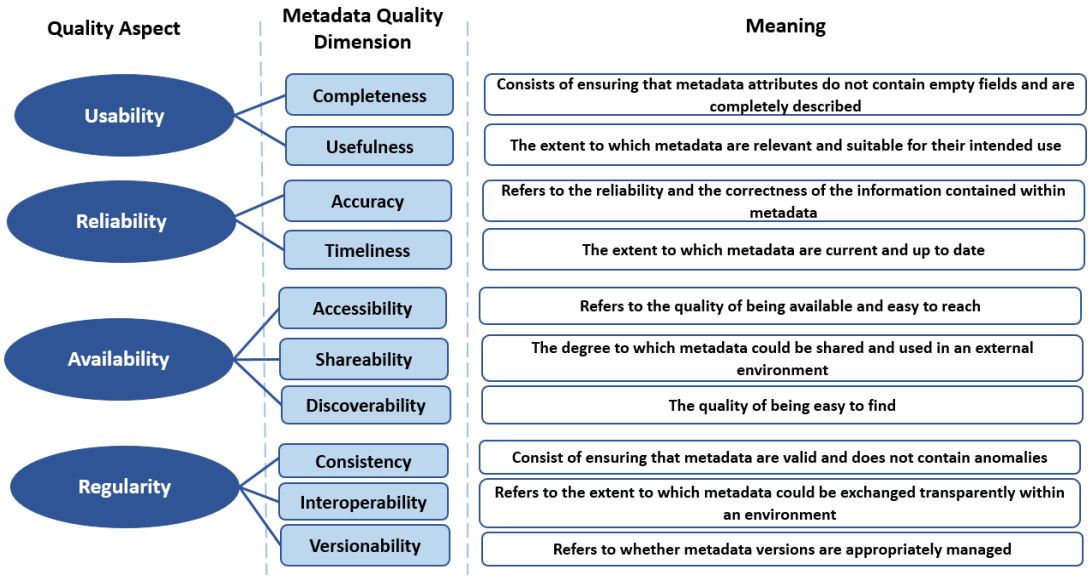

*Figure 4 : The Metadata Quality Dimensions*

These metadata quality attributes are closely interconnected. Enhancing a specific aspect of metadata quality can influence other facets of metadata quality. Consequently, individuals responsible for data management should grasp the interdependencies among metadata quality attributes before embarking on initiatives to enhance metadata quality. Moreover, it is crucial to recognize that metadata quality dimensions are profoundly context-dependent, and their interpretation may vary according to the application's context and the metadata processing phase. Importantly, it should be underscored that metadata quality evaluation should not be confined to a particular stage within the metadata management process. Instead, these metadata quality attributes should be considered throughout the diverse phases of metadata processing. In the subsequent sections, we introduce the distinct stages of the metadata management process, commonly called the "Metadata Value Chain." Subsequently, we elucidate how these metadata quality dimensions can be applied throughout the value chain.

## 4.2. New Metadata Value Chain for Big Data

The "Metadata Value Chain" (MDVC) is a systematic process delineating the distinct phases through which metadata goes within enterprises, enabling the extraction of metadata's intrinsic value. The MDVC presented below (Figure 5) encompasses four core stages: Metadata Collection, Metadata Storage, Metadata Provisioning, and Metadata Maintenance.

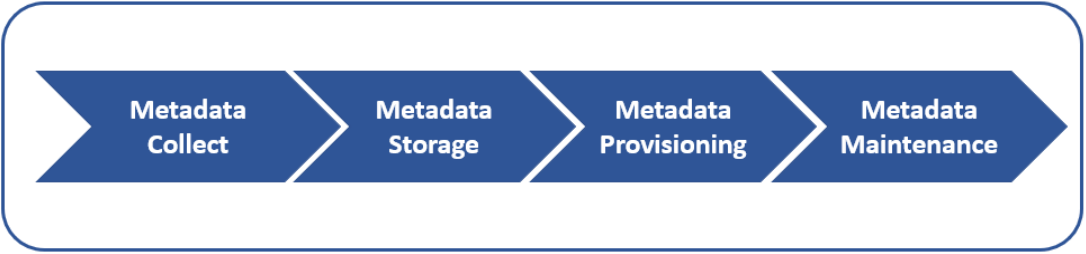

*Figure 5 : The Metadata Value Chain*

**Metadata Collect:** The initial stage entails retrieving information from diverse repositories. In big data systems, extracted metadata is typically diverse, encompassing structured and unstructured data of various types, such as text and numerical data. Following retrieval, metadata is enriched with supplementary information about the collection process and the associated semantic layer. Moreover, additional metadata may be generated to support metadata processing. It is pertinent to note that while many big data ingestion tools can be employed for metadata collection, certain technologies are explicitly tailored for handling metadata, such as Apache Tika, Open Calais, and Apache Atlas, which is widely recognized as a metadata framework.

**Metadata Storage:** Upon collecting and extracting metadata, these records are stored within a dedicated database termed the "Metadata Repository." Metadata is stored in a data lake, a centralized platform permitting data integration from numerous sources.

**Metadata Provisioning:** This phase is dedicated to making the stored metadata accessible for internal utilization, encompassing data analytics and visualization procedures. Additionally, metadata can be disseminated to external users or consuming applications via specialized APIs, such as open data portals. Nevertheless, this provisioning process necessitates the formatting of metadata according to specific metadata publishing standards, including Common Warehouse Metamodel (CWM), XML Metadata Interchange (XMI), and Web Ontology Language (OWL).

**Metadata Maintenance:** The final stage within the essential MDVC phases centers on facilitating all activities and changes linked to metadata while upholding the quality of metadata. Metadata is integral in storing information related to any data processing. It should be emphasized that stored metadata is dynamic and subject to continuous updates and enrichments by including new entries. Therefore, this phase extends throughout the entire big data value chain, requiring sustained maintenance for as long as data is processed. Advanced metadata management tools exist in big data environments like Cloudera Navigator and Apache Atlas. Nonetheless, it is imperative to underscore that these tools are most effective when integrated as part of a holistic metadata management strategy that aligns with the requisite resources and fulfills the business demands.

It is vital to comprehend that the Metadata Value Chain elucidated above is generic, encompassing phases typical to vill metadata management processes. Contextual considerations, such as the unique requirements of businesses or the nature of data, may necessitate the integration of additional processing phases. Furthermore, it should be reiterated that metadata quality dimensions should be meticulously considered throughout each stage of the MDVC to ensure the reliability and usability of metadata. The subsequent section elaborates on how the prevalent metadata quality attributes can be applied across the diverse phases of the metadata value chain.

## 4.3. The projection of the Metadata Quality Dimensions Throughout the Metadata Value Chain

This section highlights how metadata quality dimensions could be employed to support the different phases of the MDVC. For this, we present in the following table the metadata quality dimensions that should be considered at each stage of the MDVC. Additionally, practical solutions to address the raised quality concerns are suggested.

*Table 4 : The projection of the Metadata Quality Dimensions Throughout the Metadata Value Chain*

| MDVC Phase | Metadata Quality Dimensions | Explanation |
|---|---|---|
| Collect | Accuracy | Accuracy is the foremost quality dimension to scrutinize during metadata collection. In the context of big data, the precision of metadata is paramount in two dimensions:<br>**Accurate Provenance:** Given that metadata originates from multiple, sometimes unreliable sources, establishing the credibility of the metadata source is essential. This involves evaluating the originating organization's expertise in metadata standards, classifications, and any transformations applied to the metadata.<br>**Accurate Content:** Metadata should accurately describe their associated entities. Data inspection methods like sampling and profiling techniques can be employed to ensure content precision. |
| | Usefulness | Metadata within big data environments extends beyond basic data descriptions and stores information related to data storage, processing phases, and even data analytics results. These metadata records may also contain context information irrelevant to the current setting. Therefore, it is imperative to cleanse and filter collected metadata, retaining only the pertinent information. Techniques such as parsing, outlier detection, and statistical clustering are helpful. Additionally, metadata collected from various sources may exhibit redundancies, which can be addressed using metadata mapping techniques. |
| | Completeness | Post metadata collection, augmenting these records with pertinent information about the metadata source and collection process is essential. Specific data may be absent in the original context, either assumed knowledge or not applicable [20]. Consequently, it is imperative to address metadata comprehensively in two aspects: firstly, by ensuring that metadata encompasses all essential attributes, and secondly, by confirming that metadata values furnish all the requisite information. In [21], Guiding questions and best practices have been proposed to enhance metadata completeness in the realm of big data, such as identifying the pivotal metadata components necessitating completion, assigning a completeness rating to these elements, and assessing the repercussions of insufficient metadata completeness on the utility of metadata. |
| | Timeliness | Metadata collection and data collection may not be synchronized, and specific metadata may require time for generation, description, and cataloging, particularly when human intervention is involved [17]. This asynchrony can substantially impact the metadata management process, mainly when metadata contains critical information or time-sensitive analyses are imperative. In such scenarios, a thorough assessment of the risks associated with delayed metadata availability is a prudent measure before commencing metadata collection. |
| Storage | Consistency | A prevalent challenge in the storage phase is metadata inconsistency. As metadata sources utilize disparate models, the gathered metadata typically comes in diverse formats, potentially leading to anomalies, especially when all metadata is stored within a single repository. Inconsistencies may also pertain to metadata values not adhering to uniform forms or value ranges, such as disc and time attribute design. Standardized metadata models and norms should be delineated and harmonized to mitigate these challenges across all metadata records. Additionally, a conflict analysis mechanism should be implemented to assess the impact of such anomalies on data utilization. |
|  | Interoperability | Metadata stored in distributed repositories frequently employ distinct metadata models and schemas. Implementing a metadata exchange mechanism is essential to ensure interoperability across these repositories. Common Warehouse Metamodel (CWM) is a prevalent choice for enabling metadata exchange in distributed heterogeneous repositories. Nevertheless, adopting distributed metadata storage introduces novel challenges, including metadata inconsistencies and redundancy. Authors in [22] [23] have proposed new architectural frameworks to manage distributed metadata and foster interoperability effectively. |
| Provisioning | Shareability | Even if metadata within a local repository is of high quality, its quality may diminish when distributed in another environment. Therefore, "Shareability" has been introduced as a metadata quality dimension characterizing metadata that retains high quality and can be effectively utilized outside its local repository. Ensuring metadata shareability involves associating metadata with knowledge entities that elucidate its content. This can be achieved using RDF annotations and ontologies, representing metadata in a readable and machine-processable format. |
| Provisioning | Discoverability | A primary challenge for data managers is the discovery of requisite information. Difficulties locating data are frequently linked to the lack of structured and well-documented metadata that would make data discoverable. Enhancing metadata discoverability requires that metadata is appropriately linked to their associated entities and adequately documented. Utilizing ontologies is strongly encouraged, improving data and metadata discoverability likelihood. Moreover, metadata should be structured using controlled vocabulary and linked to semantic annotations to bolster semantic search visibility. |
| Provisioning | Accessibility | Prioritizing metadata accessibility is imperative, as inaccessible and unreadable metadata carries limited value. Metadata serves both internal and external purposes. Therefore, data providers must manage access permissions to provide the requisite access to metadata without compromising metadata security and usability. |

| Maintenance | Timeliness | Big data exhibits significant variability, necessitating continuous updates to maintain data timeliness. Most metadata management tools for big data facilitate the propagation of data updates to associated metadata. However, not all metadata updates can be automated, especially when metadata is closely intertwined with business processes, necessitating human oversight. |
|---|---|---|
| | Versionnability | Metadata versioning entails the creation of new metadata models in the presence of significant changes to metadata. Importantly, even upon adopting a new metadata version, previous versions must be archived and made accessible within the data system for various reasons, including recovering metadata from earlier versions, tracking metadata upgrades, enhancing previous versions, and comparing metadata versions. A version control system should be instituted to provide a collaborative workspace to manage metadata versioning and address metadata update conflicts. |

## Conclusion

In this chapter, we have covered the historical evolution of big data, its defining characteristics, and the associated tools. Additionally, we presented the different steps of big data processing, known as the Big Data Value Chain. Then, we delved into the other dimensions that define data quality as a multifaced concept. Moreover, we highlighted how big data characteristics impact data quality dimensions, making data quality a critical concern for big data. Furthermore, we went beyond the data quality to explore the realm of metadata. We have presented the different steps metadata goes through and the quality dimensions that should be considered in each stage for enhanced metadata quality. Building upon the foundational knowledge presented in this chapter, we explore the next chapter extensively, existing research, and methodologies related to big data quality. This thorough exploration of the current state of the art is fundamental in establishing a solid foundation for a comprehensive comprehension of the field of data quality, particularly the challenges that serve as the driving force behind the contributions made in this thesis.

# Chapter 3: Exploring the Landscape, A Review of Related Research in Big Data Quality

Data quality is a critical aspect of any data-driven process, as the accuracy, completeness, and reliability, among other elements of data quality, directly impact the effectiveness and credibility of subsequent analyses and decision-making processes. In the context of big data, where vast volumes of diverse and dynamic data are generated at an unprecedented scale, ensuring data quality becomes even more challenging. This chapter delves into the domain of data quality for big data, exploring the current state-of-the-art methodologies, their strengths, limitations, and the gaps that still need to be addressed. First, we delve into the historical roots of data quality to explore why and how data quality was introduced in the literature. Following this historical trajectory, we examine the transition from data quality to big data quality by highlighting the challenges and the reasons behind this shift. After an in-depth exploration of big data quality's inception, we shed light on the first primary axis of this thesis consisting of data quality assessment as previously outlined. For this first axis, we meticulously investigate how data quality has been assessed and the existing metrics used. Moving forward, we delve into the second central axis of this thesis by exploring the different anomaly detection approaches suggested in the literature. Then, we shed light on the last data quality axis, which revolves around data quality anomaly correction, by reviewing the most recent approaches suggested. Finally, we discuss the primary statements that stand out from this survey, highlight the gaps that need to be addressed, and present the main contributions of this thesis that aim to bridge the raised gaps and pave the way for enhanced data quality management. Figure 6 shows the sequence of this reviewing chapter, as described above.

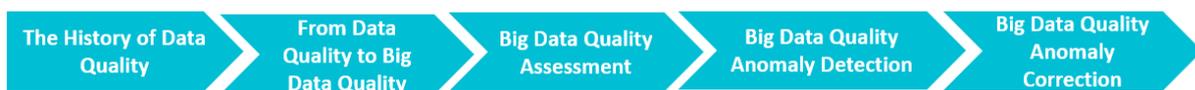

*Figure 6 : Reviewing Chapter Schema*

# 1. The history of data quality

In the previous chapter, we thoroughly discussed the foundational aspects of data quality, including its dimensions, tools, and challenges. Yet, for a deep understanding of these aspects, it is crucial to explore their historical background and origins to understand why and how these components came into existence. As mentioned earlier, data quality emerged long before the emergence of big data. Indeed, the first thoughts about data quality date back to the 19th century. This period has witnessed the rise of modern statistics and the early recognition of the significance of data accuracy and quality. The works of scholars like John Graunt, Francis Galton, and Florence Nightingale [24] laid the foundation for the future development of statistical methods, data visualization techniques, and the integration of data-driven decision-making in various fields, including government policies and public health initiatives. As the use of statistics expanded, particularly in government and public health contexts, the focus on data accuracy and quality grew substantially. Decision-makers recognized the value of reliable data in formulating effective policies and making informed decisions. The mid-20th century marked a significant turning point in the history of data processing with the advent of electronic computers. This technological advancement revolutionized how data was collected, stored, and processed, automating many tasks previously done manually. While computerized data processing offered numerous benefits, it also brought new challenges related to data quality. As data collection and storage became more automated, ensuring the accuracy, integrity, and consistency of the data entered into the computer systems became crucial, as data entry errors, hardware malfunctions, and software bugs led to inaccurate or corrupted data. In the late 20th century, the growing reliance on data for decision-making across businesses and organizations highlighted the need for systematic approaches to ensure data quality. This period witnessed the introduction of data quality frameworks and methodologies to establish guidelines and best practices for maintaining and improving the data quality used in various applications. One of the earliest and most influential data quality frameworks is the "Total Data Quality Management" (TDQM) model [25] proposed by Larry English in the late 1970s. The TDQM model focused on treating data as a strategic business asset and highlighted the responsibility for data quality throughout the entire organization. Since the introduction of TDQM, many other data quality frameworks and methodologies have emerged, each offering its unique perspective and set of guidelines for managing data quality. Among these studies, we can cite [26] where Wang et al. explores how data quality goes beyond the traditional focus on accuracy and encompasses other critical dimensions such as completeness, timeliness, currency, and relevancy. Afterward, the same author proposed one of the most common classifications of data quality dimensions [27] that fall into four distinct categories: Intrinsic, Contextual, Representational, and Accessibility. Later, many data quality approaches were suggested in the literature. In [28], the authors have explored existing data quality practices and introduced a novel framework, based on research and experience, for managing data quality in DW. The framework presents a systematic approach to define, establish, and sustain data quality management within the data warehouse environment based on the following dimensions: Accuracy, Completeness, Timeliness, Integrity, Consistency, Conformity, and Record Duplication. Also, in [29], the authors have introduced a novel approach for evaluating medical data quality. The main concept of their study involves conducting a credibility test to eliminate unreliable data before proceeding with data quality assessment. The proposed methodology consists of implementing the assessment methodology based on the Analytic Hierarchy Process to analyze the assessment results. Recent studies on data quality have shown a notable shift towards emphasizing automated methodologies based on machine learning. In [30], the authors have introduced a novel data-centric approach for monitoring data quality in AI-based applications. The authors address the challenge of integrating data from various heterogeneous sources where global domain knowledge is often lacking. They propose an automated method using reference data profiles and statistical models representing the desired data quality. Unlike traditional approaches that rely on domain expert-defined rules, this method is entirely automated, from initialization to continuous monitoring. This approach is realized in a data quality tool called DQ-MeeRKat, which was evaluated using six real-world telematic device data streams. Likewise, in [31], the authors propose an innovative method for data quality assurance based on visually representing the neighborhood relationship. They provide concrete examples to ensure defined quality aspects and demonstrate its applicability using the MNIST training data. The new tool allows for deeper insights into qualitative and quantitative aspects of the data, enabling more efficient and quantitative requirements for the quality assurance process. In [32], the authors propose an approach that clusters data field values based on their syntactic similarity. The suggested approach empowers domain experts to explore value heterogeneity, configuring clusters based on their domain knowledge. By gaining insights into the rules and practices of data acquisition and potential violations, experts can identify data quality issues in the acquisition process, system, data model, and transformations. Other recent approaches to data quality were suggested in [33] [34] [35] [36] [37]. While the literature has extensively addressed data quality concerns within traditional datasets, a fresh set of challenges arises with big datasets. This emergence introduces a rich and substantial field for researchers to explore. In the next section, we explore this transition from data quality to big data quality.

## 2. From Data Quality to Big Data Quality

The emergence of big data has brought about new and complex challenges related to data scalability, variety, and velocity. Batini et al. have shown in their article [10] " From Data Quality to Big Data Quality" that the evolution from data quality to big data quality is related to the fact that big data involves various new data types, data sources, and application domains. Also, a recent survey [38] about big data quality has stated that the main challenges of big data quality are related to the particular characteristics of big data, the data source issues, and the technology limitations. Thus, new studies and approaches have been suggested in the literature to address the challenges related to big data quality in different domains. Indeed, in contrast to the initial data quality methods discussed in the preceding section, a considerable proportion of the big data quality approaches are domain-specific and are more tailored to address the particular quality requirements of each specific area. Social media is a frequently explored domain in the literature, especially applying big data on social data sources for analyzing customer insights, including sentiment analysis and opinion mining. El Alaoui et al. [35] have emphasized vital big data quality metrics in each phase of the BDVC and their direct impact on sentiment analysis (SA) reliability and performance. They assessed how each data quality metric influenced the SA model's accuracy. In a similar context, authors in. [49] presented a new reference architecture for evaluating social media data quality, extending prior research. Their study's essential contribution was raising and considering the metadata management architecture comprising quality rule definition, metadata creation, and quality evaluation using established metrics. Big data analysis has also found extensive use in healthcare, offering benefits like disease prevention, clinical decision support, and enhanced patient care. The authors proposed a 7-step methodology for assessing medical data quality [42], involving a credibility test to remove unreliable data before quality assessment. Also, Serhani et al. [30] introduced a novel process for improving big data quality, evaluating it using the Sleep Heart Health Study dataset. The evaluation included completeness, consistency, accuracy, and processing time metrics. Likewise, [22] employed Latent Semantic Analysis to identify relevant quality dimensions in the healthcare domain. In the banking sector, the authors in [39] focus on investigating anomalies in the banking sector arising from big data technologies, credit card discrepancies, and the application of a toolkit for detecting incongruities in specific Wireless Application Protocol instruments. It is worth noting that the main challenge faced by all these approaches consists of adapting conventional algorithms and models to meet the distinct requirements of big data. As detailed in the big data tools section of the prior chapter, big data operates with its particular toolkit and architectural framework, which must also be integrated into the quality approaches for big datasets. Big data has significantly impacted the industry, but balancing data quality and security remains challenging. Talha et al. [40] examined the tradeoff between data quality and safety, revealing points of convergence and divergence between the two. While accuracy and integrity are shared properties, data security may restrict access, while data quality requires comprehensive information access. In the context of weather monitoring, the authors in [41] have proposed a preprocessing framework, addressing various aspects of raw data through cleansing, noise filters, integration, filtering, and data transformation. The framework was applied to a weather monitoring and forecasting application, incorporating global warming parameters and issuing advance alerts to users and scientists. As for data quality, recent studies on big data quality have shown a trending preference for automating big data quality management using intelligent and effective methodologies based on artificial intelligence. In [42], the authors proposed a novel approach to assess accuracy without prior knowledge using Apache Spark. The process involves a learning phase with word embeddings based on Google's Word2Vec tool, a matching phase using record linkage, and an assessment phase computing the distance between correct and new datasets. The method is applied to define contextual accuracy for three datasets from different contexts, determining the target variable based on the machine learning algorithm that yields the highest accuracy score. Other recent and pertinent approaches to big data quality were suggested in [43] [44] [2] [45] [46]. Based on the above state-of-the-art review, we can draw the following key points:

- Despite the numerous studies on big data quality, the field is still in its early stages, and various challenges must be addressed. Researchers and practitioners must work together to overcome data quality assessment, integration, and consistency issues as big data evolves.

- As highlighted in the previous chapter, big data has many particular characteristics. However, a notable observation is that most approaches addressing big data quality focus primarily on volume, overlooking other crucial big data characteristics such as variety and velocity.

- The current approaches addressing big data quality have significant limitations, particularly in managing quality dimensions and considering various aspects. Many critical data quality elements are often ignored, leading to potential inaccuracies and incompleteness in the overall data quality assessment process. There is a need to expand the scope of these approaches to encompass a more comprehensive set of quality dimensions and aspects to ensure a robust and accurate evaluation of big data quality.

- While the preprocessing and processing phases of the BDVC have garnered significant attention in the literature, existing approaches should be expanded to encompass all BDVC phases to ensure data quality is maintained throughout the entire big data life cycle.

- Understanding the interdependencies between Data Quality Dimensions (DQDs) is crucial. A comprehensive data quality evaluation should consider how modifications or enhancements in one aspect might positively or negatively influence other dimensions.

- As big data streams continuously and rapidly, there is a need for real-time data quality management solutions. Hence, novel approaches should be explored to assess and maintain real-time data quality, facilitating timely and accurate decision-making.

- Understanding the costs associated with data quality improvement efforts and the potential benefits to businesses and organizations is essential. A thorough cost-benefit analysis can justify investments in data quality initiatives and motivate stakeholders to prioritize data quality enhancement.

A notable realization emerges through an in-depth exploration of data quality in literature: the conventional qualitative approaches are insufficient for effectively addressing this critical concern. Consequently, the need to adopt a quantitative perspective becomes evident. This shift in perspective has raised new considerations surrounding data quality measurement and the definition of corresponding assessment metrics. In the upcoming section, we will explore how the concepts and implications of data quality assessment have evolved throughout history.

## 3. Data Quality Assessment

Data quality assessment has been extensively explored in the literature over the years. Indeed, the first paper introducing data quality assessment dates back to 1998 [47], when Wang showed an analogy between data quality and product quality and defined a cyclical methodology that allows a continuous improvement of data quality. In the same year, Jeusfeld et al. [48] defined the (Data Warehouse Quality) DWQ methodology and explored the relationship between quality objectives and design options in data warehousing, considering stakeholder perspectives. It classifies quality goals, defines metadata views, and addresses data and software quality dimensions, with measurement methods for each. Later, Lee et al. presented the AIMQ method [49], introducing a novel approach to assess information quality using a questionnaire to measure various dimensions of information quality. Since then, the literature has seen the introduction of several data quality assessment methodologies such as TIQM (Total Information Quality Management) [50] and CIHI (Canadian Institute for Health Information methodology) [51]. These methodologies have played a significant role in advancing data quality assessment practices. However, it is important to note that these works have primarily focused on outlining the assessment methodologies rather than providing comprehensive and detailed data quality measurements. As previously discussed, data quality is a complex concept that can be approached from various dimensions. Several quality dimensions have been identified in the literature, including well-known ones like accuracy, consistency, and timeliness. These dimensions have been further expanded and customized to catch specific requirements across diverse applications. We recommend referring to the previous section, which outlines the Data Quality Dimensions (DQD) in detail for a comprehensive understanding of data quality dimensions. Even if data quality dimensions have allowed us to define various aspects of data quality, they may not be sufficient for quantifying data quality effectively. As elucidated in the first section about the history of data quality, the early data quality methods were predominantly qualitative rather than quantitative. Thus, there was a need for quantitative and measurable elements that could provide interpretable insights into the quality of data. To address this need, data quality metrics emerged as a complementary solution to provide objective and measurable data quality assessments, offering a more precise and quantitative approach to evaluating data quality. In [52], The authors have defined a metric of data quality dimension D as a function that associates an entity with a value, usually falling within the range of 0 to 1, to quantify the data's quality level concerning the dimension D. Even if they have not provided definitions for quality metrics, Pipino et al. were there first to tackle data quality metrics and have proposed the DQA ( Data Quality Assessment) [53] methodology that aims to establish general principles for defining data quality metrics. Unlike metrics that address specific issues in particular scenarios, DQA identifies common quality measurement principles in previous research. The methodology classifies metrics into categories of subjective (measuring stakeholder perceptions) and objective (task-independent and task-dependent). Later, more specific requirements were defined for data quality metrics, such as in [54,] where the authors have proposed five requirements for data quality metrics to support economically focused data management. The suggested requirements encompass the presence of both minimum and maximum metric values (R1), the metric values being subjected to interval scaling (R2), the evaluation of the configuration parameters' quality and the derivation of metric values (R3), the proper aggregation of the metric values (R4), and the metric's economic efficiency (R5). Practical implications are also discussed in the context of applying these requirements. With the pressing need for more tangible metrics to assess data quality, there has been specific research that has developed a more explicit definition of data quality metrics. In [55], the authors have created an interactive environment, enabling users to customize data quality metrics according to their specific requirements. The research defined the following quality metrics: Completeness, Validity, Plausibility, Time Intervals, and Uniqueness. Thus, it is essential to note that the dimensions outlined in the previous chapter were not established all at once; rather, they progressively emerged through diverse studies that explored this topic. Always in the context of data quality assessment, the authors introduce in [56], the Heterogenous Data Qu ty Methodology (HDQM), designed for comprehensive Data Quality (DQ) assessment and enhancement, encompassing various data types in an organization (structured, semi-structured, and unstructured). A meta-model is proposed to describe the relevant knowledge incorporated in the methodology, and a unified conceptual representation is used for different data types. The methodology focuses on defining two crucial quality metrics in the literature: Accuracy and Currency. In [57], the authors introduce a framework for scientific data quality measurement using rule-based principles. The framework addresses data quality issues arising from incorrect execution and data collection and validation process descriptions. The framework allows for handling uncertainty and was implemented using survey data. However, the proposed approach is generic and does not specify quality metrics. In [58], the authors have introduced novel data quality metrics that consider the significance of information within the data. The defined metrics are Completeness, Relevance, Accuracy, Timeliness, and Consistency. The suggested metrics were also implemented. As the significance of data quality continues to grow in various domains, there has been a surge in the development of recent approaches for data quality assessment. In [59], the authors addressed the data quality challenges in Structural Health Monitoring systems, which are crucial for structural integrity management. The authors propose quality indicators to fill the gap and suggest deterministic and probabilistic metrics to handle uncertainties. In [60], the authors introduce a novel method for assessing data quality in blockchain-based applications used in enterprise business activities. As data from various sources may lack consistency and have differing representations, assessing their credibility and value poses challenges. The proposed method addresses this by evaluating consistency through block information similarity. A trustworthiness characterization method based on information sources and assesses data reliability. In [61], the authors introduce "QualiProvOER," a semi-automatic approach for assessing Open Educational Resources (OER) quality based on data provenance. The proposed "ProvOER Model" comprises essential metadata to describe OER history. The study outlines criteria and mathematical formulas employed to evaluate OER quality, emphasizing the influence of review and remix criteria on OER provenance for quality assessment. Other recent data quality approaches were suggested in [62] [63] [64].

The emergence of big data has brought significant data quality challenges, prompting the development of new metrics and the recognition of new requirements for data quality. A recent study [65] has stated that the primary data quality assessment challenges are related to the complexity and diversity of data sources, real-time processing needs, and vast data volume. Consequently, researchers and practitioners have responded by proposing innovative metrics and refining existing methodologies to meet the evolving demands of big data quality assessment and management. [66] introduces a hybrid approach to evaluating Big Data quality throughout the value chain. It first assesses the quality of the Big Data itself through cleansing, filtering, and approximation, followed by an evaluation of the processing and analytics processes. Experiments were conducted to evaluate data quality before and after pre-processing and the quality of the pre-processing and processing on a large dataset. The study measured quality metrics for accuracy, completeness, and consistency. In [67], the author proposes an efficient data quality evaluation scheme using sampling strategies on Big Data sets, reducing data size for representative population samples. The evaluation targets dimensions like completeness and consistency. Experiments were conducted on a sleep disorder dataset using Big Data bootstrap sampling techniques, demonstrating that mean quality scores of samples represent the original data. In [68], the authors proposed a Big Data Quality Profiling Model comprising components such as sampling, profiling, exploratory quality profiling, a quality profile repository, and a data quality profile. The repository oversees data quality dimensions, associated metrics, pre-established quality action scenarios, preprocessing tasks, their respective functions, and the data quality profile. The exploratory quality profiling method identifies processes to understand data quality trends and their impacts. Experiments testing various features of the suggested model show that quality profiling in the early stage of the Big Data lifecycle enhances data quality and yields valuable insights from the exploratory quality profiling methodology. In [69], the authors propose a model that extends machine learning to business applications for data quality assessment. The model detects data noises from a risk dataset, estimates their impacts using Gaussian and Bayesian methods, and uses sequential learning in multiple deep neural networks with an attention mechanism for assessment. Experimental results demonstrate the predictive power of machine learning and model effectiveness. The model is scalable and applicable to industries beyond banking that utilize big data. Among the most recent approaches that tackled big data quality assessment, we can cite [70], where the authors introduce a novel ISO-based declarative data quality assessment framework called BIGQA. BIGQA offers flexibility, supporting data quality assessment in various domains and contexts throughout the data life cycle. The framework enables data domain experts and management specialists to efficiently plan and execute customized data quality assessments on parallel or distributed computing platforms. In [71], the authors presented a data quality identification model designed explicitly for power big data. This model can efficiently detect abnormal data within vast datasets. Power data is grouped and mapped into diverse feature spaces using data augmentation technology. The model utilizes Tri-training to identify anomalous data from various power data subsets in different feature spaces. In [72], the authors proposed a data quality model based on the canonical data model, four adequacy levels, and Benford's law to assess the COVID-19 data reporting by the World Health Organization (WHO) in the six Central African Economic and Monetary Community (CEMAC) region countries. This model serves as an indicator of data dependability and sufficiency for big dataset analysis. Further development of this model is required to enhance its core concepts. Other recent studies about big data quality assessment were stated in [73] [74] [75]. Based on the above state-of-the-art review, we can draw the following key points:

- There is a lack of advanced big data quality assessment, and several challenges must be addressed. Addressing these challenges involves overcoming data volume, variety, velocity, and veracity issues.
- The growing number of suggested data quality dimensions in the big data context has highlighted the need for comprehensive data quality metrics. Existing data quality metrics do not fully capture big data's unique characteristics and challenges.
- Data quality assessments in the big data domain are often based on a limited number of metrics, which may not adequately represent the complexity and anomalies of the data. It is essential to expand the number of data quality metrics and consider a broader set of dimensions that reflect the unique characteristics of big data.
- Real-time big data quality assessment ensures timely and accurate decision-making in dynamic and fast-paced environments. Conventional data quality assessment methods are often batch-oriented, time-consuming, and inappropriate for real-time big data processing.
- Data quality assessments are context-correlated and not generic, meaning that data quality requirements are highly based on the specific domain, application, or use case.

- To effectively assess data quality in the big data context, specific metrics tailored to the unique attributes of big data need to be defined. Big data presents distinct challenges, such as scalability, heterogeneity, and high data volume, which require specialized metrics that can account for these factors.

While data quality assessment offers valuable insights into the overall data quality, it doesn't inherently uncover the anomalies hidden within the dataset. As a result, anomaly detection emerges as a critical component of data quality management, enabling the identification of irregularities within the dataset for subsequent intervention. In the upcoming section, we explore the current state-of-the-art practices in anomaly detection, encompassing the vast array of approaches proposed in this regard.

## 4. Big Data Quality Anomaly Detection

Anomaly detection is a widely trending topic in data analysis, given the increasing need for accurate and reliable data insights. Anomalies are deviations or outliers from a dataset's expected or regular pattern. These anomalies can arise for various reasons, such as errors in data collection, sensor malfunction, fraudulent activities, or rare events that may be important. As businesses and industries increasingly rely on data-driven decision-making, effective anomaly detection methods play a crucial role in identifying irregular patterns and potential issues in the data, enabling timely actions and informed decision-making. Thus, anomaly detection has attracted much attention from both researchers and practitioners due to its relevance across various domains. A prominent application of anomaly detection is in banking, which plays a vital role in fraud detection. In [76], the authors introduce a novel approach called UAAD-FDNet, a framework based on an Unsupervised Attentional Anomaly Detection Network for credit card fraud detection. By treating fraudulent transactions as anomalies, UAAD-FDNet uses autoencoders with Feature Attention to effectively separate them from the vast transaction data. Likewise, in [77], a deep learning model was suggested for credit card fraud anomaly detection in financial transactions. Besides the banking sector, anomaly detection has been applied in various other domains and industries. In healthcare, the authors propose in [78] a comprehensive framework that not only detects anomalies in healthcare data but also provides explanations for why they are considered anomalies by identifying outlying aspects.

Using real-world healthcare datasets, they have successfully detected anomalies and their remote elements, with the isolation-based outlying aspect mining measure demonstrating exceptional performance and promising results. In industry, the authors have suggested in [79] a novel multivariate time series anomaly detection framework. The framework uses an intuition-based neutrosophic representation method and an automatic learning graph structure. Through data preprocessing and representation, the framework captures fuzzy feature data using the intuition-based neutrosophic model after dimensionality reduction and co-frequency processing. The graph model is then used to identify rare events in the data and interpret sensor-level anomalies. Anomaly detection has also been applied to social platform data, such as in [80], where the authors have used anomaly detection to identify the emergence of specific diseases through health-related tweets, particularly pre-COVID-19 and during COVID-19. After pre-processing, the study collected over 44 thousand tweets and employed topic modeling using non-negative matrix factorization and latent Dirichlet allocation.

The resultant topics were used to design a query set, and anomaly detection was performed using a sentence transformer. Outlier tweets were also clustered based on similarity using K-means. Experimental results demonstrate that the proposed framework can effectively detect a sudden flow in unusual tweets distinct from regular tweets. In aeronautics, the authors in [81] have explored the effectiveness of three unsupervised deep generative models: variational autoencoders with Gaussian, Bernoulli, and Boltzmann priors in anomaly detection for multivariate time series of commercial-flight operations. The results show that the restricted Boltzmann machine model demonstrates robustness to changes in anomaly type and phase of flight. The study also guides the competitive nature of a discrete deep generative model compared to its Gaussian counterpart in anomaly detection tasks. It is worth noting that anomaly detection is not only limited to textual data but is also applied to media data, such as image processing. To detect anomalies in image data, the authors in [82] have introduced a novel model incorporating data distribution and reconstruction error for anomaly detection. The unsupervised model is achieved by training a generative adversarial network and an autoencoder, minimizing the harmful loss and reconstruction error, with both modules sharing the generator and decoder parameters. Other pertinent anomaly detection approaches were suggested in [83] [84] [85].

As for data quality assessment, the evolution from traditional anomaly detection to big data anomaly detection has significantly improved the ability to detect and address anomalies at scale. It has enabled the processing and analysis of vast volumes of data from diverse sources in real time, allowing organizations to uncover previously hidden insights and patterns that could have critical implications for decision-making and business operations. As a result, numerous anomaly detection techniques have emerged for big data. Among the most recent and relevant approaches, we can mention [86], where the authors propose a fusion model combining CNN and C-LSTM. This fusion model effectively merges spatial and temporal variables, offering faster training, convergence, and computation for resource-limited IoT devices. The experiments demonstrate that the proposed model outperforms existing deep learning approaches, delivering higher accuracy, precision, and recall in anomaly detection tasks within the IoT's time-sensitive big data environment. In [87] presents a novel security anomaly detection approach for big data platforms, addressing the challenges posed by the explosive growth of data on the Internet. The suggested method employs quantum optimization clustering and is structured upon a distributed software architecture using open-source big data technologies like Hadoop and Spark. This system proficiently identifies network anomalies by analyzing server log data on the big data platform. The authors in [88] introduce a complementary methodology called Exploratory Data Analysis for Big Data Analytics to detect abnormalities in cloud computing. This approach aids in gaining insights into data relationships without traditional hypothesis modeling, enhancing the understanding of patterns and enabling anomaly detection. The results demonstrate that Exploratory Data Analysis effectively facilitates anomaly detection and plays a crucial role in the Data Understanding phase. There have been some studies that have suggested new architectures for big data anomaly detection. In [89], the authors proposed a data streaming computational model for big data processing in smart manufacturing, which serves as the foundation for discrete anomaly detection engines. The architecture can handle the high volume, velocity, and variety of data generated in smart manufacturing and can be applied to various applications, such as predictive maintenance and quality control. Likewise, in [90], a developed infrastructure model was suggested for real-time anomaly detection in data streams, highlighting the method for software component requirements, algorithm selection, and tool improvement. The implemented anomaly detection tools, such as Dataflow, BigQuery ML, and Cloud DLP, enable secure real-time anomaly detection, as demonstrated in the application of detecting fraudulent software logs in an information system to enhance cybersecurity. Some efforts have also been made to extend the traditional data anomaly detection algorithms to fit big data requirements. The authors in [91] present a novel unsupervised machine learning approach by combining the K-Means algorithm with the Isolation Forest for anomaly detection in industrial big data scenarios. To achieve this, they have used the Apache Spark framework to implement the model, which was trained on extensive network traffic data. The evaluation of the proposed system on live streaming data demonstrates its potential for real-time anomaly detection in industrial setups. Other recent studies about big data anomaly detection were suggested in [92] [93] [94].

Even if various data quality dimensions have been proposed in the literature, as elucidated in the quality assessment section, most anomaly detection approaches focus on identifying anomalies only as outlier values and tend to overlook the other important aspects of data quality. Specifically, they do not consider the broader context of data quality and its various dimensions. As a result, there is a notable gap in the literature regarding anomaly detection within the scope of data quality. Few studies have explicitly addressed the integration of anomaly detection with comprehensive data quality assessment. In [95], the authors have introduced an innovative unsupervised approach to identify anomalies through model comparison, utilizing consensus learning and integrating heuristic rules with iterative hyperparameter tuning to enhance data quality. To assess the efficacy of this method, a case study was conducted. In [96], the authors introduced innovative strategies to improve the precision of Data Quality applications in the context of High Energy Physics experiments. They demonstrated the effectiveness of Machine Learning-based anomaly detection techniques in identifying unforeseen detector malfunctions, using the CMS experiment at the Large Hadron Collider as a practical illustration.

In a related study [97], the researchers introduced a data-cleaning method called "hierarchical reduction and classification cleaning" to tackle the challenge of inaccurate data within multi-source heterogeneous data settings. This method involves the construction of an improved tree Bayesian Tan network through the utilization of machine learning classification algorithms and attribute weighting. Empirical assessments highlight the proficiency of the proposed hierarchical reduction approach in extracting essential data, particularly in multi-source heterogeneous data environments. In [39], the authors explored the influence of Big Data technologies on the banking sector, focusing on credit card inconsistencies. They also investigated the application of the toolkit for anomaly detection in a particular Wireless Application Protocol (WAP) instrument. However, these approaches mainly focus on one aspect of data quality and overlook the other quality dimensions. Based on the above state-of-the-art review, we can draw the following key points:

- Despite the progress made in anomaly detection for big data, significant challenges still exist in effectively detecting anomalies at scale. The big volume, velocity, and variety of big data raise substantial challenges for anomaly detection algorithms. Ensuring real-time anomaly detection and processing vast amounts of data in a timely and efficient manner remain ongoing challenges.
- Many anomaly detection approaches are designed to address anomalies related to specific domains or application areas. For example, an anomaly detection model developed for fraud detection in financial transactions might not directly apply to detecting abnormalities in healthcare data. This domain correlation allows these approaches to effectively detect anomalies relevant to the specific context they were designed for.
- Currently, most anomaly detection methods focus on detecting "domain anomalies" specific to a particular field or application. However, they often overlook anomalies related to data quality. In other words, these approaches are primarily concerned with detecting outliers or irregular patterns within the data but do not comprehensively address various aspects of data quality, such as accuracy, completeness, or consistency. This gap in research presents an opportunity to explore how anomaly detection techniques can be harnessed to assess data quality more holistically.
- Most anomaly detection approaches aim to identify outlier data points that deviate significantly from the expected patterns. While these outlier-based techniques help detect unusual instances, they do not consider other data quality dimensions, such as data consistency or integrity. As a result, the full potential of anomaly detection in ensuring comprehensive data quality remains largely unexplored.

- Conventional data cleaning tools are commonly employed to address data quality issues. However, these tools may be limited in detecting latent anomalies that are not immediately apparent. More advanced and intelligent techniques are required to identify and address problems when sophisticated irregularities are hidden within large datasets. By incorporating such methods into data quality assessment, organizations can gain deeper insights into their data and ensure higher quality standards.

Detecting anomalies is crucial to extracting meaningful insights and making informed decisions, but the process remains incomplete until these anomalies are addressed and corrected. Therefore, correcting data anomalies – our final and ultimate axis – emerges as one of the paramount components for effective big data quality management. In the following section, we delve into the literature to explore how this critical aspect has been addressed, consistently reviewing the most recent and pertinent approaches suggested.

## 5. Big Data Quality Anomaly Correction

In the vast realm of data management, anomaly correction is one critical axis that cannot be overlooked. Anomaly correction is pivotal in bridging the gap between raw data and reliable, actionable information. Upon detecting anomalies, data management efforts must shift towards their resolution. Addressing quality anomalies entails exploring multiple possibilities, each with the ultimate goal of restoring data integrity and reliability. Among these possibilities, one of the simplest ways is to clear them from the dataset. In [98], the authors introduced three algorithms to eliminate statistical anomalies from datasets within the Data Science pipeline. The key advantages of these algorithms lie in their simplicity and minimal configurable parameters, which are determined through machine learning based on the input data properties. Moreover, these algorithms exhibit flexibility and do not rely heavily on the data's nature and source. The effectiveness of the proposed methods is validated through a modeling experiment implemented in Python, and the results are visually depicted using plots generated from raw and processed datasets. While removing data anomalies may seem like a straightforward approach to enhancing data quality, it is not always the best option due to potential drawbacks. One major concern is losing valuable information and insights in the anomalous data points. Instead of outright removal, various data cleaning techniques, such as imputation by mean, median, or max, interpolation, and other advanced imputation methods, can be employed to address anomalies. In [99], the authors have unveiled an innovative strategy for addressing anomalies in IoT data to improve data availability by rectifying irregular sensor data. The proposed method encompasses a feedback and voting-based feature selection procedure that extracts essential attributes from raw datasets through mini-batch data processing. These selected features are subsequently utilized for anomaly detection, and a subsequent anomaly imputation process is applied to replace aberrant data within the original datasets. It is worth noting that conventional methods of correcting anomalies, such as imputation by median, max, or min, while widely used, do not always offer the desired effectiveness. These approaches often introduce replacement values that may lack precision and accuracy, leading to potential distortions in the data. Modern machine learning models are increasingly used for more reliable anomaly correction. These advanced techniques utilize sophisticated algorithms that can learn patterns from the existing data and make informed decisions on handling anomalies. Thus, many studies have been suggested to address anomalies using machine learning techniques. In [100], the authors introduce a method based on an improved random forest that involves an outlier data recognition model using isolated forests to identify anomalies in the data. Subsequently, an improved random forest regression model is established to enhance adaptability to mixed abnormal data and accurately fit and predict the data trend. The improved random forest data cleaning method compensates for missing data after removing mixed abnormal data. Also, in [101], the authors outline a novel methodology for detecting and rectifying anomalies in sensor data, encompassing four fundamental stages: data preprocessing, identification of abnormal data, correction of detected anomalies, and tool life prediction and assessment. In the initial phase, raw condition monitoring data is subject to preprocessing to extract features and construct health indices. Historical training samples are grouped via the dynamic time-warping algorithm to pinpoint abnormal data by calculating errors against a predefined threshold. Subsequently, the identified anomalies are fine-tuned through similarity matching via the k-nearest neighbors approach employing dynamic time warping. In agriculture, the authors introduce in [102] a novel data anomaly detection and correction algorithm tailored for the tea plantation IoT system, considering the complex characteristics of abnormal data with multiple causes and features. The algorithm begins with Z-score standardization of the original data and dynamically determines the sliding window size based on the sampling frequency. The process involves constructing a Convolutional Neural Network model to extract abnormal data, then utilizing the Support Vector Machine algorithm with Gaussian Radial Basis Function and One-to-One multiclassification techniques for classifying the anomalies. Subsequently, a Long Short-Term Memory network is established for correction using multifactor historical data. To address the complexities of identifying and correcting outliers in multi-source data, a novel intelligent identification and order-sensitive correction method based on historical data mining is proposed in [103]. Firstly, an intelligent identification method for single-source data outliers is introduced, utilizing Neural Tangent Kernel K-means clustering. Next, a framework called Order-Sensitive Missing Value Imputation is developed for multi-source data. Constructing a similarity graph based on multidimensional similarity analysis, the filling order decision for missing values in multi-source data transforms into an optimization problem, facilitating optimal imputation sequence determination. Lastly, a Neighborhood-Based Imputation algorithm is presented, enhancing outlier correction in sources with missing data. Other pertinent and recent approaches for data anomaly correction were proposed in [104] [105] [106].

As for anomaly detection, the emergence of big data has brought new challenges to anomaly correction in terms of scalability, variety, and complexity. Thus, new approaches were suggested in the literature to address anomalies in the big data context. In [107], the authors analyzed the causes and distribution characteristics of different abnormal load data types. Then, they used an improved density estimation algorithm to identify and modify the anomalous data, enabling swift processing and repair of distribution network load big data. The proposed algorithm is tested on the power load data of a province in 2020, demonstrating its ability to accurately and rapidly identify and correct abnormal load data. Likewise, in [108], the authors introduced a novel model proposed to cleanse drinking-water-quality data effectively. Firstly, data that deviates from the normal distribution is identified as outliers and removed. The optimal control theory of nonlinear partial differential equations is then integrated into the cart decision tree with a specified depth, serving as a weak AdaBoost classifier. Using the suggested model, missing values in the data stream are predicted and compensated for, achieving data cleaning for drinking water-quality data. To detect erroneous measurements in the vast amount of real-time big data streams generated by sensors, the authors implement [109] an effective data cleaning process to eliminate traffic sensor faults. They also introduce a traffic model that benefits from detecting anomalous data traffic sensors capture. Also, in [110], the authors suggested a novel architecture to address anomalies in big data in a power grid. In the proposed system, the context of the sub-system is first analyzed to fulfill the overall functional process of the system. Specific application scenarios are then examined to list and describe the operational requirements of the data-cleaning subsystem.

Furthermore, the functional requirements of the power grid big data cleaning subsystem are analyzed from a software engineering perspective. In [111], the researchers studied anomalies within the banking domain, precisely irregularities concerning credit card data in the context of Big Data. The study focuses on inconsistencies found in credit card data and introduces a novel approach to identify and address these anomalies by detecting invalid data numbers. Furthermore, the study proposes effective measures for eliminating such invalid data.

The concern regarding the lack of comprehensiveness in anomaly detection was similarly identified in the context of anomaly correction. Indeed, despite extensive discussions on data anomaly correction in various domains, there is a notable gap in approaches addressing anomalies related to data quality. Existing methods focus on correcting outliers and handling missing values but overlook other quality anomalies like inconsistencies, duplicates, or unreadable values. As a result, most of the reviewed approaches focus on dealing with abnormal data values. On the other hand, numerous studies have addressed missing values, such as in [112], where the authors have introduced a method known as "Full Subsequence Matching" for handling missing values within telemetry water level data. This technique entails recognizing a series of absent values and substituting them with constant values to form a contiguous sequence. Subsequently, it identifies the most comparable subsequence from historical data and adjusts it to fill the gaps in the missing segment based on similarity. Also, in [113], the authors have introduced an innovative approach to address missing data through generative adversarial networks. This enhanced generative adversarial imputation network incorporates the Wasserstein distance and gradient penalty to manage missing values effectively. The data preprocessing stage is optimized by capitalizing on domain knowledge related to ships and utilizing isolation forests for anomaly detection.

Additionally, the proposed model is informed by statistical analyses, including ship design parameter correlation analysis, outlier analysis, and a comprehensive examination of the types of missing data. For the imputation of health data, as detailed in [114], the authors focus on addressing missing values within the diabetes dataset generated by IoMT devices. Two distinct imputation techniques are employed: statistical methods (such as zero, mean, median, and mode) and machine learning algorithms (specifically SVM and KNN) for data imputation. Accuracy is evaluated for each technique to determine its effectiveness in imputing missing data. The analysis is conducted on an electronic healthcare dataset comprising 1000 rows and eight attributes, with 10% (800) missing values. The results reveal that KNN outperforms the other algorithms, achieving a 72.87% accuracy in imputing missing data. To address missing data in marine systems' sensor data, the authors analyze in [115] the use of variational autoencoders for assigning missing values. A comparative study includes widely used imputation techniques such as mean imputation, Forward Fill, Backward Fill, and KNN. The case study focuses on marine machinery system parameters collected from sensors installed on a tanker ship's generator. The results demonstrate that variational autoencoders handle missing values in marine machinery system sensor data. Other studies addressing missing values were suggested in [116] [117] [118]. Based on the above state-of-the-art review, we can draw the following key points:

- The existing research on data anomaly correction is relatively limited compared to many studies focused on anomaly detection. This calls for significant efforts and attention in data anomaly correction, especially in big data, where unique challenges and characteristics must be considered.

- While some approaches for data anomaly correction in big data have been explored, the current coverage is insufficient to address the complexities of big data quality fully. Therefore, there is a demand for more comprehensive and tailored approaches that consider the specific characteristics of big data.

- Many existing approaches concentrate solely on specific aspects of data quality anomalies, such as outliers and missing values, neglecting other crucial elements that may impact data reliability. A more inclusive approach is required to handle the diverse data quality issues effectively.

- The prevailing approaches often exhibit domain-specific correlations, catering to specific anomalies within particular domains. To promote broader applicability, it is imperative to develop generic strategies that can handle data quality anomalies without being overly context-focused.
- There is an apparent necessity for intelligent and sophisticated methods for data quality anomaly correction, especially when dealing with correlations and patterns that surpass the capabilities of conventional techniques. Advanced methodologies are crucial to ensure accurate and precise correction of complex data anomalies.
- There is a growing need for anomaly correction approaches explicitly tailored for real-time data streams. The dynamic nature of real-time data requires specialized methods to promptly and efficiently address data quality issues as they arise.

After conducting an extensive review of the principal components within big data quality management and capturing the primary findings of each addressed component, our survey remains incomplete without a comprehensive discussion. Thus, in the following final section, we synthesize the primary outcomes from the study and highlight the gaps identified within each explored component. Furthermore, we present the main contributions of this thesis, aiming at addressing and bridging the raised gaps.

# 6. Discussion

This comprehensive review of the state-of-the-art literature has identified crucial points and gaps in data quality, data quality assessment, data anomaly detection, and data anomaly correction. One of the main gaps we have identified in all the tackled axes is the insufficient comprehensiveness in addressing data quality aspects, as most approaches tend to focus on a limited set of data quality dimensions, often overlooking the multifaceted nature of data quality. Therefore, there is a need for approaches with a more holistic perspective considering various aspects of data quality. This involves prioritizing data quality as a primary objective, particularly within anomaly detection and correction methods, where most studies are more domain-oriented than data quality-oriented. This leads us to our second key observation - the high context dependency of existing data quality approaches. Indeed, most approaches heavily focus on domain-specific knowledge, making them less adaptable to diverse contexts. While these approaches undoubtedly allow for effectively addressing anomalies within a specific field, there is an equally vital requirement for more generic approaches that prioritize data quality as the central goal without any correlation to one particular, offering a higher level of versatility and adaptability.

Another recurring point we have noted is the relative scarcity of approaches designed to address data quality concerns in a big data context compared to their traditional counterparts. This discrepancy underscores a significant gap in the literature, particularly given the distinct challenges big data pose in terms of scalability, velocity, and complexity, especially since data quality concerns might even be more pronounced than in traditional datasets due to the particular characteristics of big data. Furthermore, our review has also underscored the evolving significance of automated and machine learning-driven methodologies across data quality assessment, anomaly detection, and correction. We have noted that these approaches are generally more efficient than traditional approaches as they use the power of algorithms and data-driven insights to streamline the process, making them more efficient and adaptive. However, their adoption and integration into existing practices are still areas of ongoing research and development, representing a promising avenue for the future of data quality enhancement.

In response to the identified challenges, we aim through this thesis to address the underlined gaps with four key contribution directions. First and foremost, we aim to introduce greater comprehensiveness by expanding the scope to encompass a more comprehensive array of data quality dimensions. Secondly, we strive for greater genericity by placing data quality at the central core of our research, making it the ultimate goal, and allowing our approaches to be more adaptative and applicable across diverse domains instead of being correlated to one specific field. Moreover, given the significance of data quality in big data, we have tailored our contributions to meet the particular requirements of this context. Furthermore, we use the potential of automation and machine learning to empower our approaches with greater efficiency and adaptability. We firmly believe that the application of our collective contributions to the fundamental axes of data quality, including data quality assessment, data anomaly detection, and data anomaly correction, has the potential to make a significant impact on the current state of the art, ultimately leading to the development of a robust, automated, and comprehensive framework for big data quality management.

## Conclusion

In this section, we have conducted an exhaustive review of the existing body of work within the domains of data quality, data quality assessment, data quality anomaly detection, and data quality anomaly correction. We systematically identified and elucidated vital points and insights gained from our comprehensive analysis of the state-of-the-art literature for each of these axes. The key issues identified revolve around the insufficient comprehensiveness in addressing data quality aspects, the high context dependency of existing approaches, the limited number of studies tackling challenges specific to big data context, and the need for automated and machine learning-driven methodologies. These identified challenges drive our contributions to enhance the different axes of data quality management, including data quality assessment, data anomaly detection, and data anomaly correction. In the following sections, we delve into the first axis of data quality management: data quality assessment. Thus, we will highlight our contributions within this realm, their implementation and the obtained results.

# Chapter 4: A Comprehensive Approach for Data Quality Assessment

## 1. Introduction

In the past decade, data analytics has emerged as a powerful tool for businesses to enhance their operations and connect better with their customers. Data is now recognized as a crucial resource for decision-making in various industries. However, the usefulness of data depends on its quality. Using unstructured and inaccurate data can introduce bias into data analytics and lead to incorrect managerial decisions. Therefore, data quality has become a focal point for both academic researchers and organizations, resulting in the development of various approaches to address this issue. Despite the importance of high-quality Big Data, there has been limited work on big data quality assessment. Indeed, most current methods for data quality assessment in big data rely on traditional measures and overlook specific issues unique to big data.

Moreover, while the number of defined dimensions has expanded significantly, few metrics are currently measured, leaving a gap that needs to be addressed. Furthermore, data weights, which greatly influence measurement accuracy, are often disregarded by existing quality assessment frameworks, raising concerns about the precision of the measurements. To tackle these challenges, this chapter aims to contribute to the ongoing discussion on big data quality assessment with four main contributions:

1) The introduction of four novel data quality metrics: Accessibility, Integrity, Security, and Ease of manipulation.
2) The formulation of a comprehensive Big Data Quality Assessment Framework encompasses 12 metrics: Timeliness, Completeness, Volatility, Conformity, Uniqueness, Ease of Manipulation, Consistency, Readability, Relevancy, Security, Integrity, and Accessibility.

3) Enhanced precision in measurement through incorporating data field, data quality metric, and data quality aspect weights.
4) Offering a holistic view of Big Data Quality by establishing and evaluating five quality aspects: Availability, Reliability, Pertinence, Validity, and Usability.

The remainder of this chapter is as follows: Section 2 highlights the importance of big data quality metrics. Section 3 reviews the most recent approaches to data quality assessment. Section 4 presents the suggested big data quality assessment framework. Section 5 describes the implementation of the proposed framework.

## 2. The importance of considering BDQM

Nowadays, incorporating data quality metrics has become a standard practice in various domains and contexts of big data processing. The rise of Big Data has further boosted the importance of data quality metrics, as ensuring data quality has always been a top priority in big data environments, given the prevalent poor quality of big data. Several advantages arise from considering Big Data Quality Metrics (BDQM). Among them, we can cite the following:

**Providing measurable insights into data quality:** BDQM assessment offers valuable insights into the data's health, enabling data managers to make an accurate idea about the quality state of data. A study [26] exploring the influence of data quality on businesses has revealed that less than 50% possess confidence in the quality of their internal data, with only 15% expressing satisfaction with the data supplied by third-party sources.

**Thresholds and Standards:** Organizations can set predefined thresholds and quality standards for each BDQM. When data falls below these thresholds, it indicates poor data quality, while data meeting or exceeding the set standards indicates good data quality. These thresholds act as benchmarks to distinguish between acceptable and unacceptable data quality levels.

**Identifying data quality defects:** Measuring metrics raises awareness of hidden data weaknesses in data that may go unnoticed and allows data managers to identify data defects. A study [43] that evaluated the effects of Big Data Quality Metrics (BDQM) when employing sentiment analysis techniques for big data has demonstrated that completely disregarding big data quality metrics significantly skews prediction outcomes, resulting in a sentiment analytical accuracy of only 32.40%. This underscores the compelling impact of neglecting quality metrics for big data on the accuracy of predicted results.

**Leading to enhanced data quality:** Assessing BDQM allows organizations to manage data quality proactively. Regular assessments enable data managers to monitor data quality over time and take corrective actions based on the measured metrics before poor data quality adversely impacts decision-making processes and downstream applications.

**Root Cause Analysis:** By using data quality assessment metrics, organizations can perform root cause analysis to understand the underlying reasons behind data quality issues. This helps identify the sources of poor data quality, such as data entry errors, integration issues, or data processing inaccuracies. Addressing these root causes leads to sustained data quality improvements.

**Data Integration and Interoperability:** Data quality assessment metrics facilitate integration and interoperability. When organizations exchange data with external partners or merge data from different sources, assessing the quality of the integrated datasets becomes essential to ensure data consistency and accuracy. With the immense importance of BDQM, data quality measurement has gained increasing attention from practitioners and researchers. Thus, many approaches to data quality metrics have been suggested in the literature. The following section provides a more in-depth exploration of the data quality assessment.

## 3. Big Data Quality Assessment

Data quality assessment stands as a crucial area in the realm of data management and analysis. As mentioned in the previous chapter, there has been a significant shift from qualitative assessment methodologies focused on subjective evaluations to quantitative methods that offer measurable metrics and objective insights about data quality. Despite the advancements made in the field of data quality assessment, there are still areas within this field that require further attention and improvement. Indeed, while well-established dimensions like completeness, uniqueness, and timeliness have gained significant attention and have transparent assessment methodologies, there remains a notable gap in addressing the less common metrics such as conformity, readability, or volatility. These emerging data quality dimensions are equally vital but have not received the same level of attention in the existing literature. As a result, there is a noticeable gap in terms of comprehensive definitions and standardized assessment techniques for these metrics. This gap leads to confusion when data managers attempt to evaluate and enhance these aspects of data quality.

On the other hand, new and specialized metrics have emerged with the expanding use of data analytics in diverse domains, each addressing specific aspects of data quality. One such metric is Compliance, which evaluates the degree to which data aligns with predetermined regulations, standards, or industry guidelines, ensuring the legal and ethical integrity of the data. Another noteworthy metric is Relevance, which assesses the suitability and importance of data about specific analyses, ensuring that only the most relevant information is employed for the task at hand. There is also the metric Understandability, which assesses how easily users can understand data, considering elements such as data labeling, clear documentation, and the general ease of use for the end user. Plausibility is another metric that examines data's logical consistency and rationality, ensuring it aligns with real-world scenarios. These emerging metrics reflect the dynamic nature of data analytics, adapting to the ever-growing complexities of data quality requirements across different sectors. Various techniques have been employed to define such metrics for data quality assessment. Some metrics are established based on standards like ISO/IEC 15939 and ISO/IEC 25000, relying on qualitative techniques to ensure data quality. Others use mathematical methodologies such as normalizing probability and distance entropy, disturbed entropy, and formulas for rule-based measurement of data quality dimensions like completeness, consistency, accuracy, and uniqueness. Sampling and profiling are widely adopted algorithms in defining metrics. Moreover, some metrics, such as accuracy, were determined based on machine learning approaches such as recurrent neural networks (RNNs), long short-term memory (LSTM) techniques, or word embeddings.

Concerning big data quality assessment approaches, as mentioned earlier, there is a pressing need for methodologies tailored to the unique requirements and challenges related to big data. It's important to acknowledge that a significant portion of the existing research in this field has attempted to adapt and apply conventional data quality metrics to big data. Although this strategy may seem effective, it also introduces crucial factors to consider: Firstly, the applicability of the existing metrics to big data environments must be evaluated. While many traditional quality metrics are applicable, the scale, diversity, and speed at which big data operates introduce complexities that require adjustments or the development of new metrics. Big data often encompasses diverse data sources and types, including unstructured and semi-structured data, streaming data, and data from various domains. Assessing the quality of such heterogeneous data necessitates a more flexible and context-aware approach. Secondly, the performance of data quality assessment tools becomes a critical concern in big data. Traditional tools designed for smaller datasets may struggle to cope with the scale and complexity of big data, leading to potential performance issues. Lastly, the accuracy of measurements is of utmost importance. As big data often involves real-time or near-real-time analysis, any inaccuracies or delays in assessing data quality can have significant consequences. The speed at which data is generated and consumed requires methodologies that can provide timely and reliable quality insights. In addition to adapting existing metrics, there have been efforts to introduce novel quality metrics specific to big data. These emerging metrics are specifically designed to address the characteristics and challenges of big data's sheer volume, variety, and velocity. For instance, one such metric is the "spread," which measures the extent to which data sources are widely distributed and interconnected within a big data ecosystem. A low spread indicates that only a few top sites must be identified and wrapped to compile a comprehensive set of data sources. Another innovative metric is the "Value of Tail," which assesses the importance of capturing and including less popular or niche entities in a comprehensive database within the context of big data. If constructing a comprehensive database requires extracting and incorporating data related to less popular entities (relevant to a smaller user group), the value of the tail is high. "Connectivity" is another quality metric that measures the ease with which data sources can be discovered and accessed through bootstrapping-based Web-scale extraction algorithms. High connectivity indicates a robust network of interconnected data sources, simplifying data extraction and ensuring the accessibility of a wide range of relevant data. These new metrics, among a few others, demonstrate the evolving nature of data quality assessment in the context of big data, emphasizing the need for tailored approaches to address the unique dimensions of this data paradigm. After reviewing the current state of the art regarding big data quality assessment, three prominent gaps emerge: First and foremost. At the same time, the literature offers numerous dimensions of data quality. Only a small portion of these dimensions can be quantified and assessed through existing metrics. This large gap between the defined and measured dimensions highlights the pressing need to develop new metrics to effectively measure the different dimensions of data quality, especially for big data. Secondly, not all data elements carry the same weight in evaluating data quality, and some may substantially influence the overall data quality. Ignoring this crucial factor may drastically impact the accuracy of the data quality assessment. Lastly, there is a need for measurable data quality aspects that consolidate various dimensions and provide a holistic view and global insights about data quality, enabling a more thorough understanding and management of data quality on a global scale. In response to the challenges identified, this chapter endeavors to broaden the scope of measured dimensions. We introduce four new data quality metrics: Accessibility, Integrity, Security, and Ease of manipulation. As a result, we present an inclusive Big Data Quality Assessment Framework comprising 12 metrics, enumerated as follows: Timeliness, Completeness, Volatility, Conformity, Uniqueness, Ease of manipulation, Consistency, Readability, Relevancy, Security, Integrity, and Accessibility. To provide a comprehensive perspective on data quality, we have organized these metrics into five overarching quality aspects: Availability, Reliability, Validity, Usability, and Pertinence. This categorization enhances the clarity of data quality assessment. Additionally, to improve the precision of quality evaluations, we apply data weights to data fields, quality metrics, and quality aspects. In the subsequent section, we detail the data quality aspects under consideration in this research, expound upon and provide measurements for the 12-quality metrics, and elucidate the concept of weighted data quality, highlighting its application across various levels.

# 4. Big Data Quality Assessment Framework using Weighted Metrics

## 4.1. Big Data Quality Aspects

To comprehensively understand data quality, we propose classifying quality metrics into distinct categories known as "data quality aspects." The data quality aspects group metrics that share similar traits, offering a higher level of abstraction and valuable insights into data quality, which could support data managers during data analysis. Quality metrics can be classified and grouped through various criteria, such as their nature, meaning, or the study's specific context. In our research, we have defined 12 data quality metrics, which we grouped into five key quality aspects, as shown in Figure 7. The five quality aspects are Reliability, Pertinence, Validity, Usability, and Availability defined as follows:

**Reliability** pertains to the accuracy and trustworthiness of the data.

**Availability** refers to the data's accessibility and the ability to share it while ensuring appropriate data security.

**Usability:** focuses on the relevance and user-friendliness of the data.

**Validity:** ensures data adheres to specified formats and complies with the defined business rules.

**Pertinence:** relates to what renders data pertinent and suitable for its intended utilization context.

Although various classifications have been proposed in the literature [65] [27], none of these studies have measured the quality aspects. Typically, metrics classification highlights specific properties or defines general aspects of data quality rather than for measurement or assessment purposes. Our approach involves assessing the quality aspects using a weighted average of the metrics relevant to each element. The averaging factors are adjusted based on the relevance of each metric. In the following section, we delve into the specific quality metrics associated with the five quality aspects and propose suitable quality measures for each of these metrics.

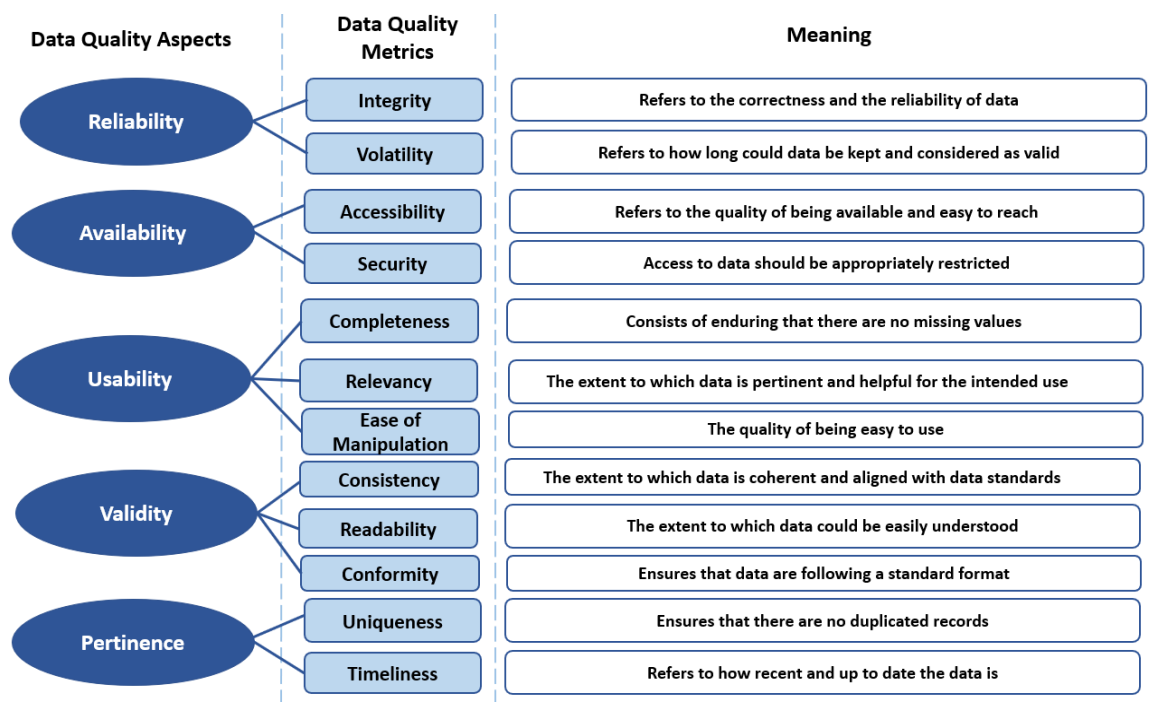

*Figure 7: Big Data Quality Aspects and Metrics*

## 4.2. Big Data Quality Metrics

To contribute to the ongoing discussion about big data quality assessment, we outline in this section 12 distinct big data quality metrics designed to be applicable in the context of big data. The defined metrics are measures for the 12 metrics: Completeness, Timeliness, Volatility, Uniqueness, Conformity, Consistency, Ease of manipulation, Relevance, Readability, and Security. Moreover, we explore the influence of big data characteristics on these metrics, highlighting their specific impact on each metric.

### 4.2.1. Completeness

In big data environments, the raw data often suffer from incompleteness and lack essential contextual information. Consequently, data completeness emerges as a crucial criterion during data quality evaluation. Data completeness revolves around assessing the extent to which the data are adequately filled and satisfy the necessary information requirements. Traditionally, data completeness measurement primarily revolves around the quantification of missing values. However, it is also pertinent to consider completeness at the attribute level, mainly when specific obligatory fields are present. Furthermore, completeness assessment can be conducted horizontally, at the row level, which proves more relevant for reference data like countries and currencies. In this study, our definition of completeness centers on the ratio of non-missing values, encapsulating the essence of this vital data quality metric.

$$Completeness(\%) = \frac{Number\ of\ non\ empty\ values}{Total\ values} * 100$$

### 4.2.2. Uniqueness

Large-scale datasets often suffer from redundancy due to their collection from various sources. As a result, identical information can be replicated multiple times. Thus, ensuring data uniqueness is essential before delving into the analytical phase, as duplicated records can impact the accuracy and integrity of the analytical results. The ratio of non-duplicated values essentially quantifies data uniqueness.

$$Uniqueness(\%) = \frac{Number\ of\ unique\ rows}{Total\ rows} * 100$$

### 4.2.3. Consistency

Inconsistency within a dataset refers to the absence of coherence or concordance among multiple records about the same real-world entity. Consequently, a consistency anomaly arises when records referring to identical entities exhibit conflicting information. A classic illustration of such an anomaly involves two records with the same name and ID number, yet their addresses differ. These inconsistencies may arise due to various factors, including errors during data entry or data integration from multiple sources. We define the consistency metric as the number of identical values among duplicated records divided by the total number of repeated values.

$$Consistency(\%) = \frac{Number\ of\ identical\ values}{Total\ duplicated\ values} * 100$$

### 4.2.4. Conformity

In big data environments, the diverse origins of collected data often result in inherent inconsistency and the potential inclusion of anomalies. As a consequence, the process of transforming raw data into a structured and uniform format is a challenging task when dealing with big data. Achieving data consistency, which involves presenting information in a coherent manner consistent with established data schemas and standards, poses a significant challenge. In this chapter, our focus lies on quantifying consistency by using pre-defined data types. Therefore, we define data consistency as the proportion of values that adhere to the expected data type about the total number of values. This provides a meaningful metric to evaluate the alignment and reliability of the structured data within the big data ecosystem.

$$Conformity(\%) = \frac{Number\ of\ values\ with\ consistent\ format}{Total\ values} * 100$$

### 4.2.5. Timeliness

Variability is a prominent attribute of big data, characterized by the rapid and frequent information updates as new data becomes accessible. In this dynamic environment, maintaining data currency is of utmost significance, as outdated data can potentially lead to biased results of data analysis. The Timeliness metric, also known as currency or freshness, is a vital tool for evaluating the relevance and accuracy of data within such a context. This metric considers the recency of data and its alignment with real-world values. The measurement of Timeliness revolves around assessing the time elapsed since the data was last modified and comparing it to the current date.

$$Timeliness(\%) = \frac{Current\ Date - Last\ Modification\ Date}{Current\ Date - Creation\ Date} * 100$$

### 4.2.6. Volatility

Initially introduced as a characteristic of big data, volatility has evolved into a significant data quality metric primarily concerned with the duration of data's utility and relevance. Unlike timeliness, which is associated with preprocessing tasks, volatility is closely tied to the inherent nature of the data itself. Specifically, it pertains to the challenges posed by processing big data that is continuously updated and subject to frequent changes. As a data quality metric, volatility determines when data can be retained and deemed valid for analysis and decision-making purposes. The definition of volatility as a quality metric is based on calculating the time delay between the data storage date and its last modification.

$$Volatility(\%) = \frac{CreationDate - Modification\ Date}{Current\ Date - Creation\ Date} * 100$$

### 4.2.7. Readability

Data validity extends beyond the scope of data format and encompasses data semantics, ensuring that the information presented is meaningful and accurate. Raw data, particularly in extensive human data entries in social media datasets, may be prone to misspelled or nonsensical words. Moreover, semantic challenges may arise in digital data and non-digital forms like images and audio. Data readability is pivotal in addressing these issues, focusing on effectively processing and extracting relevant information from the data. In this context, readability can be defined as the ratio of processed values free from misspellings, capturing the degree to which the data maintains its meaningful representation.

$$Readability(\%) = \frac{Number\ of\ non\ misspelled\ values}{Total\ values} * 100$$

### 4.2.8. Ease of Manipulation

The preprocessing phase of big data occurs before its use, involving various crucial processes like cleaning, data integration, and reduction. However, implementing these transformations on a vast volume of data can be costly in terms of financial resources, workforce, and time. Hence, we introduce the concept of "ease of manipulation," which evaluates the simplicity with which data can be readily used, requiring minimal effort during the preparation phase. To measure this metric, we assess the invested effort in preparing the data for manipulation by comparing the data in their original schemas to their state after preprocessing. Accordingly, ease of manipulation is defined as the ratio of differences observed between the raw and preprocessed data concerning the total data.

$$Ease(\%) = \frac{Number\ of\ differences\ between\ original\ and\ cleaned\ table}{Total\ data} * 100$$

### 4.2.9. Relevancy

Data relevancy and usability are critical dimensions of data quality in big data environments, recognizing that not all captured data hold equal importance for the intended use. Relevancy, in this context, pertains to the degree of alignment between the information contained within the data and the specific information requirements. Assessing relevancy is inherently contextual and relies on the intended purpose of the data. Our framework adopts the definition proposed in [58], where the authors associate data relevancy with data access frequency. Accordingly, the most frequently accessed data is deemed the most relevant, signifying its higher significance and usability for analysis and decision-making purposes.

$$Relevancy(Field\ F) = \frac{Number\ of\ access\ to\ F}{Total\ access\ to\ the\ table\ that\ includes\ F} * 100$$

### 4.2.10. Security

Data Security refers to the appropriate restrictions placed on data access to protect it from unauthorized access or breaches. In light of the rising instances of large-scale privacy breaches and security attacks, prioritizing data confidentiality and security has become crucial. Measuring data security involves a comprehensive and specific examination. While a detailed assessment is required, we suggest guiding questions can help in assigning a score to the data security level:

**Security Policy Compliance (20%):** This aspect evaluates whether an established security policy governs data usage. It assesses the extent to which data usage is aligned with the security policies.

**Data Transfer Protocols (20%):** This aspect examines using security protocols during data transfer. It focuses on the security measures in place to safeguard data during transmission.

**Threat Detection Measures (20%):** This aspect assesses the implementation of measures for threat detection. It looks at the strategies and technologies in place to identify and respond to potential security threats and breaches.

**Data Encryption (20%):** This aspect evaluates whether data is appropriately encrypted. It considers the methods and strength of encryption techniques to protect data from unauthorized access.

**Security Documentation (20%):** This aspect examines the presence of security documentation accompanying the data. It assesses whether comprehensive documentation is available to provide insights into the security measures, access controls, and data handling procedures in place.

### 4.2.11. Accessibility

Data accessibility is a crucial metric that guarantees data availability and ease of retrieval. It holds significant importance as data becomes useless if it remains inaccessible. In today's interconnected world, data should be easily accessible and efficiently retrievable, even beyond the local repository, when distributed for external use. The definition of accessibility in this context revolves around quantifying the ratio of accessible values, indicating the proportion of data that can be obtained and used.

$$Accessibility(\%) = \frac{Number\ of\ Accessible\ values}{Total\ values} * 100$$

### 4.2.12. Integrity

Data integrity is a fundamental aspect of data quality, referring to the accuracy and trustworthiness of data throughout its lifecycle. In the dynamic realm of big data, data undergo various transformations and processes before being used, making it imperative to ensure the data's integrity and validity during these operations. Ensuring data values remain unaltered and dependable throughout the data processing journey is critical for reliable analysis and decision-making. A comparative study is conducted between data values before and after processing to quantify data integrity. Data integrity is the ratio of differences observed between the original and processed data values about the total number of values.

$$Integrity(\%) = \frac{Number\ of\ differences\ between\ original\ and\ processed\ values}{Total\ values} * 100$$

### 4.3. Data weights

Assessing the quality of big data requires incorporating data weights, as the relevance of information varies when viewed from a business perspective. In most organizations, specific data hold greater importance than others, thus influencing data quality evaluations. Table 5 presents an illustrative dataset schema for customers, featuring fields such as First Name, Last Name, Age, Address, Email, Phone Number, Country, and City, alongside their completeness scores.

*Table 5:Customers' dataset fields and their completeness scores.*

| First Name | Last Name | Age | Address | Email | Phone Number | City | Country |
|---|---|---|---|---|---|---|---|
| 90% | 90% | 80% | 40% | 30% | 20% | 65% | 70% |

Data completeness is considered to be the percentage of non-lacking values. Thus, the completeness score is 60.62% (1).

$$Completeness = \frac{90 + 90 + 80 + 40 + 30 + 20 + 65 + 70}{8} = 60.62\%$$

From a business perspective, if we examine the mentioned dataset aimed at a company's marketing campaign to gather data about potential customers, the company will prioritize contact data, specifically email and phone numbers, as these facilitate customer communication and product/service promotion. To ensure more precise data quality measurement, giving these fields a higher influence on the completeness metric is essential. Therefore, we propose adopting weighted metrics by allocating weights to data fields based on the following approach:

**Step 1:** Arrange the fields in order of relevance and their expected impact on the quality score. In the previous instance, the prioritized order for fields is as follows:

1- Email.

2- Phone Number.

3- Address.

4- City and Country.

5- First Name, Last Name, and Age.

**Step 2:** Assign a factor of 1 to the less critical fields. In the above example, the fields First Name, Last Name, and Age are assigned a factor of 1. Subsequently, allocate a factor ranging from 2 to 10 to each field based on its intended impact compared to the less significant field(s). To assist with this, we recommend consulting Table 6 as a guide for assigning factors. The factors assigned to the fields in the example above are displayed in Table 7.

*Table 6: factor's impact degree.*

| Factor Range | 1–2 | 3–4 | 5 | 6–7 | 8–10 |
|---|---|---|---|---|---|
| Impact Degree | Very Low | Low | Moderate | Significant | Very High |

*Table 6: Field's factors.*

| First Name | Last Name | Age | Address | Email | Phone Number | City | Country |
|---|---|---|---|---|---|---|---|
| 1 | 1 | 1 | 3 | 6 | 4 | 2 | 2 |

**Step 3:** Determine each field's weight by calculating the assigned factor's ratio to the total factors. In this illustration, the total sum of factors is 20.

As a result, applying the designated weights to the fields, as indicated in Table 8, leads to a completeness score of 45.5%.

$$Completeness = \frac{90 \times 0.05 + 90 \times 0.05 + 80 \times 0.05 + 40 \times 0.15 + 30 \times 0.3 + 20 \times 0.2 + 65 \times 0.1 + 70 \times 0.1}{8} = 45.5\%$$

*Table 7: Field's weights.*

| First Name | Last Name | Age | Address | Email | Phone Number | City | Country |
|---|---|---|---|---|---|---|---|
| 1/20 = 0.05 | 1/20 = 0.05 | 1/20 = 0.05 | 3/20 = 0.15 | 6/20 = 0.3 | 4/20 = 0.2 | 2/20 = 0.1 | 2/20 = 0.1 |

Thus, not considering data weights could make a dataset incomplete even when the missing details are irrelevant. However, considering data weights allows a dataset with empty values to be considered comprehensive if it meets the business needs. Therefore, data weights should be meticulously considered when measuring data quality, especially in big data systems where not all information holds equal relevance. However, despite its importance, most studies have yet to adopt data weights in their assessment approaches, raising concerns about the correctness and precision of their measurements.

It is important to note that this approach extends beyond data fields and can be applied to various levels of data units. For instance, data rows may also have varying relevance. From a business standpoint, specific customers may hold more excellent value in sales and profitability, making complete information on these valuable customers more crucial. In such cases, the database's completeness score should reflect the importance of having complete data for targeted customers in specific regions. Moreover, this approach can be applied to other data units, such as tables and data quality dimensions. To achieve more precise measurements, we propose in this chapter a weighted big data quality score, where data weights are applied at three levels:

**Data Fields:** Each data attribute has varying relevance and should have different weights when measuring a quality metric.

**Quality Dimensions:** Different quality metrics have varying significance and should be assigned different weights when measuring quality.

**Quality Aspects:** Various quality aspects have different levels of relevance and should be weighted differently when measuring the overall quality score.

In the subsequent section, we implement the previously defined quality metrics and aspects in a case study involving Twitter Stream data while considering data weights at different levels. Additionally, we conduct a comparative study of existing assessment approaches and draw conclusions based on our findings.

# 5. Implementation

## 5.1. Dataset Description

The data quality metrics were assessed on an extensive dataset of Twitter posts from the Twitter Stream related to COVID-19 chatter [119]. The selection of this particular case study stemmed from technical considerations and its contextual significance. On the technical side, the choice was driven by factors such as the dataset's scale, its origin, the nature of its attributes, and its inherent simplicity. The case study was also chosen due to its pertinence around social media content dealing with a pertinent subject. The dataset encompasses 283 million tweets and supplementary details surrounding language, country of origin, and attributes like creation date and time. These tweets were gathered from January 27th to March 27$^{th,}$ 2020, averaging a daily count exceeding four million. In the context of our specific case study, the data collected were initially well-structured and exhibited a relatively high level of quality. However, to underscore the efficacy of our quality assessment approach, some modifications were performed to disorganize the dataset intentionally, which affected all the quality metrics under evaluation.

## 5.2. Tools and Libraries

The big data quality assessment framework was implemented using Apache Spark, tailored to manage extensive datasets in big data environments. Quality metrics were implemented in Python using Pyspark libraries, including the Great Expectation Package [122], Numpy, Spell Checker, Matplotlib, and Pandas. Before initiating any assessment, the dataset underwent loading and preprocessing. The data schema was adjusted to align with the measurement requisites. As such, specific fields, like creation date and time, were formatted. Moreover, supplementary fields were introduced into the dataset, such as the "last modification DateTime" attribute, enabling measurement of Timeliness and Volatility metrics.

## 5.3. Results

As the goal of the implementation was to test the suggested measurements, arbitrary values were assigned to the additional fields while maintaining the case study's consistency and logic. Additionally, the number of accesses to the data was added to the dataset to calculate the Relevance metric. Then, weights were assigned to the data fields, the metrics, and the quality aspects. The performed experiments aimed to accomplish the following goals:

- Measuring the following quality metrics: Timeliness, Completeness, Volatility, Conformity, Consistency, Uniqueness, Ease of manipulation, Security, Readability, Relevancy, Accessibility, and Integrity, while considering the attribute weights presented in Table 9.

*Table 8: Field's Weights*

| Tweet | Language | Country | Creation Date Time | Last Modification Date Time |
|---|---|---|---|---|
| 0.4 | 0.15 | 0.15 | 0.15 | 0.15 |

- Measuring the following quality aspects: Availability, Reliability, Usability, Validity, and Pertinence while considering the metric weights presented in Table 10.

*Table 9: Metrics Weights*

| Reliability | Integrity | 0.7 |
|---|---|---|
| | Volatility | 0.3 |
| Availability | Security | 0.8 |
| | Accessibility | 0.2 |
| Pertinence | Timeliness | 0.7 |
| | Uniqueness | 0.3 |
| Validity | Consistency | 0.4 |
| | Conformity | 0.4 |
| | Readability | 0.2 |
| Usability | Completeness | 0.5 |
| | Relevancy | 0.3 |
| | Ease of Manipulation | 0.2 |

- Measuring a global weighted data quality score while considering the aspect weights in Table 11.

Table 10: Aspects Weights

| Reliability | Availability | Pertinence | Validity | Usability |
|---|---|---|---|---|
| 0.3 | 0.1 | 0.1 | 0.3 | 0.2 |

We assessed the quality metrics associated with each field following the abovementioned weighting scheme. Based on the results obtained, an evaluation of the data quality aspects was carried out. Finally, the overall big data quality score was measured.

It's important to note that not all metrics were subject to field-specific weighting. Indeed, specific metrics could not be applied to every field and, as a result, were not computed for all data attributes. Metrics such as Uniqueness, relevance, Security, and Accessibility fell into this category.

Data preprocessing was imperative to assess the ease of manipulation metric. In addition to the earlier adjustments, we enhanced data consistency and conformity by modifying data types and formats to align with the expected data model. Subsequently, we compared the original and preprocessed datasets, assessing the degree of similarity between them. Figures 8 and 9 show the obtained scores after considering the big data quality metrics and aspects.

Following the measurement of data quality metrics and the evaluation of data quality aspects, the overall big data quality score was then deducted to get this final score:

Quality Score (%) = 58.35%

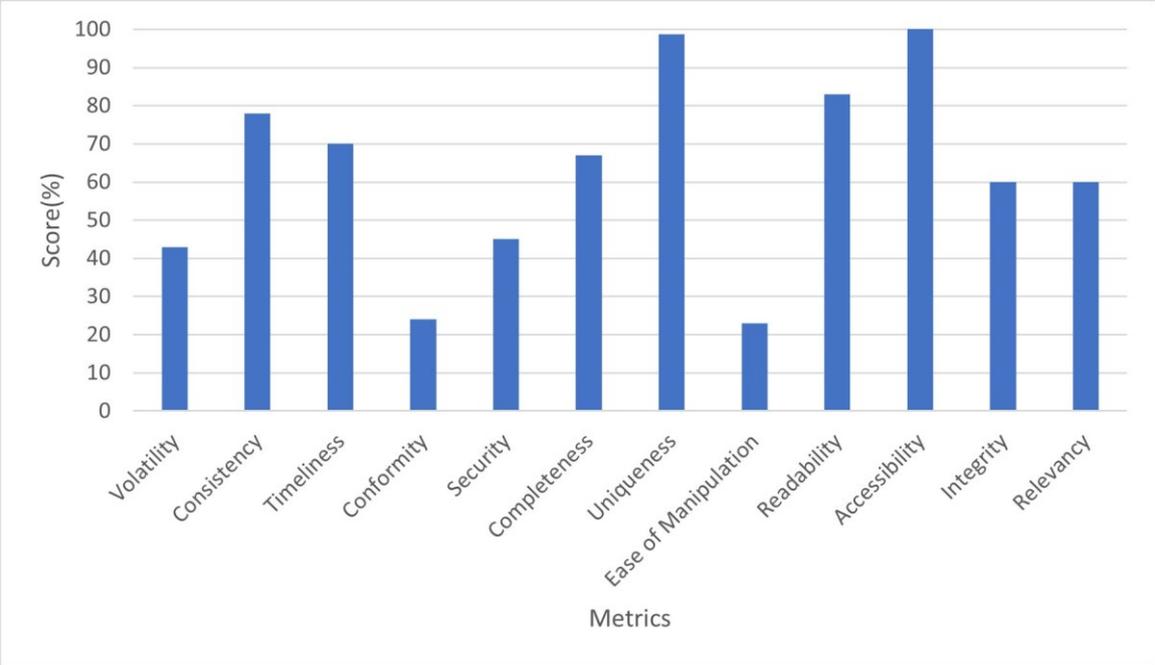

Figure 8: Data Quality Metric Scores.

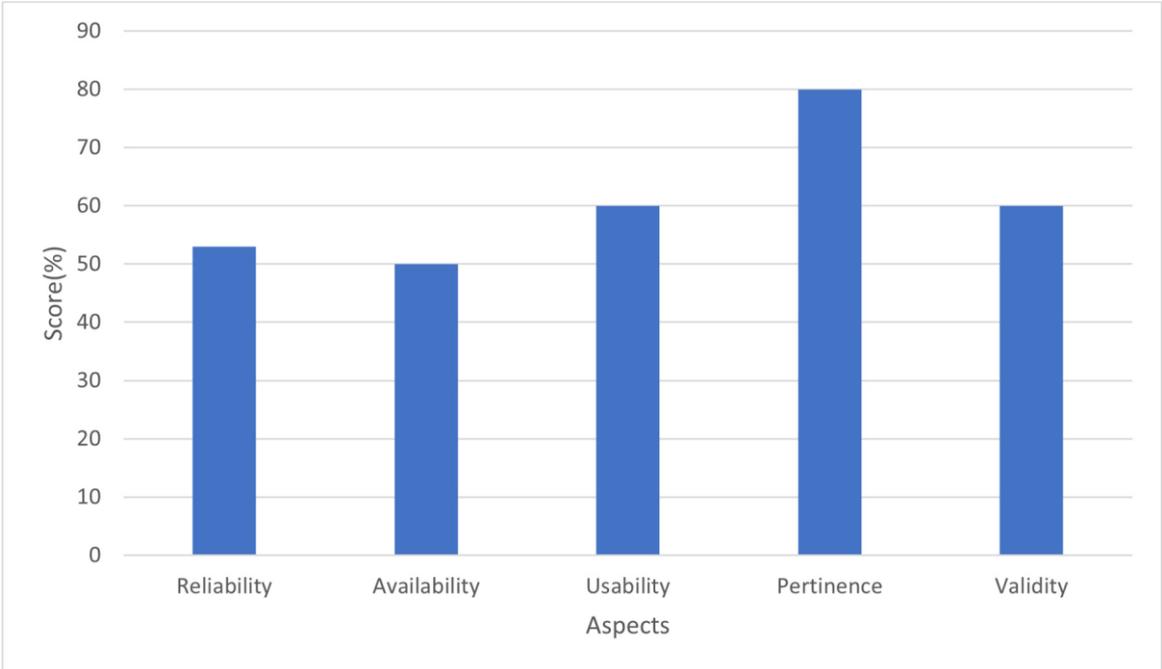

Figure 9: Data Quality Aspects Scores

## 5.4. Comparative Study

In this section, we undertake a comparative analysis to evaluate the proposed framework of existing data quality assessment methodologies. We focus on various aspects, including the framework model, incorporating pre-existing metrics, introducing novel metrics, and considering big data scope and precision.

A recent study [120] into data quality assessment for big data introduced a quality evaluation model that enables the assignment of a score to existing quality assessment frameworks designed for big data. The scoring mechanism is established as a formula (S), utilizing seven specific evaluation criteria outlined below. The values in this formula are binary, with 0 or 1. Notably, the highest score achieved in the survey was 7.5, as reported in the paper [67]. Therefore, we evaluated both our framework and existing frameworks using this model.

$$S = D + 2 \times M + 2 \times Mt + 1.5 \times F + 1.5 \times S + A + P \text{ (S)}$$

1) **Criterion D (Providing a Data Quality explanation):** This assesses the clarity and articulation of the concept of data quality within the study.
2) **Criterion M (Proposing a Quality structure):** It evaluates the introduction of a structured model or framework for characterizing data quality.
3) **Criterion Mt (Offering evaluation indicators):** This examines the presentation of specific indicators or metrics used to gauge data quality.
4) **Criterion F (Proposing a methodological approach):** It assesses the inclusion of a systematic approach or framework for data quality assessment.
5) **Criterion S (Developing a simulation or prototype):** This considers the creation of a practical simulation or prototype related to data quality.
6) **Criterion A (Presenting a Current State of Knowledge):** It looks at the coverage and analysis of the existing body of knowledge in data quality.
7) **Criterion P (Conducting a survey):** This examines the involvement of a survey or data collection process to gather insights or opinions regarding data quality.

We presented a clear data quality definition, thereby D = 1.

Our framework introduced a novel quality model incorporating 12 metrics and 5 distinct quality aspects, thus M = 1.

A total of 12 evaluation metrics were defined within our framework, equating to Mt = 1.

We did not employ polling mechanisms, resulting in P = 0.

The field's current state was provided, aligning with A = 1.

A simulation was conducted, yielding S = 1.

Our framework proposed an assessment framework that delivers a weighted data quality score, signifying F = 1.

Table 12 offers a classification and comparison of existing data quality frameworks and the corresponding scores assigned to each, incorporating preexisting metrics, introducing novel metrics, and considering big data scope and precision.

*Table 11 : Comparative Table of Data Quality Assessment Frameworks*

| | Ref. | [54] | [55] | [121] | [122] | [58] | [123] | [52] | [66] | [67] | [124] | [42] | Our Framework |
|---|---|---|---|---|---|---|---|---|---|---|---|---|---|
| Score | D | 1 | 1 | 1 | 1 | 1 | 1 | 1 | 1 | 1 | 1 | 1 | 1 |
| | M | 1 | 1 | 1 | 1 | 1 | 1 | 1 | 1 | 1 | 0 | 0 | 1 |
| | Mt | 1 | 1 | 0 | 1 | 1 | 0 | 1 | 1 | 1 | 1 | 1 | 1 |
| | F | 1 | 0 | 0 | 0 | 0 | 0 | 0 | 0 | 1 | 1 | 1 | 1 |
| | S | 0 | 1 | 1 | 0 | 0 | 1 | 1 | 1 | 0 | 0 | 1 | 1 |
| | A | 1 | 1 | 1 | 1 | 1 | 1 | 1 | 1 | 1 | 1 | 1 | 1 |
| | P | 0 | 0 | 0 | 0 | 0 | 0 | 0 | 0 | 0 | 0 | 0 | 0 |
| | Score | 7.5 | 7.5 | 5.5 | 6 | 6 | 5.5 | 7.5 | 7.5 | 7.5 | 5.5 | 7 | 9 |
| Metrics | Completeness | * | * | | * | * | | * | * | * | * | | * |
| | Timeliness | * | * | | * | * | | | * | | | | * |
| | Volatility | | | | | | | | * | | | | * |
| | Uniqueness | | * | * | * | | * | | | | * | | * |
| | Accuracy | * | | | * | * | | | * | * | | * | |
| | Conformity | | * | | * | | | | | | | | * |
| | Consistency | * | * | | * | * | | | * | * | * | | * |
| | Correctness | * | | | * | | | | | | | | |
| | Relevancy | | | | | * | | * | | | | | * |
| | Readability | | | | | | | | * | | | | * |
| | Spread | | | | * | | | | | | | | |
| New Metrics | Introducing New Metrics? | yes | no | no | yes | yes | no | no | no | no | no | no | yes |
| | Number of New Metrics | 1 | | | 2 | 1 | | | | | | | 4 |
| Scope and Precision | Big Data | * | | | * | | | * | * | | | * | * |
| | Weighted Quality | | | | | * | | | | | | | * |
| | Number of considered | 5 | 5 | 1 | 8 | 5 | 1 | 3 | 5 | 3 | 3 | 1 | 12 |

## 5.5. Discussion

The obtained overall big data quality score shows the low quality of big data and sheds light on the importance of cleaning and preprocessing big data before using them. Considering big data characteristics, getting deficient scores for the Volatility, Conformity, and Ease of manipulation metrics is normal. As discussed previously, the data quality scores, as presented in Figure 8, show a high dependency between the quality metrics. Hence, we obtained a low-security score for increased accessibility and a balanced score between completeness and relevancy. Thus, the suggested framework allows the accurate assessment of big data quality and the evaluation of how the quality metrics impact each other. Based on the results presented in the previous tables, we notice that, even if many efforts have been made to assess data quality, there is still a significant lack of work defining new metrics. Indeed, standard data quality metrics, such as Completeness, Consistency, and Accuracy, are the most addressed metrics in the literature.

On the other hand, new metrics, such as Volatility, Spread, and Readability, are less considered in the literature. Moreover, we notice that there is confusion in the naming of metrics. Some metrics refer to the same meaning but are named differently in each study, such as Freshness, which refers to Timeliness, and Copying, which relates to Conformity. Therefore, while setting up the comparative table, we considered the meanings of the defined metrics instead of their names. Table 12 also shows a lack of consideration of data weights by the existing data quality assessment approaches despite their high importance and impact on the accuracy of the measures. Thus, the results show that the suggested methodology outperforms the current data quality assessment frameworks with a score of 9/10 while including 12 defined metrics, considering data weights, and addressing big data characteristics.

On the other hand, with the 12 defined metrics, there is still a large gap between the specified and measured dimensions, highlighting the need to measure and implement new quality metrics. In addition, some metrics are contextual and, therefore, can be measured differently depending on the data context, such as relevance and Accessibility. For these metrics, generic and non-context-aware approaches need to be implemented.

In data analytics, while data quality undoubtedly plays a pivotal role, it's crucial to recognize another indispensable component: metadata. Often overlooked but equally vital, metadata serves as the contextual backbone that empowers accurate interpretation and analysis of data. Indeed, data quality could not be effectively enhanced without considering metadata quality. To comprehensively address data quality, we explore metadata quality in the following section and propose an initiative approach to improve metadata quality through the metadata value chain for big data.

## Conclusion

In this chapter, we have suggested a comprehensive big data quality assessment framework based on 12 metrics. Moreover, we have defined and measured five data quality aspects to provide a macro-view of data quality. Furthermore, we have weighed data at three levels for accurate assessment: data fields, data quality dimensions, and data quality aspects. The implementation and evaluation of the framework have shown that the suggested framework outperforms the existing approaches in terms of the number of addressed metrics, scope, and comprehensiveness. With a comprehensive big data quality assessment framework, we have covered the first primary axis of data quality among the three axes we identified in the first chapter. If data quality assessment is essential to gain insights about the quality level of data, it constitutes only a first step towards enhancing data quality. In the next chapter, we will tackle the second axis of data quality management: data quality anomaly detection. Anomalies can significantly impact data quality and indicate underlying data collection or processing issues. Therefore, detecting and addressing these anomalies is crucial for ensuring that the data utilized for decision-making is accurate, reliable, and valuable.

# Chapter 5: Big Data Quality Anomaly Detection Framework

## 1. Introduction

Suppose data quality assessment provides valuable insights about the dataset's quality level. In that case, it does not allow locating quality anomalies within the dataset to address them. Thus, detecting quality anomalies is a central component of effectively managing data quality. While detecting quality anomalies often involves elementary and conventional methodologies, identifying data quality anomalies, particularly within big data, is more challenging and requires advanced and intelligent techniques. Among the spectrum of quality anomalies, outlier values hold considerable importance and can significantly influence the precision of Big Data analysis. Outliers refer to data points that deviate from the typical distribution of data values. As highlighted in Chapter 3, many Big Data anomaly detection approaches have been suggested in the literature to address issues related to outlier values. These approaches have substantially contributed to the advancement of anomaly detection across diverse domains, such as finance, healthcare, transportation, and security. However, it is imperative to recognize that while outlier values represent a pivotal facet of data quality, they constitute merely one among several other quality dimensions that impact the integrity and precision of Big Data analysis.

Consequently, despite the significant progress achieved in anomaly detection, the existing methodologies remain somewhat inadequate in comprehensively tackling the spectrum of quality anomalies. While most existing methods are primarily oriented toward identifying deviations from expected patterns, they overlook the other anomalies linked to data quality. Detecting these quality-related anomalies is challenging, as such anomalies do not exhibit distinctive patterns typically associated with irregular behavior. Consequently, a problematic demand arises for developing and implementing more sophisticated and pragmatic approaches to identifying and resolving all aspects of data quality anomalies, particularly within Big Data. While exploring the process of detecting different data quality anomalies, we found that identifying uniqueness and consistency anomalies requires a preliminary entity resolution phase. Thus, this raised requirement motivates us to develop a novel big data entity resolution framework as a foundational step for comprehensive anomaly detection. To address the concerns raised about the necessity of encompassing all aspects of data quality anomalies, we present a comprehensive approach for Big Data Quality Anomaly Detection, featuring the following significant contributions:

1) We introduce a novel Big Data Entity Resolution Framework founded on Online Continuous Learning as an initial step toward anomaly detection. This framework is further extended to encompass an end-to-end big data deduplication framework.
2) We have established a new end-to-end Data Quality Anomaly Detection Framework, which employs predictive techniques to proactively identify potential generic data quality anomalies in big data, enhancing data accuracy and reliability.
3) Our approach detects anomalies in six key quality dimensions: Consistency, Accuracy, Conformity, Completeness, Readability, and Uniqueness.
4) We introduce and calculate a novel metric, the "Quality Anomaly Score," which quantifies the degree of anomaly and low quality associated with the detected irregularities for each quality dimension and the entire dataset.

The rest of this chapter is organized as follows: Section 2 reviews existing research on anomaly detection and data quality for Big Data. In the third section, we present in the first part our suggested entity resolution framework serving as a preliminary stage for anomaly detection. Then, we offer the steps of our proposed Big Data Quality Anomaly Detection Framework in the second part. Section 4 presents the implementation of deduplication and quality anomaly detection frameworks for other datasets. Finally, the obtained results are discussed, and conclusions are made.

## 2. Big Data Anomaly Detection

In the realm of data characterized by low quality, anomaly detection has emerged as a practical approach across diverse domains to address errors and inconsistencies within the data. One of the foremost sectors where anomaly detection plays a crucial role is the banking sector, which is primarily used for fraud detection. Indeed, anomaly detection algorithms analyze transactional data to uncover atypical patterns or deviations from well-established norms, thus enabling the identification of suspicious transactions. Anomaly detection is also used to detect credit card inconsistencies and identify irregularities, discrepancies, or potentially fraudulent activities. Healthcare is another domain where anomaly detection is widely used to detect anomalies in medical records and patient data. It thus provides timely alerts, enabling healthcare providers to take prompt action, make informed clinical decisions, and ensure patients' safety. Anomaly detection is also used in network security as it identifies unusual patterns of network traffic that may signify cyberattacks or intrusions. Another application of anomaly detection is manufacturing, mainly used in predictive maintenance to identify potential anomalies in equipment behavior or sensor readings, such as unusual temperature spikes or vibrations, before they evolve into costly losses. Anomaly detection finds applicability in various domains, including finance, transportation, agriculture, quality control, etc. Across these domains, anomaly detection aims to identify anomalies, like outliers and irregularities within data, which may vary based on the particular scenario and nature of the data. Indeed, the anomaly detection principle involves identifying data points or patterns that deviate significantly from the expected or normal behavior within a dataset. It assumes that most data instances conform to a well-defined way or distribution, while anomalies represent rare occurrences or deviations from this pattern. Anomaly detection typically involves three key steps: data representation, modeling, and identification. Firstly, the data is represented in a suitable format, often involving extracting relevant features or attributes. Then, a model is built to capture the expected behavior of the data, which can vary from simple statistical methods to more complex machine learning algorithms. Finally, anomalies are identified by comparing the new points to the established model. If a data point falls significantly outside the expected range or exhibits patterns that diverge from the norm, it is flagged as an anomaly.

To effectively identify and flag irregularities within datasets, anomaly detection uses various algorithms, such as machine learning classification algorithms, to distinguish between normal and abnormal data points based on their learned patterns. Online learning and distributed learning techniques enable real-time and scalable anomaly detection in dynamic and large-scale data environments. Traditional statistical methods, such as Z-score analysis and Gaussian distribution modeling, are also used for identifying anomalies in numeric data by measuring deviations from established norms. Deep learning and neural networks are also used to capture complex patterns and anomalies in high-dimensional data, such as images, audio, and time series, allowing effective anomaly detection in applications like image recognition, fraud detection, and fault prediction. Textual analysis methods are widely used to detect anomalies in unstructured data, like natural language text, where anomalies might represent fraudulent activities or suspicious behavior. Despite the notable progress made in anomaly detection and its provision of valuable insights across diverse domains, challenges still need to be addressed. A frequently encountered challenge in anomaly detection stems from managing imbalanced datasets, where anomalies are rare compared to typical instances, potentially leading to biased models that overlook anomalies. The dynamic nature of anomalies and the occurrence of concept drift introduce further complexities in the realm of anomaly detection. Indeed, anomalies are not static; they evolve and adapt over time, making it essential for anomaly detection systems to remain flexible and responsive to changing data patterns. Scalability is another significant concern in anomaly detection, particularly in light of the exponential growth in data volumes. Thus, there is a growing need for more efficient big data anomaly detection techniques.

A notable observation regarding anomaly detection is that it is often associated with outlier values, which typically involve identifying extreme or unusual data points that may impact data quality. While outliers undoubtedly play a role in data quality, it's essential to recognize that they constitute a single dimension within the broader spectrum of data quality. Thus, anomaly detection should not be confined solely to identifying outlier values; instead, it should be oriented toward addressing data quality comprehensively. This means detecting anomalies related to various aspects of data quality, such as inconsistencies, errors, missing values, and more. Anomaly detection can significantly enhance data quality by broadening the scope beyond outliers. Another noteworthy observation we raised about the anomaly detection approaches is that they are usually domain-specific. Thus, they tend to focus on "domain anomalies" specific to a given field rather than comprehensively addressing anomalies in data quality.

Consequently, to the best of our knowledge, there is a noteworthy gap in the academic literature concerning the exploration of anomaly detection techniques for a holistic evaluation of data quality, encompassing its diverse dimensions. To overcome the raised points, we present an innovative end-to-end data quality anomaly detection framework capable of intelligently identifying generic data quality anomalies in Big Data, independently of any specific domain, through a predictive model. This framework is designed to detect a broad spectrum of data quality anomalies across six key data-quality dimensions: Consistency, Accuracy, Conformity, Completeness, Readability, and Uniqueness. Additionally, we introduce and calculate a novel metric called the "Quality Anomaly Score," which quantifies the extent of deviation and low quality of the anomalies detected within each quality dimension and the entire dataset.

## 3. Frameworks Conceptualization

## 3.1. An end-to-end big data deduplication framework: A preliminary process toward quality anomaly detection

As mentioned earlier, the datasets need to go through an entity resolution phase to detect quality anomalies related to uniqueness and consistency. We have developed a novel entity resolution framework for big data that we have extended as a standalone deduplication framework. Within this initial section, we introduce the proposed deduplication framework tailored for Big Data, as illustrated in Figure 10, encompassing a sequence of 5 distinct stages. The first phase involves preprocessing, wherein the data is meticulously prepared for deduplication due to the inherent quality challenges related to big data. Subsequently, the second step consists of constructing a training dataset, performed by an automated data labeling process. Then, the dataset undergoes fuzzy matching to detect instances of duplication. The identified duplicates are subsequently subjected to cleansing using the appropriate strategies. Finally, the model is deployed, supported by a real-time continual learning approach, enabling enhanced accuracy during the operational phase. Subsequently, we provide a detailed description of each stage of the framework.

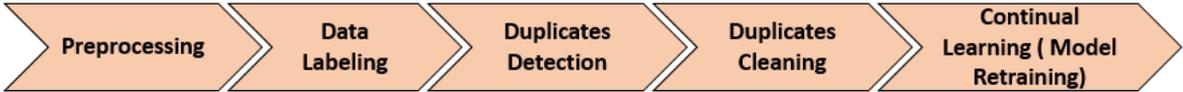

*Figure 10: Big Data Deduplication Framework*

### 3.1.1. Preprocessing

In light of the multifaceted nature of big data, information extracted from big data environments often exhibits unstructured, noisy, and poorly formatted traits. The initial data preprocessing phase becomes a vital requirement before processing the data. During this pivotal phase, raw data transforms into a more orderly and understandable format, rendering it suitable for application with Machine Learning (ML) algorithms. This preliminary step profoundly influences the efficiency and accuracy of the model, underlining its critical importance. Here, we outline the necessary transformations needed to prepare substantial data for deduplication:

**1) Feature Selection and Extraction:** The high dimensionality of Big Data requires the identification of critical data attributes, as not all extracted data within Big Data settings are pertinent to the intended purpose. The aim is to retain only the relevant information, which is achieved by selecting the most informative variables (Feature Selection) or creating new ones (Feature Extraction). This procedure is crucial for data deduplication, as it ensures the preservation of the most significant features upon which the deduplication model depends to identify duplicates.

**2) Encoding:** Encoding entails the conversion of categorical variables into numeric types. Indeed, most ML algorithms are designed to work on numeric input and encounter challenges when dealing with categorical data. Thus, various encoding techniques, such as label encoding, one-shot encoding, and vector indexer, among others, can be applied. Furthermore, encoding allows for ensuring data consistency, a critical factor within the context of data deduplication. Since big data derives from multiple sources, categorical values may be represented differently, creating inconsistency issues that can hinder duplicate detection.

**3) Uppercasing/Lowercasing:** Standardizing text data by converting it to lowercase or uppercase is a crucial transformation. For the sake of simplicity, it is often preferred to convert all data to lowercase, particularly in applications involving Natural Language Processing (NLP). This standardization proves essential in deduplication, as failing to employ this transformation could result in the same word (e.g., "Good" vs. "good") being treated as distinct words in vector space models.

**4)** **Stop words and Symbols removal:** Eliminating irrelevant, frequently occurring words (commonly known as stop words) from text data is crucial. The purpose of removing stop words is to emphasize the information within the data by excluding non-essential terms, as these words do not significantly impact meaning. Additionally, the dataset should be cleansed of special symbols and punctuations, as these elements do not contribute to identifying similarities. Other text elements, such as URLs, HTML tags, and more, may need to be removed depending on the context.

**6)** **Normalization:** Due to the heterogeneity of data sources, variables in the data may exhibit varying scales. Such inconsistencies in scale can bias duplicate detection since records should ideally be compared based on a unified scale. To address this challenge, data normalization is implemented, ensuring that the range of all variables is similar, typically scaled between 0 and 1.

These transformations are regarded as the most critical for preparing data for deduplication. However, based on the specific characteristics of the dataset, additional text cleaning may be necessary, encompassing tasks such as spell correction and stemming.

### 3.1.2. Data Labeling

Once the dataset has undergone preprocessing and reached a state fit for utilization, the subsequent step revolves around constructing a labeled dataset. This labeled dataset serves as the foundation for training the deduplication model. Labeling data is often considered one of the most challenging aspects of AI projects. According to a study [125], labeling data consumes up to 80% of the time allocated for AI projects. While many labeling approaches rely on human annotators, such an approach becomes impractical when dealing with large-scale big data. The impracticality arises not only from the sheer quantity of data but also due to quality-related concerns. To address these challenges, we employ an automated approach for generating labeled data dedicated to entity resolution. This approach is based on Record Linkage techniques and consists of the following steps:

**1)** **Indexing:** Indexing involves generating pairs of candidate records. This step aims not to create all combinations of record pairs within the dataset, as doing so would lead to quadratic time complexity. Instead, the objective is to select likely duplicate pairs that need examination.

Various indexing techniques, including Blocking, Sorted Neighbourhood, and TF-IDF, are available for record linkage. For this framework, we use sorted neighboring for pairs indexing as it is the most suitable indexing technique for big data [126].

**2)** **Comparison and Similarity:** These candidate records are compared once the record pairs are generated. Subsequently, a similarity score is assigned to each pair. The choice of comparison measures depends on the field type (string, numerical value, date, etc.). Many comparison methods, such as Jaro-Winkler, Levenshtein, Cosine, and Jaccard, can be employed for this task. For more precise measurements, weights can be assigned to data fields, as some fields could be more pertinent than others in identifying duplicate records.

**3)** **Labeling:** After calculating similarity scores, pairs are categorized using supervised or unsupervised techniques. Methods such as the Optimal Threshold, Support Vector Machines (SVM), K-Means, and Farthest First, to name a few, are applied to classify pairs into two categories: matches and non-matches. It's important to note that, within this approach, we recommend grouping duplicates into clusters instead of pairs, allowing each cluster to encompass more than two records. Non-duplicate records are excluded, leaving only matching records in the training dataset. In this framework, we use optimal threshold, as the goal is only to keep a small part of the data as a training dataset.

### 3.1.3. Duplicate Detection

We can train the deduplication model with a labeled dataset and subsequently use it to identify matches. In our proposed framework, this approach does not rely on direct text comparisons but instead on deduplication predicates and indexing rules generated by the model post-training. The entity resolution is then performed following these subsequent steps:

**1)** **Training:** The initial phase involves training the model to classify records as duplicates or non-duplicates using the training dataset. Post-training, the model generates indexing rules, referred to as Predicates, for identifying potential matches. These rules are applied during the operational phase to block records that align with the deduced indexing rules.

**2)** **Testing and Parameter Tuning:** This phase evaluates the model's accuracy and fine-tuned performance by determining the most suitable clustering threshold for optimal results. Parameter tuning can be done manually or automatically, employing techniques such as Bayesian Optimization, Random Search, or Tree-structured Parzen Estimator (TPE). The parameter tuning process is relative to the desired balance between precision and recall when detecting or excluding matches, as this trade-off exists inherently.

3) **Serving:** In the final step, the trained and optimized model is used to identify matches and classify records as either "duplicate" or "not duplicate." The model returns clusters of matching records, considering the transitive nature of duplication. Clustering is performed on matching pairs, resulting in records within the same cluster being considered duplicates.

### 3.1.4. Duplicate Correction

Once matching records are consolidated into clusters, the next crucial task involves aggregating data from multiple records into a single representation. The approach to data fusion varies based on the strategic priorities of the data team. Depending on the objectives, the process may focus on data accuracy, wherein the record from the most reliable source is retained. Alternatively, the complete record is preserved if the goal is to gather as much data as possible. Another data fusion strategy involves generating a new record by merging existing records. In this case, conflict resolution mechanisms must be implemented to integrate duplicated data values.

### 3.1.5. Continual Learning (Model Retraining)

Due to the inherent variability of big data, data undergoes constant change, encompassing alterations in a schema, shifts in statistical distribution, changes in data quality, and more. These variations are referred to as data drift. In addition, data can be subject to concept drift, where the statistical properties of the target variable evolve. The model's predictive performance may degrade due to data drift and concept drift. Therefore, adapting the model to these data changes is imperative to maintain model accuracy. This adaptation process aligns with the machine learning strategy known as continuous learning.

Continual Learning is a method that automatically and continuously retrains an ML model with new data, rendering the model adaptive and capable of improving its performance over time. It is particularly relevant in use cases where data evolves, such as recommendation systems that must continuously update with new data as user behavior changes. There are two fundamental approaches to executing Continual Learning:

**Offline Mode (Batch Learning):** The model is periodically retrained with the accumulated new data in this mode.

**Online Mode (Incremental Learning):** The model is continually retrained using a live data stream in this mode.

Online Continual Learning ensures that the model does not deteriorate following data or concept drift, as it continually updates itself with emerging data patterns. This online mode is also a time-efficient solution, as it prevents the need to store and manage substantial batches of accumulated data. Nevertheless, monitoring the input data constantly is essential, as insufficient data can instantly impact performance. The online mode is especially suitable for big data environments and real-time applications. For data deduplication, even when the deduplication model is initially trained with high-quality pairs, features defining duplications may change over time, particularly in scenarios involving human input data. New duplication features may come into play, while previously utilized features may become less reliable. Deduplication models are susceptible to these changes; even minor feature drifts can significantly affect model performance. As a response to this, we advocate for the adoption of an Online Continual Learning approach for deduplication involving the following steps:

1. **Building a Dataset:** Comprising new data and the pairs identified during the operational phase.
2. **Comparison and Similarity Score Calculation:** Comparing and computing similarity scores for the newly constructed dataset.
3. **Selecting Similar Pairs:** Choosing the most similar pairs using a predetermined threshold.
4. **Labeling:** Labeling the selected pairs as duplicates.
5. **Retraining the Model:** Revise the model with the newly labeled pairs.
6. **Performance Evaluation:** Evaluating the model's performance based on the updated data.

This approach operates online, making it efficient regarding memory and time utilization. It proves particularly suitable for large datasets and has demonstrated promising results in enhancing model accuracy. The continual training ensures the model is consistently updated with new pairs, refining the indexing rules to align with the evolving dataset characteristics.

## 3.2. Big Data Quality Anomaly Detection Framework

This section introduces a framework for detecting data quality anomalies using an extended isolation forest model for anomaly identification. The core concept behind anomaly detection revolves around identifying irregularities within the data that deviate significantly from the expected patterns. In this subsequent section, we propose an advanced predictive anomaly detection model that proactively identifies potential data quality issues, focusing on ensuring high-quality Big Data. Notably, this model possesses the capacity for continuous improvement through exposure to additional data, making it especially effective at identifying emerging data quality anomalies or evolving patterns over time. The main idea behind this framework involves reformatting the dataset to reveal hidden anomalies associated with each metric, highlighting their divergence from the dataset's statistical norms. Within this framework, the dataset transforms various patterns, each of which uncovers anomalies related to a quality metric, rendering them discernible in terms of their values and, consequently, easily detectable. The data quality framework consists of four primary phases. The initial preprocessing phase focuses on cleansing and preparing the raw data for the anomaly detection model. Subsequently, the data is structured and molded into multiple patterns, with each pattern highlighting the quality anomalies associated with a specific quality metric in the form of aberrant values, thus facilitating their detection. These resultant patterns are then attributed to the anomaly detection model, which relies on an advanced isolation forest model to pinpoint anomalies linked to each quality metric. Finally, the model computes an anomaly quality score for each anomaly identified, ultimately deriving an anomaly score for each metric and an overall anomaly score for the entire dataset. Figure 11 visually shows the various phases of the Big Data Quality Anomaly Detection Framework.

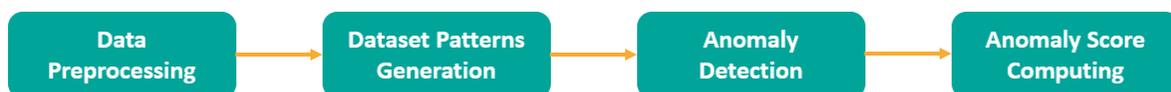

*Figure 11: Big Data Quality Anomaly Detection Framework*

### 3.2.1. Data Preprocessing

Data preprocessing is a pivotal initial phase in preparing data for anomaly detection. Within the realm of Big Data, data preprocessing takes on even greater significance due to the sheer magnitude, speed, and diversity of the data that must be handled. Additionally, large datasets often exhibit unstructured and poorly formatted characteristics, which can introduce bias into models and yield inaccurate results. In this context, preprocessing is a critical enabler for anomaly detection, ensuring that data is accurate, consistent, and primed for analysis. We present the following essential data preprocessing tasks for precise anomaly detection in big datasets:

**Feature selection:** The importance of feature selection in anomaly detection is underscored by its ability to keep only the most relevant features linked to anomalies, thereby enhancing prediction accuracy. This process also allows faster and more effective anomaly detection by reducing data noise and irrelevant features.

**Normalization and Scaling:** Since anomaly detection models primarily rely on the statistical distribution of data values, data scale or range variations have a substantial impact. Normalization and scaling techniques are applied to address this, creating uniform value ranges that enhance the accuracy and accessibility of identifying data points falling outside the expected range.

**Uppercasing/Lowercasing:** Anomaly detection models can exhibit sensitivity to letter cases. Therefore, converting text to either uppercase or lowercase format ensures data uniformity and normalization, contributing to more consistent data representation.

**Dimensionality reduction:** Implementing dimensionality reduction enhances the scalability and performance of the anomaly detection process, effectively reducing data dimensionality.

**Stop Words and Symbol Removal**: Removing stop words and punctuation marks is crucial in preparing data for anomaly detection. This action refines the data, presenting it in a more focused and meaningful pattern by eliminating superfluous elements in textual data. Simultaneously, this process involves the removal of whitespace, "N/A," "NA," "NULL," and "NaN" values, allowing them to be identified and treated as null values.

It's worth noting that the specificities of data cleaning may require additional transformations based on the dataset in question. Furthermore, particular data cleaning tasks are performed in the following step, where data is transformed into new patterns specific to each quality dimension.

### 3.2.2. Dataset Patterns Generation

After preprocessing and cleansing the dataset, the next step involves generating new patterns by transforming the original dataset. Each pattern is meticulously designed to expose deviations associated with specific data quality dimensions and uncover anomalies. In the following, we explain the process of generating the dataset pattern associated with each data quality dimension:

#### 3.2.2.1. Accuracy

Data accuracy refers to the trustworthiness and correctness of information within the dataset. It measures how closely the gathered data aligns with actual values. Inaccurate values, such as ages exceeding 120 or future birth dates, represent outliers deviating significantly from the expected range of these attributes. These outliers are deemed inaccurate, as they vary from the typical value range within the dataset. Their presence can distort statistical analyses and skew results, often arising due to data collection errors, measurement inaccuracies, or erroneous human entries. Detecting accuracy anomalies necessitates no dataset transformation, as accuracy is intricately tied to data values and the information they contain. Hence, for this metric, the dataset is kept in its original form, allowing for an assessment of the accuracy of its values—be they accurate or not—by directly applying the anomaly detection model.

#### 3.2.2.2. Conformity

Conformity anomalies encompass values that do not conform to the expected data type. These anomalies are particularly prevalent in Big Data, where data from multiple sources may employ distinct data types and formats for the same column. To avoid errors and inconsistencies in data analysis and processing, it's imperative to address conformity anomalies. In light of their data type relationship, conformity patterns present data values in terms of their types. This entails a three-tier classification system encompassing numeric, string, and date data types. The conformity pattern is designed by transforming each data value into three binary digits: the first digit indicates whether the data value is entirely numeric, the second signifies complete alphabetic composition, and the third designates a date value. Subsequently, numeric values are represented as "100," string values as "010," date values as "001," and alphanumeric values as "000." The binary format facilitates the anomaly detection model's comprehension and processing of data values. Consequently, nonconforming records featuring different data types are distinguished in value in the new pattern and identified as outliers by the anomaly detection model. This representation can be extended to encompass other data types depending on the dataset.

### 3.2.2.3. Completeness

Completeness anomalies revolve around missing values, which can arise for various reasons, including human non-response, incomplete data collection, or data unavailability. The impact of missing data on analysis can be substantial, mainly when many values are absent. This can introduce bias into data analysis and compromise results. The completeness pattern is designed to reflect the extent of dataset completeness by converting each data value into a single binary digit, indicating whether it's missing. Consequently, data values are represented as '1' if they are absent and '0' if they are present. This classification distinguishes missing data values from the rest, flagging them as anomalies detectable by the anomaly detection model. As established during preprocessing, white spaces, 'N/A,' 'NA,' 'NULL,' and 'NaN' are cleaned and treated as missing values.

### 3.2.2.4. Uniqueness

A uniqueness anomaly signifies redundant entries referring to the same real-world entity. In expansive datasets, redundancy frequently occurs due to data collection from various sources, leading to multiple records with identical information in different formats. Such datasets, with redundant entries, misguide data analysis and impact accuracy. The dataset pattern should spotlight the degree of similarity among records to detect redundant entries as anomalies. This is achieved by employing the entity resolution framework detailed in the prior section to form duplicate pairs and compute the similarity score for each pair. Consequently, a dataset pattern is constituted by a collection of similarity scores for potentially matching records. Exceptionally high similarity scores signify the most similar records and are flagged as anomalies by the anomaly detection model.

### 3.2.2.5. Consistency

Inconsistency denotes a lack of alignment or agreement between two or more records or data about the same entity. Consequently, a consistency anomaly occurs when records referencing the same real-world entity expose conflicting information. For instance, two records may share the same name and ID number but have different addresses. Inconsistencies can stem from various factors, such as data input errors or the utilization of multiple data sources for integration. Detecting data consistency anomalies requires the preliminary identification of potentially redundant records, followed by a comparative analysis to unveil inconsistencies. Thus, this process employs the output of the Uniqueness anomaly detection as a dataset pattern, assessing the degree of similarity in data values among the potential redundant records.

### 3.2.2.6. Readability

Data quality encompasses not only data format but also semantic elements. Human-generated data, especially in large datasets, can enclose misspelled words and non-readable values, impeding users' ability to search, filter, or accurately analyze data. Consequently, data readability patterns highlight the extent to which data is intelligible and offers meaningful insights.

The data readability pattern is designed by converting each data value into a binary digit, signifying whether the data is readable using semantic libraries. As such, a data value is designated as '1' if misspelled and '0' if readable. This coding sets apart non-readable data values, marking them as anomalies discernible by the anomaly detection model.

### 3.2.3. Anomaly Detection

Once dataset patterns are formed for each quality dimension, these patterns are fed into an anomaly detection model to uncover the quality anomalies linked to each dimension. This framework deploys an extended isolation forest model for anomaly detection, building upon the original Isolation Forest algorithm. Isolation Forest is an unsupervised learning technique that isolates anomalies from regular data points by constructing isolation trees. Based on decision tree principles, these trees generate binary partitions via the random selection of a feature and a split value within that feature's range. Partitioning persists until all data points are isolated from the remaining samples. This process yields shorter paths in decision trees for anomalies, as outliers typically require fewer isolated partitions than standard instances. This allows for distinguishing anomalies from regular data. Notably, Isolation Forest excels in anomaly detection without requiring the creation of a normality model, making it especially suitable for the ambiguity often inherent in Big Data scenarios.

Moreover, Isolation Forest is known for its robustness against noise, enabling effective outlier detection even in noisy or irrelevant features, a common issue in Big Data scenarios. The algorithm further benefits from its swift computation, as it operates with a linear time complexity of $O(n)$, rendering it exceptionally well-suited to high-volume datasets. In this framework, an Extended Isolation Forest (EIF) version is employed for outlier detection. In the original algorithm, split branches can only align with the axes, leading to over-partitioning in regions with limited or single observations and resulting in inaccurate anomaly scores for some cases. In the extended version, the dataset is sliced with hyperplanes characterized by random slopes. Instead of selecting an unexpected feature and value, this approach establishes random gradients and intercepts for the branching cut. The result is a model that can capture more intricate anomalies and effectively manage high-dimensional data using hyperplanes rather than straight lines to isolate anomalies. As mentioned, the isolation forest operates as an unsupervised model, obviating the need for labeled data during training. The generation of decision trees within the isolation forest constitutes the training phase while scoring entails traversing data points through these previously trained trees. Data points are then classified as anomalies or non-anomalies based on the number of tree depths required to reach them. Typically, anomaly data points exhibit shorter tree paths than regular data points, as they are more readily isolatable. In this framework, the isolation model is applied individually to each dataset pattern to detect anomalies associated with each quality metric. This method facilitates the model's focus on the most pertinent features for each metric, contributing to a more precise identification of anomalies based on unique feature values. Once the model processes all dataset patterns, the detected anomalies are consolidated into a single dataset, effectively pinpointing quality anomalies associated with each data value.

### 3.2.4. Quality Anomaly Score Computing

The isolation forest operates by iteratively partitioning a subset of the dataset until each data point is isolated. Subsequently, the depth of each data point within each tree is determined. This depth represents the number of splits required to separate the data point within the tree. To measure the quality anomaly score of a data point, an average depth is computed across all trees within the forest. The underlying rationale is that anomalous data points require fewer splits and have a smaller average depth.

Consequently, the quality anomaly score measures the likelihood of a data point being an outlier regarding data quality. Higher anomaly scores indicate data points with a higher probability of being inferior quality, while lower scores refer to regular data points. For quality metrics wherein the dataset patterns are binary, such as Completeness, Readability, and Conformity, quality anomaly scores for detected anomalies are fixed at 100%. This is due to the categorical nature of these quality metrics, wherein the associated quality anomaly score can exclusively take values of either 0% or 100%.

Consequently, each data point concludes with detected anomalies corresponding to each quality metric and their respective quality anomaly scores. For example, applying the quality anomaly detection to the data value "05/2030" within the Birth Date column leads to an outcome of: "Accuracy (48%), Conformity (75%)," signifying that the data value is inaccurate by 48% and nonconforming by 75%. Once quality anomalies are evaluated for all data values, a global anomaly score associated with each quality metric is computed by averaging the quality anomaly scores for all detected irregularities within that metric.

Each quality metric thus possesses a resultant quality score, indicating the anomaly level within the entire dataset concerning that metric. Formally, the global anomaly score linked to a metric M is expressed as:

$$S_M = \frac{\sum_i S_{MDi}}{\sum_i D_i} \qquad (1)$$

$S_{MDi}$ refers to the anomaly score of the detected $D_i$ anomaly related to the Metric M. The quality anomaly score for the entire dataset refers to the overall degree of abnormality and the data's quality. Additionally, a global quality anomaly score for the whole dataset can be computed as the average of the anomaly scores for each quality metric. For enhanced precision, the global quality anomaly score of the dataset can be determined using a weighted average.

Given that quality metrics may vary in relevance, assigning higher weights to the most significant metrics ensures a more pronounced impact on the measured quality anomaly score. Formally, the global anomaly score is computed as:

$$S_A = \frac{\sum_i w_i \times S_{Mi}}{\sum_i S_{Mi}} \quad (2)$$

Where $w_i$ refers to the weight attributed to each metric, and $S_{Mi}$ refers to the average anomaly score related to the metric M. The dataset's quality anomaly score, denoted as $S_A$, provides valuable insights into the overall anomaly level.

A high-quality anomaly score implies a substantial presence of anomalous data points, indicative of poor data quality. Conversely, a low-quality anomaly score indicates a dataset with relatively few anomalous data points, signifying good data quality. Thus, the global quality anomaly score indicates the dataset's quality level and facilitates comparisons of anomaly levels among different datasets.

Moreover, it can be employed as a quality diagnostic tool, estimating the effort required to rectify the dataset based on detected anomalies and their respective quality anomaly scores.

Figure 12 is an in-depth visualization of the quality anomaly detection pipeline within the proposed framework. The subsequent section will delve into the framework's implementation, encompassing the tools and datasets employed in the process.

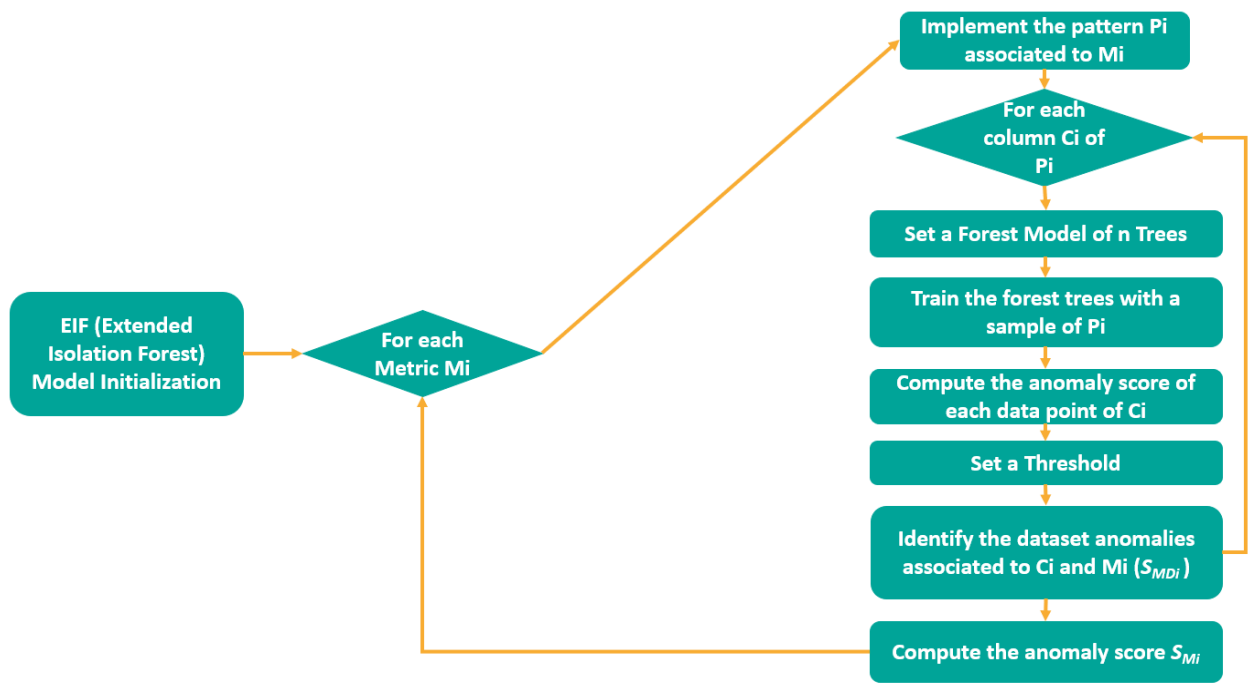

*Figure 12 : The Big Data Quality Anomaly Detection Framework Pipeline*

## 4. Implementation

### 4.1. Use Case and Datasets Description

This section describes the practical application of the deduplication and anomaly detection frameworks introduced in the previous section. We have implemented these frameworks using the following datasets:

**Dataset 1:** Our initial dataset comprises over 2 million synthetic records generated using a Python script [127]. This dataset contains different anomalies and is used to evaluate the various contributions of this thesis, namely data quality assessment, detection, and correction. It emulates personal information and features columns such as Name, Address, Gender, Age, and Salary. The anomalies within this dataset were introduced through alterations in data distribution, incorporating outlier values to scrutinize the accuracy metric, and the insertion of missing and erroneous data to address completeness, conformity, and readability metrics. Approximately 10,000 random rows were also subject to slight modifications and introduced as duplicates to assess Uniqueness and Consistency metrics. Using a synthetic dataset stems from the unavailability of pre-labeled datasets with known quality anomalies. Indeed, to accurately evaluate the effectiveness of our proposed frameworks, a dataset with pre-identified anomalies is required. Labeling such a dataset can be arduous and time-consuming, mainly when dealing with a massive dataset exceeding 1 million records. Hence, creating a synthetic dataset with pre-defined anomalies becomes essential for evaluating our framework's performance. Moreover, the synthetic dataset is a good option for assessing the framework's capabilities and limitations, as it contains intricate and challenging data anomalies. To verify the framework's real-world applicability, a representative subset of 100,000 records from the first dataset was meticulously labeled as anomalies or non-anomalies, thus enabling the assessment of Precision and the framework's accuracy for this first dataset.

Considering that the first dataset is synthetic, we have applied the deduplication framework on a second dataset (Dataset2) of real-world values to ensure the framework's accuracy:

**Dataset 2:** This dataset encompasses a small collection of 864 records containing information about restaurants, of which 112 are duplicates [128]. The dataset has been pre-labeled and previously employed in other research to evaluate the suggested deduplication approaches. Thus, despite its small size, we used this dataset to compare our deduplication framework against existing models in terms of accuracy.

For the same purpose, we have applied the quality anomaly detection framework on an additional dataset (Dataset 3) to evaluate our framework's performance in detecting real-world anomalies:

**Dataset 3:** The Taxi and Limousine Commission (TLC) Trip Record dataset [129] contains detailed information on every New York City taxi and for-hire vehicle trip. This dataset is regularly updated by the TLC, the governing body responsible for regulating and licensing the city's taxi and for-hire vehicle services. In this research, we extracted the green taxi trip data over 2022. The dataset is extensive, comprising over 1 million records, and encompasses data on green taxi trips, including pick-up and drop-off locations, travel distances and durations, fare details, payment methods, and more. Notably, the dataset contains a multitude of quality anomalies, including inaccuracies, missing values, duplicate records, and non-conforming records. The data relating to taxi trips is sourced from taxi meters, which are susceptible to various error sources such as hardware and software malfunctions, issues with wireless signals, and human interventions.

## 4.2. Adopted Architecture and Libraries

### 4.2.1. Architecture

Our frameworks have been implemented within the Cloudera Data Platform [130], a powerful platform for processing Big Data. The Cloudera Data Platform incorporates various tools and technologies designed for processing and analyzing large-scale datasets in a scalable and secure manner. We have opted for this Big Data platform for its advanced performance, ensuring efficient data processing. It is also compatible with various Big Data tools, including Hadoop, Spark, HBase, Kafka, and more. Data storage is managed in a distributed manner through the Hadoop Distributed File System (HDFS), designed to provide scalable storage for big data. HDFS offers horizontal scalability by seamlessly accommodating increasing data volumes by adding cluster nodes. It also allows fault tolerance by replicating data across multiple nodes, safeguarding against data loss in the event of node failures. Data processing is executed through Apache Spark, significantly leveraging Pyspark, a Python API of Spark. Spark is a robust Big Data processing framework known for its efficiency compared to traditional MapReduce processing in Hadoop. In our implementation, Spark collaborates with the Hadoop YARN cluster to deliver scalable and parallel data processing, benefiting from Hadoop's scalability and fault tolerance. In our adopted architecture, Zeppelin serves as the development platform, monitored by Ambari to orchestrate processing operations. Figure 13 offers a comprehensive overview of the architecture used in the implementation.

### 4.2.2. Libraries

The first step to implement the deduplication framework involves indexing data via Sorted Neighborhood to construct the training dataset. The sorted neighborhood is an indexing method that consists of sorting data values using a blocking key value and then employing a moving window across the sorted values to identify potential duplicates. The sorted neighborhood indexing method is particularly effective when dealing with large amounts of unstructured data. To enhance precision, weighted indexing was applied to the second dataset, enabling the accurate detection of most pairs. Subsequently, a similarity score is computed using Cosine Similarity. While alternative methods such as Euclidean distance and the Jaccard coefficient are plausible, Cosine Similarity is the most suitable for measuring text similarity, as supported by several studies [131]. Pairs are filtered based on predetermined minimum and maximum threshold ranges to heighten accuracy, retaining only the most similar records. The selected records are then grouped into clusters and employed as a training dataset for the deduplication process. In this approach, duplicate detection is not contingent on text comparison; instead, it relies on deduplication predicates and indexing rules generated by the model post-training. To this end, we have employed Dedupe, a Python library well-suited for accurate and scalable data deduplication and fuzzy matching based on machine learning [132]. The initial step involves creating a dedupe instance tailored to the dataset. The dedupe instance is then trained using the dataset crafted in the previous phase. After training, the model generates indexing rules that identify similar records. For example, in the second dataset, one of the generated predicates is (CommonTwoTokens, name), (SameSevenCharStart, name), (CommonThreeTokens, address). This indicates that records matching two-token names and three-token addresses are considered duplicates. Once trained, the model is employed for deduplication via a semi-supervised clustering method, clustering similar records based on a provided labeled dataset.

Data preprocessing is executed for anomaly detection using string functions and various Python libraries, including NLTK (Natural Language Toolkit), DateTime, and RE (Regular Expressions). Additionally, dataset patterns are generated using PySpark functions, transforming the original data into the required format. The pattern associated with Uniqueness and Consistency metrics is designed utilizing the labeling component of the deduplication framework. For Uniqueness and Consistency, data is indexed using the proposed entity resolution method to compute the similarity score of potential records. Each record is represented as a feature vector, and a weighted similarity score is calculated through Cosine Similarity. The Isolation Forest model uses the Sparkling Water library [133]. Sparkling Water is an open-source library integrating Apache Spark and H2O.ai [134], an open-source machine learning platform. This integration allows the combination of the high-speed, scalable machine-learning algorithms of H2O with the capabilities of Spark. To this end, an Isolation Forest model consisting of 100 trees and a sample size 265 is initialized. The sample size pertains to the number of randomly sampled observations used for training the isolation tree.

As mentioned, the model is augmented with the appropriate dataset pattern for each quality metric. The iterative application of the isolation model to each dataset pattern column culminates in detecting quality anomalies associated with that specific metric. Initial analysis of a subset of the dataset reveals the proportion of outliers within each dataset. Subsequently, the relevant quantile score is estimated and employed as a threshold for predicting quality anomalies. Once the anomalies are detected, their related quality anomaly scores are measured. Finally, the quality anomaly score for each metric and the global quality anomaly score are calculated.

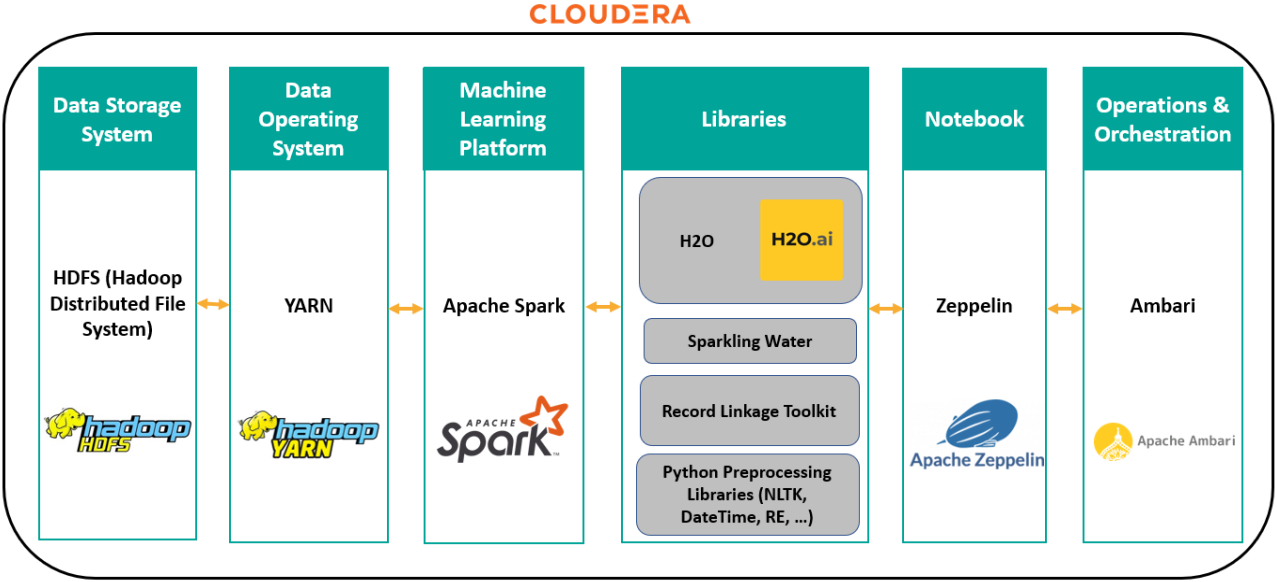

*Figure 13: Implementation Architecture*

## 4.3. Results

The performance of both the entity resolution framework as well as the quality anomaly detection framework was evaluated using the confusion matrix defined by the following metrics:

$$Precision = \frac{TP}{TP+FP} \quad (3)$$

$$Recall = \frac{TP}{TP+FN} \quad (4)$$

$$F-score = 2 * \frac{Precision*Recall}{Precision+Recall} \quad (5)$$

$$Accuracy = \frac{TP+TN}{TP+TN+FP+FN} \quad (6)$$

TP, FP, TN, and FN are True Positive, False Positive, True Negative, and False Negative, respectively. In evaluating the deduplication framework, "true negatives" refer to the instances where the framework correctly identifies and treats distinct records as different. In deduplication tasks, most records in a dataset are often non-duplicates. This is because most real-world datasets contain a much larger number of unique records than duplicate ones. Given the prevalence of non-duplicates, this would result in a high accuracy score even if the framework does not detect duplicate records accurately. Thus, we have not considered the accuracy metric when evaluating the deduplication framework, as this metric does not accurately reflect the framework's actual performance in identifying duplicates. The following figure shows the results obtained when applying the full deduplication on the first and the second datasets.

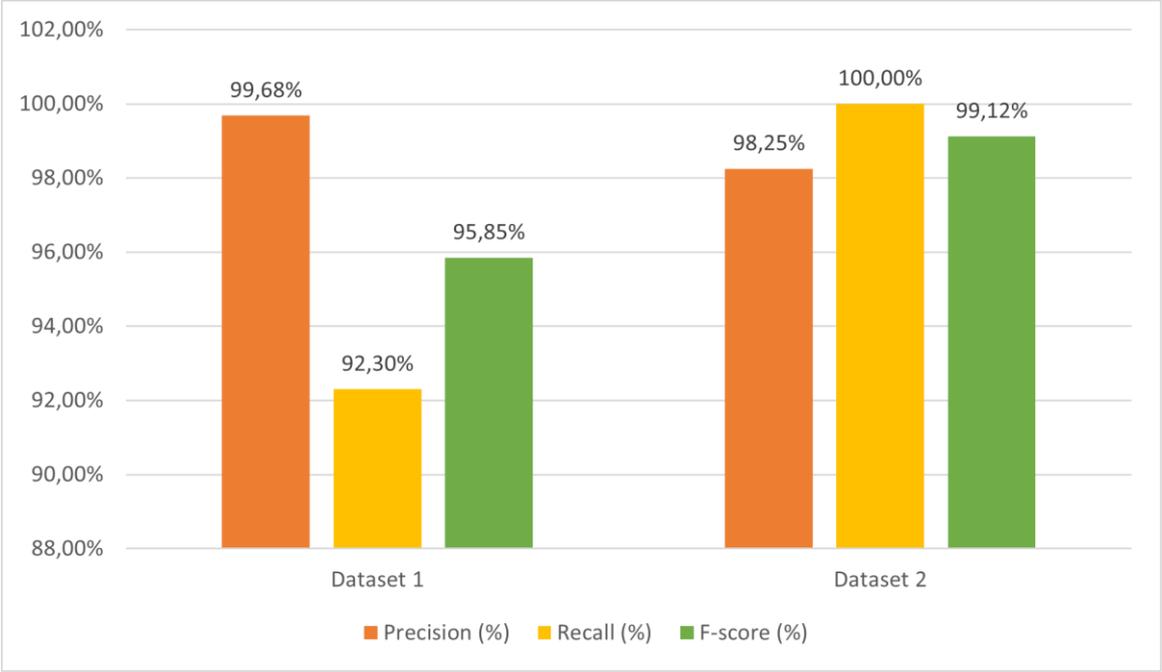

*Figure 14: Deduplication Framework Evaluation*

As mentioned earlier, we select the second dataset to assess our deduplication framework's performance against existing models previously applied to the same dataset. The models that have been used for this dataset include SDLER [135], DeepER [136], Magellan [137], and Dedupe [138]. The figure below presents a comparative analysis of the F-scores achieved by each of these models when used on the restaurant dataset. The outcome of this analysis distinctly demonstrates that our proposed framework outperforms the other models in terms of accuracy.

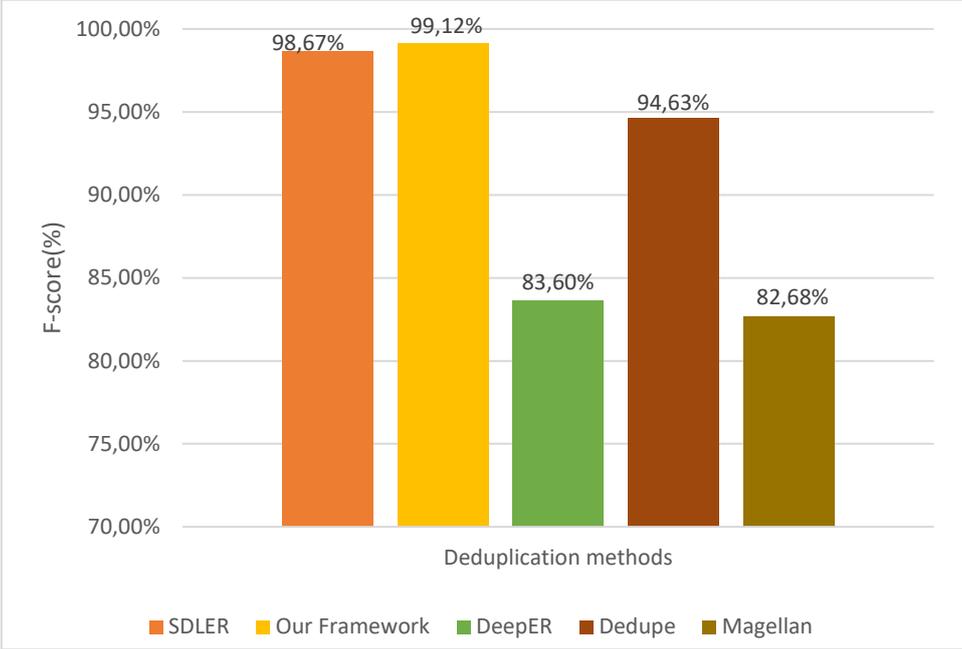

*Figure 15: F-Score Comparison*

To assess the effectiveness of the anomaly detection framework, a series of preprocessing steps were executed on both the first and third datasets. These steps involved the conversion of data values into appropriate formats to ensure that the anomaly detection model was not misled and the elimination of white spaces and stop words. Then, dataset patterns corresponding to each quality metric were generated for both datasets. The anomaly detection model was applied to each dataset pattern to identify anomalies specific to the corresponding metric. This procedure was carried out individually for each metric. It is worth noting that we introduced additional anomalies for the third dataset by eliminating certain values and converting specific data values into non-conforming formats. The results for each quality metric were computed by averaging the metric's scores across all columns. However, it should be noted that the assessment of the readability metric was deemed less pertinent for the third dataset, primarily due to the predominant numeric nature of the data entries. Consequently, no readability metric was evaluated for the third dataset. The outcomes for each quality metric for both the first and third datasets are presented in Figures 16 and 17.

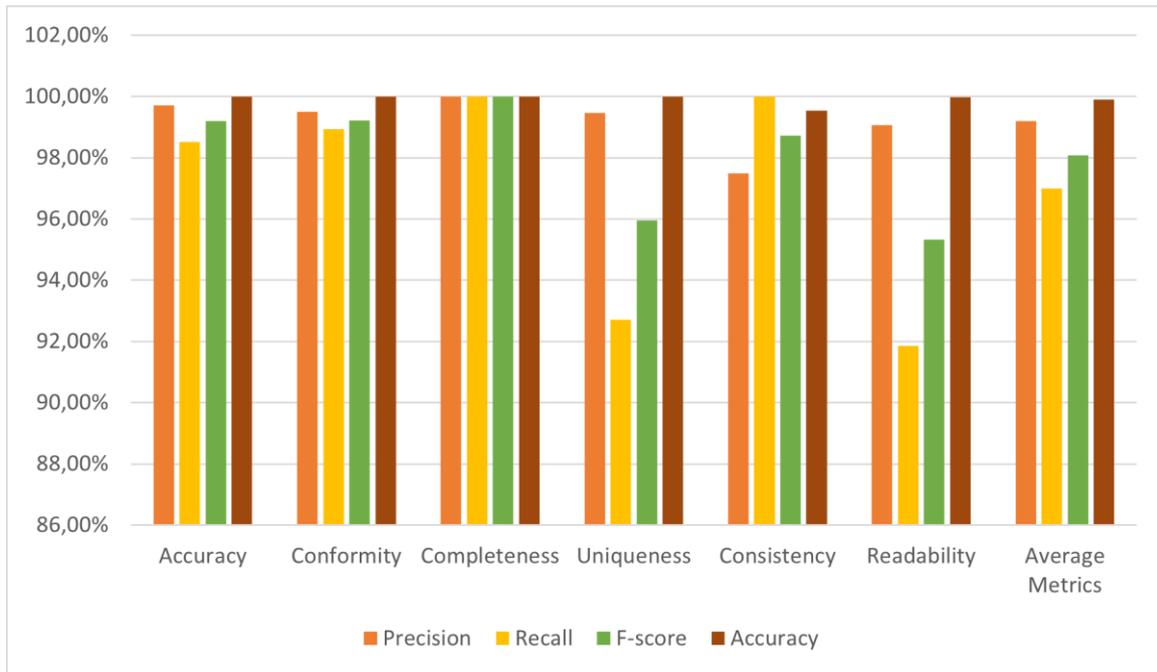

*Figure 16: Anomaly Detection - DATASET 1 CONFUSION MATRIX*

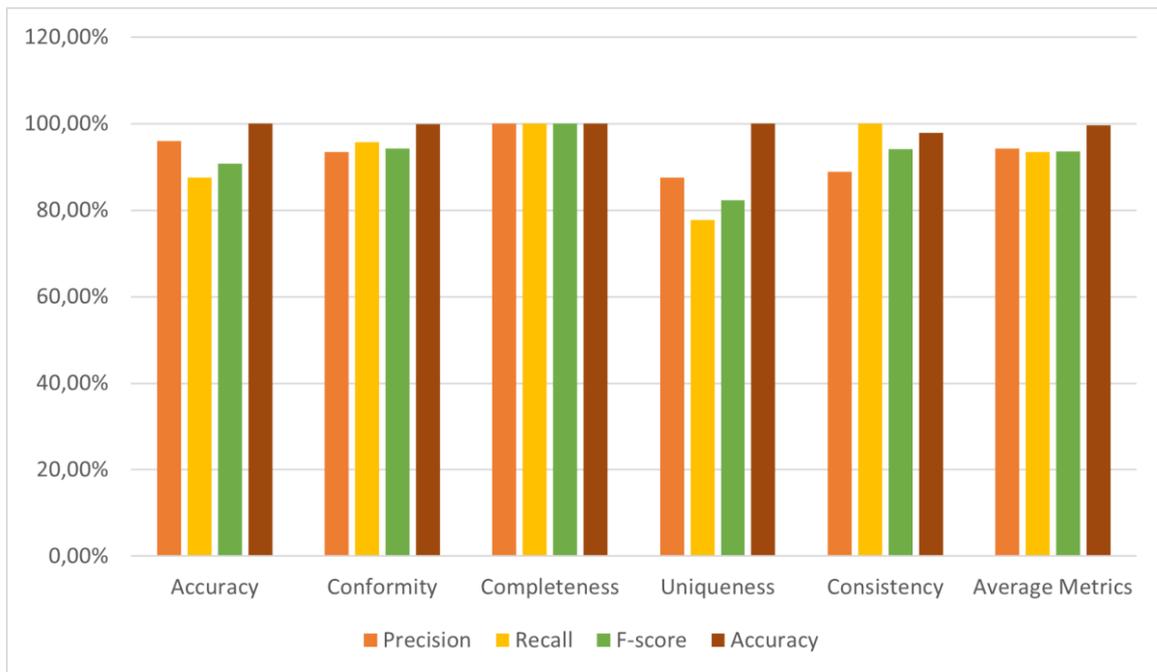

*Figure 17: Anomaly Detection - DATASET 3 CONFUSION MATRIX*

For the computation of the anomaly score, initially, the anomaly score for each metric and each column was measured based on the accumulated anomaly scores of the detected anomalies. Subsequently, the global anomaly score associated with each metric was derived by averaging the anomaly scores across all columns. Finally, the global anomaly score for each dataset was computed as the mean of the anomaly scores for all metrics. It is important to note that in this particular implementation, no specific weights were applied to the columns. The resulting anomaly scores for both datasets, encompassing each metric and the global anomaly score, are comprehensively presented in Figure 18.

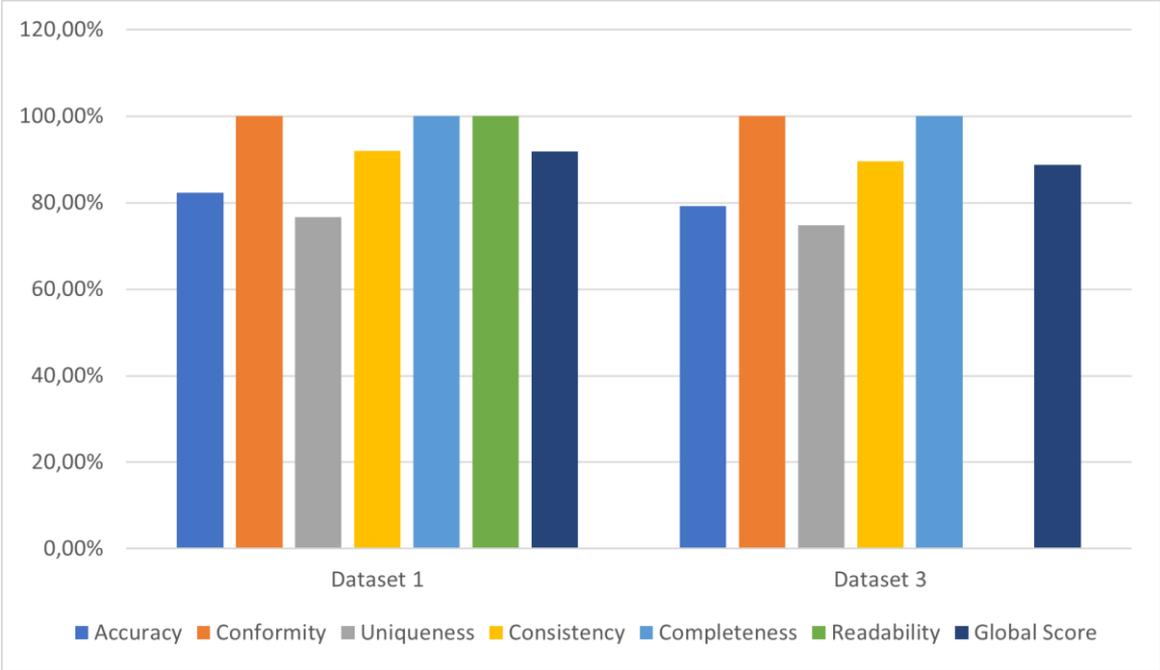

*Figure 18: Anomaly Detection - ANOMALY SCORES*

Given that the framework is designed to work for Big Data, it is essential to evaluate its scalability. To this end, we assessed the processing time entailed in the model's anomaly detection process and the execution time of the framework's entire pipeline. This holistic evaluation encompassed the preprocessing phase, the generation of dataset patterns, anomaly detection, and the computation of anomaly scores. A comprehensive schema of the processing time for both datasets is presented in Figure 19.

The findings from this evaluation underscore the framework's good performance in terms of processing time, which adheres to a linear $O(n)$ complexity. The framework's scalability is related to the implementation tailored to the Big Data requirements. This involved strategically using architectural elements such as Hadoop Distributed File System (HDFS), Spark, and distributed processing, collectively contributing to the solution's scalability. Furthermore, the implementation harnessed scalable libraries and models, such as Sparkling Water's isolation forest for anomaly detection and the sorted neighborhood method for indexing.

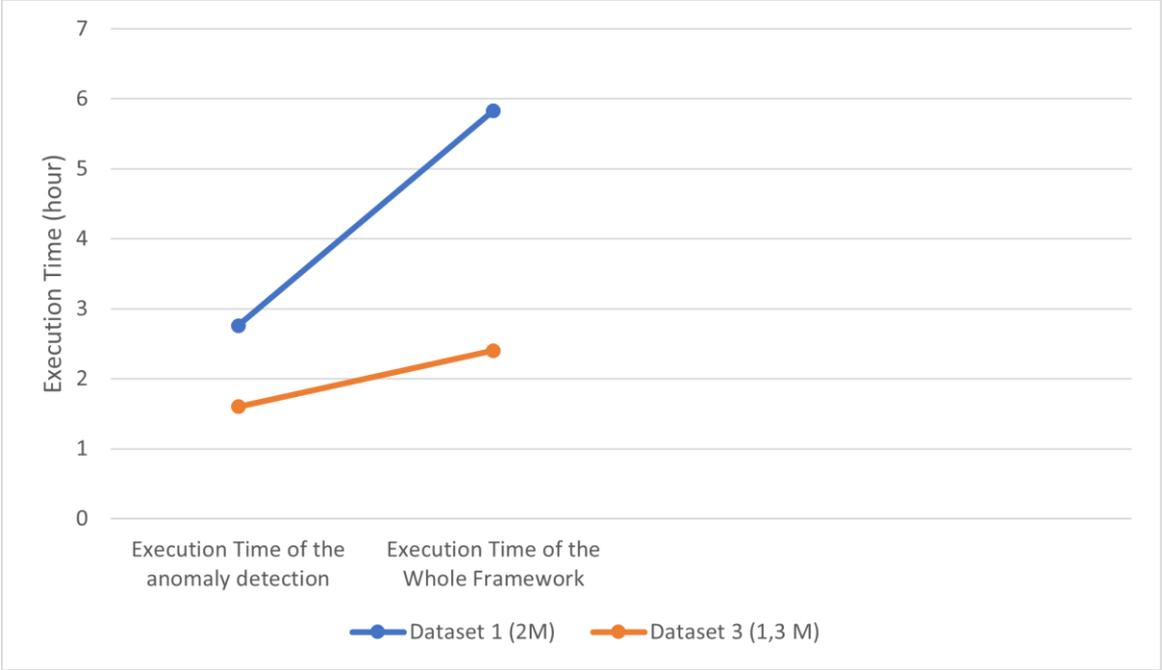

*Figure 19: Anomaly Detection - Execution Time*

## 4.4. Discussion

Based on the gathered results, the framework has demonstrated good performance in accuracy and scalability. It has shown a high capability to detect a wide spectrum of quality anomalies, with only a few extreme cases escaping detection. These undetected anomalies are linked to semantic nuances. The framework assessed the accuracy metric in the first dataset, effectively identifying inaccuracies in numeric and date values. However, it has shown a limitation in flagging inaccuracies within text data, as it fails to detect erroneous text values that conform to the prescribed format. Only text values with a correct format were seen as anomalies, particularly when present in duplicate records, which were rightly identified as inconsistent. Non-conform text values were generally detected due to the dataset pattern generated. The framework addressed the uniqueness metric, particularly in seeing duplicate records. It handles even challenging cases where partial modifications are introduced to the columns. Only extreme cases of duplicates were not detected where the entire content is completely inconsistent while referring to the same real-world entity.

For the third dataset, the accuracy metric evaluation excluded categorical encoded features due to the absence of outlier values. Instead, the model's focus was exclusively on continuous elements. Notable examples include detecting extreme cases such as trips exceeding 50 miles in distance, durations surpassing 2 hours, and fares exceeding $300. As for the first dataset, the model identified all missing values. Regarding the uniqueness dimension, the model detected most of the duplicated records. These identifications were based on attributes such as pickup datetime, locations, and other features. The framework executed a comparison of each record pair to pinpoint instances of consistency anomalies and successfully detected inconsistencies between these pairs. However, it's noteworthy that data values were also flagged as inconsistent while conveying the same information yet being presented in different formats, highlighting potential improvement that needs to be made to the framework. It is essential to emphasize that our proposed framework wasn't subjected to a comparative analysis with existing methodologies given the unique scope of our framework, which focuses explicitly on the realm of data quality anomalies. Thus, comparing our framework's performance with existing anomaly detection frameworks would be inappropriate due to the distinctive scope of each.

Nevertheless, our proposed framework outperforms the existing approaches in scope as it not only detects outlier values but addresses a broad range of data quality anomalies encapsulated within the six data quality metrics. Moreover, as applied to both datasets, the framework demonstrates noteworthy accuracy based on the metrics derived from the confusion matrix. Furthermore, its design aligns seamlessly with the requirements of Big Data environments, delivering efficient processing times and scalability characterized by a linear $O(n)$ complexity.

## Conclusion

The potential of Big Data to revolutionize operational and business performance across diverse domains is significant. Nevertheless, realizing these advantages hinges upon the enhancement of data quality. At the same time, literature has proposed numerous anomaly detection frameworks designed to address outliers in various domains. These frameworks overlook anomalies tied to data quality, encompassing missing, non-conforming, or inconsistent data. These anomalies can bias data analysis and impede an organization's capacity to derive meaningful insights from their data. In response to this pressing concern, we present a holistic approach to Big Data Quality Anomaly Detection. This approach allows the identification of generic data quality anomalies within the realm of Big Data in a generic way and without any correlation to a specific field. The detected anomalies are related to six quality dimensions: Accuracy, Consistency, Completeness, Conformity, Uniqueness, and Readability. To develop the suggested quality anomaly detection framework, an entity resolution framework was first developed as a preliminary stage for detecting uniqueness and consistency anomalies. The entity resolution framework was extended as an end-to-end standalone deduplication framework. Both frameworks were implemented using different datasets and have shown acceptable results in terms of accuracy and scalability. Once quality anomalies are detected, they must be addressed effectively to enhance the dataset's quality. The next chapter delves into the third axe of data quality management and suggests a practical framework for big data quality anomaly correction.

# Chapter 6: Big Data Quality Anomaly Correction Framework

## 1. Introduction

Detecting data quality anomalies is crucial, as it enables the identification of irregularities and inconsistencies within datasets, which is necessary to avoid wrong conclusions and bad decisions. However, anomaly detection does not fix the damage caused by these anomalies on data quality. Thus, to truly achieve effective data quality management, rectifying these anomalies and restoring the integrity of the data is imperative. In this last chapter, we focus on the most crucial part of managing data quality, which consists of correcting data quality anomalies. Even if existing studies have addressed data anomaly correction, many focus on specific contexts and lack a general approach. These context-specific studies are limited in their applicability to different situations and industries. Also, studies that addressed data quality anomalies have often been limited to addressing only one specific aspect, such as outliers or duplications. This limited focus fails to capture the entirety of data quality anomalies that may exist, neglecting other critical issues that could impact the reliability and accuracy of big data. A recent survey [73] about big data quality has stated the need for a comprehensive context-aware approach that handles big data quality anomalies. Therefore, there is a pressing need for intelligent and sophisticated techniques to address data quality anomalies holistically, considering the multifaceted nature of big data and encompassing all relevant quality dimensions. Such an approach would provide a more comprehensive and effective solution to enhance data quality and enable organizations to derive accurate insights and make informed decisions based on reliable and trustworthy data. To address the gaps in the field, we propose a Big Data Quality Anomaly Correction with three main contributions:

• The primary contribution of this research is the proposal of a novel framework for Big Data Quality Anomaly Correction. This framework adopts an intelligent and sophisticated methodology based on a predictive model, allowing them to address challenging quality anomalies.

• Another noteworthy contribution of this study is the comprehensive treatment of big data quality. This framework addresses six key dimensions of data quality: Accuracy, Completeness, Conformity, Uniqueness, Consistency, and Readability.

• Furthermore, this research proposes a framework not restricted to a specific field or domain. It is designed to be applicable across various areas, offering a generic approach to address data quality anomalies.

By introducing this framework, we aim to provide a comprehensive solution for addressing big data quality anomalies, offering a more holistic approach to data quality management. The rest of this chapter is organized as follows: The next section reviews the most recent related works addressing data quality. Section 3 presents the different steps of the proposed big data quality anomaly correction. Then, the proposed framework is implemented, and the obtained results are discussed. Finally, conclusions are made, and future word directions are highlighted.

## 2. Big Data Quality Anomaly Correction

Data anomaly correction is a crucial process in data quality management aimed at rectifying anomalies, errors, or inconsistencies within a dataset. Anomaly correction requires a preliminary anomaly detection phase that consists of identifying anomalies within the dataset. Then, the specific data points or records responsible for these anomalies are precisely located. This localization step is crucial for isolating and addressing the issues within the dataset. Then, anomaly correction is performed according to different approaches:

**Removing Erroneous Values:** In some cases, the most straightforward approach to anomaly correction is simply removing the erroneous data points from the dataset. This is often a suitable solution when the anomalies are erroneous, or their impact on the overall dataset is minimal. However, this approach can result in information loss if the anomalies represent infrequent occurrences.

**Smoothing Out Irregular Data Patterns:** An alternative strategy involves smoothing out irregular data patterns. This method is beneficial when dealing with time-series data or data with inherent noise. Techniques like moving averages or exponential smoothing can be applied to moderate the impact of anomalies while preserving the overall trend in the data. This approach is valuable when maintaining the continuity of data is essential.

**Rectifying with Replacement Values:** In some cases, rather than removing anomalies, replacing them with new values consistent with the rest of the dataset is more appropriate. This might involve imputing missing values or substituting erroneous entries with more accurate estimates based on statistical methods or machine learning algorithms. The goal is to maintain the dataset's integrity and completeness while effectively addressing the anomalies.

In this chapter, we adopt the third approach to anomaly correction as it proves to be one of the most efficient methods, enabling us to address anomalies without information loss. Several techniques are employed to replace anomalies with accurate values. One common strategy involves statistical methods, which replace anomalous data points with measures such as the dataset's mean, minimum, maximum, or most frequent values. Instead of relying on fixed rules or heuristics, there is also a better option of using machine learning algorithms that learn from the data to make predictions about the correct values. Thus, the algorithm goes through a training phase where it learns to map input features to output values. In the context of anomaly correction, the input is the dataset with anomalies, and the output is the corrected values. After training, the model can predict the correct values for anomalies based on the learned patterns. Among the machine learning algorithms commonly used for anomaly correction, linear regression models the relationship between input features and the target variable (corrected value) using a linear equation. K-NN also determines the fixed value by averaging the values of the k-nearest data points from the anomaly based on specific distance metrics. Decision trees, Support Vector Machines (SVM), and random forests are also among the notable algorithms used in this domain. Deep learning techniques, such as neural networks, have also demonstrated their proficiency in capturing complex data relationships, making them suitable for addressing anomalies within complex datasets. As previously highlighted in the context of anomaly detection, anomaly correction approaches also tend to handle a single aspect of data quality with a predominant focus on completeness by imputing missing values. However, data anomalies can manifest in different forms, such as inaccuracies, outliers, inconsistencies, duplicates, etc.

Therefore, despite the broad scope of anomalies encompassing multiple data quality dimensions, anomaly correction has rarely been performed comprehensively, addressing data quality. This underscores the need to develop more generic and versatile anomaly correction models that tackle all quality anomalies, considering their diverse aspects without being limited to a specific field or domain. Such generic models offer a holistic approach to data quality enhancement, ensuring that data is reliable, consistent, and fit for various applications, regardless of the domain it originates from. Thus, through this chapter, we aim to enhance the quality of big datasets by suggesting a novel Big Data Quality Anomaly Correction Framework. This framework adopts an intelligent and sophisticated methodology based on a predictive model, allowing them to address quality anomalies by predicting the correct values in an automated way. Another noteworthy contribution of this study is the comprehensive treatment of big data quality. Unlike previous research that often focuses on specific aspects of data quality, this framework addresses six key dimensions of data quality: Accuracy, Completeness, Conformity, Uniqueness, Consistency, and Readability. Furthermore, we propose a framework not restricted to a specific field or domain. Instead, it is designed to be applicable across various areas, offering a generic approach to address data quality anomalies.

## 3. Big Data Quality Booster Framework through Predictive Analysis

In the realm of Big Data analysis and decision-making, the accuracy and reliability of datasets play a pivotal role. Nevertheless, anomalies such as inaccuracies, missing values, or inconsistencies frequently raise significant challenges to data quality. To tackle this issue, we've introduced an innovative framework for correcting quality anomalies in Big Data quality. The principal goal of our framework is to enhance the overall quality of big datasets by correcting and addressing data quality anomalies. Given that our framework is generic and applicable across various data contexts, we acknowledge that predicting exact corrections for every quality anomaly in a generic way without any field correlation is unrealistic. Indeed, factors like inherent variability of data types, contextual dependencies, and data uncertainties and complexity make it extremely challenging to achieve absolute precision in predicting values for all data use cases. Instead, our objective through this framework is to provide an improved and more precise version of big datasets by addressing data quality anomalies and substituting them with enhanced and more reliable data values that closely align with the valid ones. As demonstrated in Figure 20, the proposed framework comprises four primary phases.

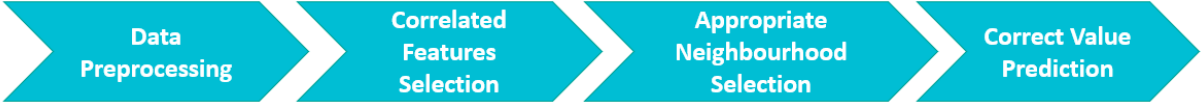

Figure 20:The Big Data Quality Anomaly Correction Framework

## 3.1. Data Preprocessing:

An initial data preprocessing phase is imperative in big data, characterized by its typically unstructured nature and inherent irregularities. Below, we present the essential preprocessing techniques considered vital for the anomaly correction framework. Nevertheless, additional preprocessing methods may be integrated based on the contextual requirements of the dataset use case.

**Feature Selection:** Feature selection identifies and selects the most pertinent attributes from a dataset. It allows for reducing data dimensionality and eliminating irrelevant or redundant features. Given that predictive models heavily rely on the existing data for making accurate predictions, it is imperative to cleanse the dataset of non-relevant information that could introduce bias into the model's accuracy so only the relevant features are retained.

**Feature Extraction:** Sometimes, some vital information for prediction may be hidden within the raw data. This is where feature extraction comes into play. Feature extraction transforms the raw data into a derived attribute set that encapsulates the crucial information. Its objective is to decrease data dimensionality while preserving the essential characteristics. Through feature extraction techniques, we can unearth additional valuable attributes that enhance the predictive capabilities of our models.

**Encoding:** As mentioned earlier, encoding pertains to converting categorical or textual data into numerical representations suitable for processing by predictive models. Encoding is crucial because many predictive models require numerical inputs for analysis. However, real-world data frequently comprises categorical or textual variables that these models cannot directly use. Therefore, encoding techniques translate non-numerical data into numerical representations, making them compatible with predictive models.

**Normalization and Scaling:** Inconsistent scales can distort analysis and potentially yield misleading or biased results. To address this issue, we employ normalization, which entails scaling numeric attributes to an expected range. Normalization is indispensable as it enhances the convergence of predictive models, prevents attributes with larger magnitudes from dominating the analysis, and ensures equitable comparisons between different attributes.

**White space and symbol removal:** As explained earlier, this process involves data cleansing by eliminating unnecessary white spaces, punctuation marks, or special symbols that contribute nothing to the analysis. This step is particularly pertinent for predictive models to enhance the consistency and reduce the noise of the training dataset.

## 3.2. Correlated Features Selection

After the initial preprocessing phase, the next crucial step involves selecting correlated features specific to each detected quality anomaly in the dataset. The imputed elements significantly impact the predictive model, which is critical to its performance and effectiveness. For this, only features correlated with the detected anomaly should be considered by the predictive model. By considering only the correlated features, we avoid introducing unnecessary noise or irrelevant information into the model so it can effectively learn the patterns and relationships specific to the detected anomaly, leading to more precise predictions and improved anomaly correction. For example, if an abnormality is detected in a sales dataset related to "product price," it would be sensible to prioritize correlated features like "product category," "time of sale," or "customer segment." These features will likely directly impact the product price and provide valuable context for understanding and addressing the anomaly. Correlated feature selection can be done manually through human intervention, leveraging domain knowledge and expertise, or automatically using various techniques, such as statistical measures or machine learning algorithms. In this framework, the correlated features are selected iteratively for each detected quality anomaly using a hybrid approach combining manual and automatic correlation selection strengths. This approach involves leveraging automated techniques to identify correlated features, followed by human supervision to refine and validate the selections. First, the correlated features are captured based on a data-driven approach that consists of a statistical measure of the correlation coefficients that can quantitatively assess the relationships between components and identify the ones that reveal a significant correlation with the detected anomaly. Correlation coefficients provide a numerical value that indicates the strength and nature of the correlation between the variables, ranging from -1 to +1. Then, domain experts can review and adjust the automatically selected features based on their contextual knowledge, ensuring the inclusion of domain-specific insights that may not be captured by automated methods alone. This hybrid approach combines the advantages of both human expertise and data-driven analysis, leading to more accurate and robust feature selection. In some scenarios, the correlation coefficients of the dataset features may be close, indicating that all features are slightly correlated with the detected anomaly. Even though the individual correlations may be weak, collectively, these features may provide valuable insights when analyzed together. In such a scenario, we recommend including and considering all the dataset features in the subsequent phases of the framework and attributing them to the predictive model.

## 3.3. Appropriate Neighborhood Selection

After selecting the correlated features related to the anomaly, another crucial step is filtering the records to include only the closest ones to the anomaly. This phase focuses on identifying the neighborhood of the anomaly, which refers to the subset of data points or records that will be attributed to the predictive stage for more precise anomaly correction. The selection of this neighborhood varies depending on the quality dimension the anomaly addresses. In the following, we outline how the neighborhood is chosen for each of the six Big Data quality dimensions addressed in this framework:

### 3.3.1. Completeness

Neighborhood selection involves including records similar to the missing or incomplete data points when addressing completeness-related anomalies by considering other complete data values related to the selected correlated features. The proposed framework is designed to handle missing values in both categorical and continuous data. For example, let's consider a dataset related to customer information where some customer records have missing values for their purchase history. In this case, the neighborhood selection aims to identify records with similar customer attributes, such as age, gender, and location, with complete purchase history information. By including these records in the neighborhood, the framework will use their purchase patterns and behavior to impute missing values for customers with incomplete purchase histories. Our proposed framework addresses missing values in categorical and continuous data. The framework predicts the most appropriate value for categorical features based on the available complete data. Since the missing values in categorical features belong to a limited set of possible options, the framework can predict accurate and exact values for the missing categorical values.

On the other hand, predicting exact values for missing data in continuous features is more challenging. In such cases, the framework adopts a different approach. It leverages feature extraction techniques to extract relevant features from the dataset, providing additional insights about the missing data. This enables the framework to estimate a potential range of values for the missing data, offering a valuable approximation that helps fill the gaps in the dataset. For instance, consider the prediction of a person's ID number. The framework uses the available features such as the person's date of birth, city, and province. The framework can generate a range of possible values within which the missing ID number may fall by analyzing the relationship between these features and the ID number. This approximation, such as transitioning from null values to a value like "AB1234XX," provides a practical solution for completing the dataset and addressing completeness-related anomalies.

### 3.3.2. Accuracy

When an inaccurate value, such as an outlier, is identified, the framework treats it as missing. Indeed, the inaccurate values cannot be used to predict the correct values, so they are just cleared and addressed as if they were missing data. As for missing data, the framework leverages the available data and analyzes the relationships among variables within the dataset to rectify inaccurate values. By examining the patterns and dependencies among the features, the framework learns how the features are interconnected and then uses the gained knowledge to predict a more accurate value for the outlier.

### 3.3.3. Uniqueness & Consistency

In this process, the framework introduces a unique and empty row that serves as a placeholder for the merged record. By leveraging the collective information from all instances of the duplicated data, the framework predicts accurate values for each feature within the introduced empty record. This prediction considers the data from all repeated cases, allowing for a comprehensive analysis of the available information. Through this approach, the framework resolves the duplication issue and addresses any inconsistencies among the duplicated instances.

### 3.3.4. Conformity

Addressing Big Data conformity is essential to ensure interoperability and effective data integration, enabling improved decision-making based on homogenous and well-structured big data. Our framework incorporates a semantic-based approach to manage non-conforming data that deviates from the expected data types. Using word embedding techniques, the framework identifies the most closely related values regarding semantics. These selected rows presenting similar semantic characteristics are then imputed into the predictive model. Leveraging the power of machine learning and statistical analysis, the model can accurately correct the non-conforming values based on the patterns and relationships within the selected data subset. For instance, let's consider a conformity anomaly within a categorical feature where the expected values are "1st class" or "2nd class." Suppose we encounter a non-conforming value of "1" within this feature. By applying word embedding techniques, the framework can identify rows that exhibit similar semantic characteristics to the non-conforming value. In this case, the framework would select rows that contain the value "1st class" for the definite feature. These selected rows, which share a semantic similarity with the correct value, are then used for imputation in the predictive model. As a result, the non-conforming value of "1" is transformed into the valid value of "1st class." This semantic-based methodology ensures that data non-conformities are effectively addressed, leading to accurate corrections.

### 3.3.5. Readability

To address readability anomalies, the framework employs a method that selects the most similar correct values to the erroneous data entry to correct non-readable data values. Using string matching algorithms and Cosine similarity measures, the framework identifies the closest matching values within the dataset. These selected values are then used to correct the misspelled and erroneous entries using the predictive model embedded within the framework. For instance, consider a city feature where an incorrect entry, "Moreal" is present. The framework applies string-matching algorithms to find the closest matching values within the dataset. In this case, the algorithm identifies "Montreal" as a close match to the erroneous entry. By selecting rows that contain the correct value, "Montreal," the framework can impute these values and correct the incorrect entry.

## 3.4. Correct Value Prediction

After identifying and selecting the relevant records for the detected anomaly, the subsequent step in the framework involves imputing these selected records into the predictive model for accurate value prediction. In the proposed framework, we employ the XGBoost predictive model. XGBoost is a robust and widely used machine learning algorithm known for its effectiveness in handling structured and unstructured data. It is an ensemble learning algorithm that corrects the detected quality anomaly by iteratively building decision trees. It uses a gradient-boosting framework to make accurate predictions by combining multiple weak models into a robust ensemble model. One of the key advantages of using the XGBoost model in this framework is its exceptional predictive performance. It is widely recognized for achieving high accuracy in regression and classification tasks. This makes it reliable for accurately predicting the correct values for the selected records.

Moreover, XGBoost is well-suited for handling large-scale datasets, making it appropriate for big-data scenarios. It uses parallel computing techniques to process vast amounts of data and is compatible with distributed computing frameworks such as Apache Hadoop and Apache Spark. Another reason XGBoost is well-suited for big data scenarios is its robustness against noisy data and outliers. Indeed, XGBoost employs regularization techniques that help reduce the impact of noise and outliers on the model's performance. Thus, the XGBoost model is well-suited for anomaly correction tasks as it can handle complex relationships and patterns within the data. By feeding the selected records into the XGBoost model, the framework employs the model's ability to learn from the existing data and enhance Big Data quality. The imputed values provided by the XGBoost model serve as a correction for the anomalies and contribute to improving the overall data quality. The model analyzes the input features and uses the patterns and relationships within the dataset to predict the correct values for the detected anomalies. It starts by preparing the imputed correlated features and closest records selected from the previous stages. Then, it trains an initial decision tree and calculates the errors. The algorithm constructs subsequent trees, learns from the mistakes of previous ones, and updates predictions by summing up the outputs of all trees, giving more weight to accurate ones. This iterative process continues until it reaches a level of improvement. Finally, the corrected value for the anomaly is obtained by combining the predictions from all trees. Through this iterative tree building and prediction updates, XGBoost learns from the imputed features and records to accurately predict the correct value for the anomaly. This framework is applied to all detected quality anomalies, ensuring that the valid values are predicted based on the imputed features and records specific to that anomaly. Thus, by leveraging the power of XGBoost and the information from the selected close records, our framework effectively corrects data quality anomalies, enhancing overall data accuracy and reliability. Figure 21 shows the global pipeline of the suggested big data quality anomaly correction framework. Figure 22 shows the pseudo-code of the anomaly correction framework using the predictive model XGBoost.

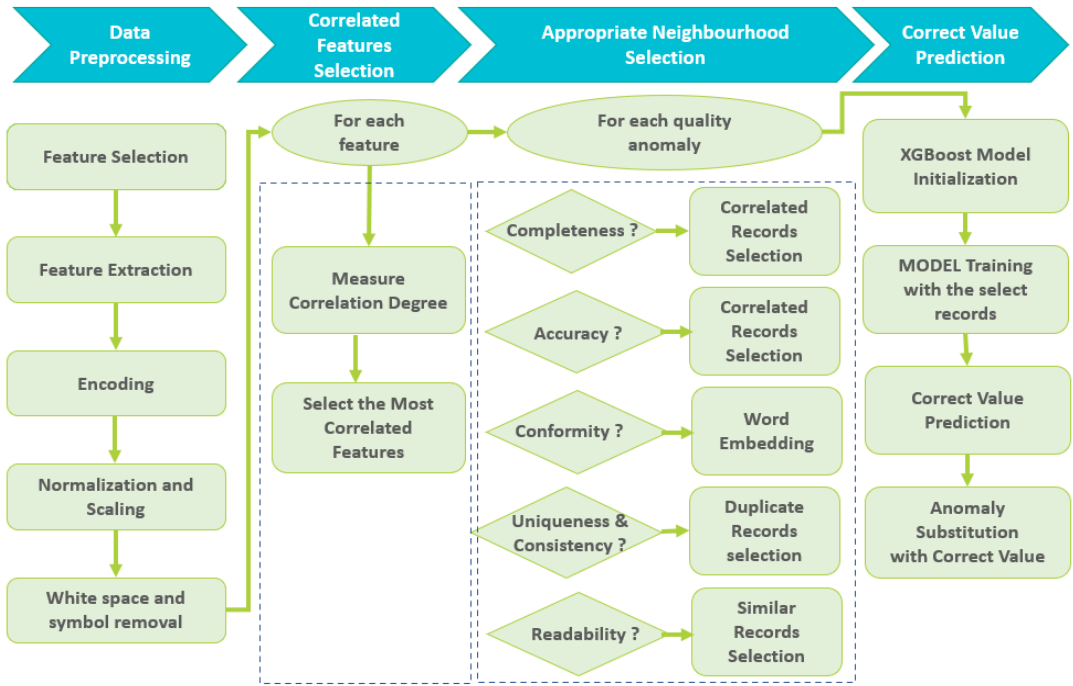

*Figure 21: The Global Pipeline of the Big Data Quality Anomaly Correction Framework*

| Algorithm : Anomaly correction using XGBoost model |
|---|

**Input :** D : Preprocessed Dataset D with previously detected quality anomalies

**Output :** D : Dataset D with corrected anomalies

**1 :** D'=D

**2:** XGBoost = H2OXGBoostEstimator(booster='dart', normalize_type="tree", seed=1234)

**3: for each** F in D.columns **:**

**4:**     **for each** F' in D.columns **:**

**5:**         if( correlation(F,F') < corr_tresh )

**6:**             D'.drop(F')

**7:**     **end**

**8:**     **for each** row in D':

**9:**         value = row[column F]

**10:**         **switch** (value):

**11:**             **case** is_completeness_anomaly :

**12:**                 training_dataset=D'.drop(row)

**13:**             **case** is_accuracy_anomaly :

**14:**                 set value = null

**15:**                 training_dataset=D'.drop(row)

**16:**             **case** is_readability_anomaly :

**17:**                 sim_scores= calculate_similarity(D'[value_index], D'.rows)

**18:**                 similar_idxs = indices where sim_scores > we_tresh

**19:**                 training_dataset = select rows[similar_idxs] from D'

**20:**             **case** is_conformity_anomaly :

**21:**                 f_we = transform_to_word_embedding(D'[column F])

**22:**                 sim_scores= calculate_similarity(f_we [value_index], f_we.rows)

**23:**                 similar_idxs = indices where sim_scores > sim_tresh

**24:**                 training_dataset = select rows[similar_idxs] from D'

**25:**             **case** is_consistency_anomaly :

**26:**                 training_dataset=duplicaterows(row)

**27:**             **else :**

**28:**                 break

**29:**         XGBoost.train(training_dataset)

**30:**         correct_value = XGBoost.predict(value)

**31:**         D.F[index_value]=correct_value

**32: end**

*Figure 22: Pseudo-code of the anomaly correction framework*

# 4. Implementation

## 4.1. Datasets Description

Among the various potential use cases presented in the previous section for the proposed framework, we have specifically chosen the domain of personal information. This extensive use case includes a variety of sub-use cases, including datasets related to customers' data, passengers' data, or users' data on various platforms. This use case is selected because of its wide prevalence and relevance in the practical real world and data-driven era. Indeed, human information datasets are extensively used in numerous industries and domains, from e-commerce and healthcare to transportation and social media platforms. Moreover, these datasets contain various quality anomalies due to multiple factors, such as human errors during data entry, system inconsistencies, or evolving data formats. Thus, ensuring the accuracy and quality of such datasets is crucial for maintaining data integrity and making informed decisions.

**Dataset 1:** The first dataset used in this study is the same synthetic dataset used in the previous chapter [127]. We have opted for this dataset for the same reasons previously mentioned. Indeed, it is essential to have a dataset with preexisting abnormalities to assess the suggested framework's performance accurately. To address various quality metrics, anomalies were intentionally introduced into the dataset. These anomalies encompass changes in data distribution with outlier values to evaluate the accuracy metric, modifications to existing data values, and presenting missing or incorrect data to assess completeness, conformity, and readability metrics.

Additionally, approximately 10,000 random rows were slightly modified and added as duplicates to evaluate uniqueness and consistency metrics. By introducing challenging and confusing data anomalies, we can thoroughly assess the framework's ability to handle diverse and complex scenarios. Thus, using a synthetic dataset enables us to determine the effectiveness and limitations of our framework more comprehensively.

**Dataset 2:** The second dataset is The Titanic [139], a well-known data analysis and machine learning dataset. It consists of information about passengers on the Titanic, including their demographic data, such as age, gender, class, and survival status. The dataset contains 1309 rows, each representing a passenger, and several columns containing different attributes. Despite being a relatively small dataset, the Titanic dataset was selected to assess the performance of the suggested framework in handling real-world data quality anomalies. The Titanic dataset is particularly relevant due to quality anomalies in the context of quality anomaly correction. These anomalies include missing values, inaccurate entries, duplicate records, inconsistencies, and other data quality issues. To further enhance the evaluation of the framework and test its capabilities, the Titanic dataset was deliberately augmented with additional quality anomalies. These different anomalies were introduced to challenge all aspects of data quality, including completeness, consistency, uniqueness, conformity, readability, and accuracy.

## 4.2. Implementation Details

### 4.2.1. Implementation Architecture and Libraries

We used the same architecture employed in the previous chapter to implement the proposed framework. In data preprocessing, a combination of Python string functions and specialized libraries such as NLTK (Natural Language Toolkit), DateTime, and RE (Regular Expressions) was employed. These libraries offer tools tailored for text manipulation, date-time handling, and pattern matching. Special symbols were systematically removed from text values using Python's string functions to ensure consistent data representation. Moreover, specific placeholders commonly encountered in datasets, like 'N/A,' 'NA,' 'NULL,' and 'NaN,' were systematically cleared, making them recognized as missing values. Furthermore, all textual data were lowercase to represent data values consistently. The XGBoost model was implemented using the Sparkling Water library.

Sparkling Water integrates Apache Spark with H2O [134], an open-source machine-learning platform known for its fast and scalable algorithms. The XGBoost model for the Titanic dataset was built and trained using the H2OXGBoostEstimator with specific parameter settings. The booster parameter was set to 'dart,' which indicates using the Dart boosting algorithm. This algorithm incorporates dropout regularization, which helps prevent overfitting and improves the model's generalization. Additionally, the normalize_type parameter was set to "tree." This parameter specifies the normalization type used during tree construction in XGBoost. This normalization technique helps balance the influence of individual trees and maintain stability during the learning process. The seed parameter was set to '1234', a random seed value used for reproducibility. To prepare the dataset for addressing conformity anomalies, we used the word embedding Word2Vec capabilities of the H2O library [140], which allows computing and applying word embeddings to the dataset in a scalable and distributed way. Observation embedding techniques enable the framework to capture the semantic relationships between data values and identify closely related values. To address readability anomalies, we employed various measures from the Python library, including Levenshtein distance, cosine similarity, and Jaccard similarity. These measures compare the similarity between text values and identify the most closely matching values.

### 4.2.2. Execution Process

After preprocessing the dataset to handle data quality anomalies, the proposed framework is applied by following a series of steps. Firstly, the relevant features correlated to the specific quality anomaly being addressed are selected. Then, the rows that do not exhibit any quality anomaly in the selected feature are chosen. Next, a filtering process is applied to retain only the closest records based on the nature of the addressed quality dimension. The selected rows are then used as the training dataset for the XGBoost model. The XGBoost model is initialized with the abovementioned settings, such as the booster type and normalization technique. The training phase involves optimizing the model's parameters and leveraging the powerful predictive capabilities of XGBoost to learn patterns and relationships within the dataset. Once the training phase is completed, the model is applied to the entire dataset, including the rows with quality anomalies. Using the imputed data and features, the XGBoost model employs its predictive capabilities to correct the detected quality anomalies or provide approximate values, depending on the quality anomaly being addressed. This approach ensures that the data quality issues are effectively fixed or compensated for, leading to improved accuracy and reliability of the dataset.

## 4.3. Results

To evaluate the performance of the proposed framework, we have measured the accuracy metric for each addressed quality dimension. The accuracy metric measures the proportion of correct anomaly corrections compared to all managed values. It indicates the framework's ability to correct quality anomalies in the dataset. We also measured the error rate, which represents the proportion of incorrect corrections about all addressed values, to identify areas that require further improvement or fine-tuning in the framework.

$$Accuracy = \frac{Number\ of\ correct\ predictions}{Total\ predictions}$$

$$Error\ Rate = \frac{Number\ of\ incorrect\ predictions}{Total\ predictions}$$

We employed different approaches depending on the dataset used to classify and determine the accuracy of anomaly corrections. For the first synthetic dataset with pre-labeled anomalies, we compared the corrected values to the known correct values in the dataset. Since the valid values were already available, we could directly assess the accuracy of the anomaly corrections by comparing the corrected values and the actual values. We manually labeled the dataset for the second dataset, which lacked pre-labeled data, by carefully inspecting the values. This manual labeling process was feasible due to the relatively small size of the dataset. We examined each value and determined whether it was an anomaly based on our domain knowledge and understanding of the dataset.

Additionally, we considered any additional anomalies that were intentionally included in the dataset during the evaluation process. This manual labeling allowed us to establish a reliable reference for evaluating the accuracy of the anomaly correction framework for real datasets. Figures 23 and 24 show the accuracy and error rate measurements obtained for the first and the second datasets, respectively.

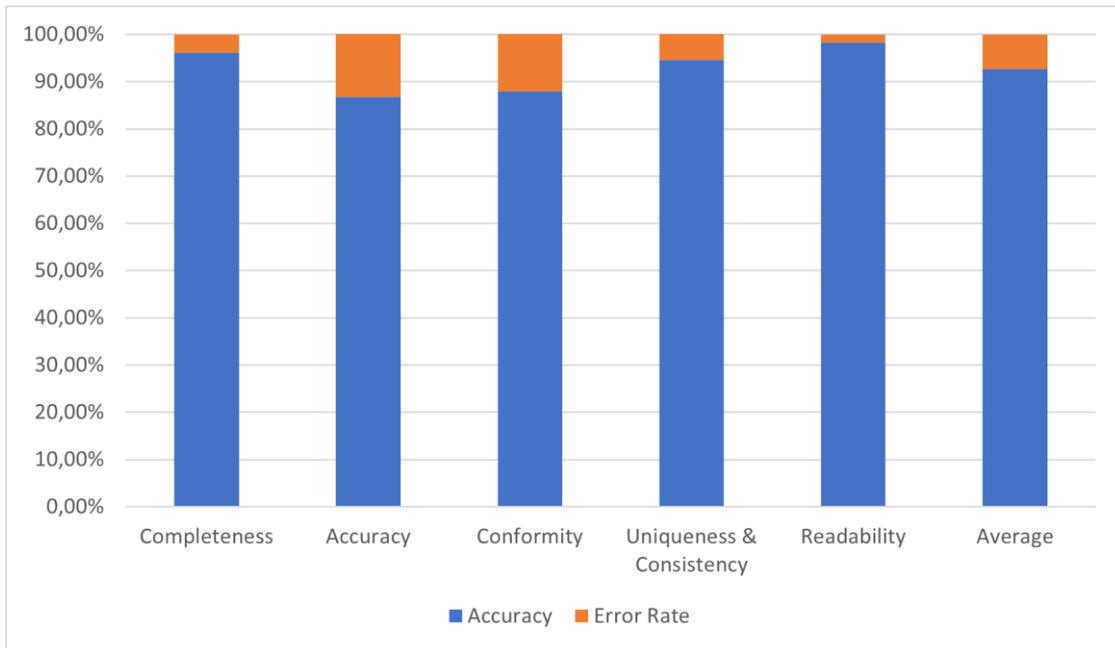

*Figure 23: The Accuracy and Error Rate of the First Dataset*

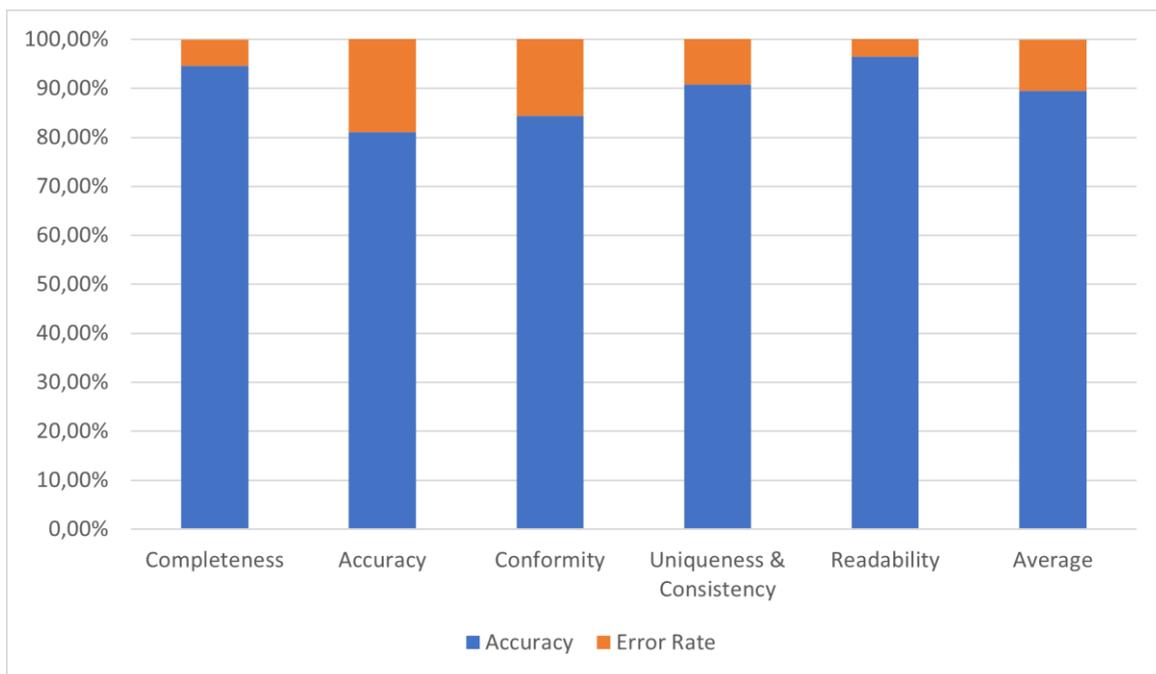

*Figure 24: The Accuracy and Error Rate of the Second Dataset*

Moreover, we calculated the quality score for each dimension before and after implementing the anomaly correction framework. This allowed us to evaluate the extent of improvement achieved for each specific measurement and the overall quality of the dataset. We employed metrics defined in Chapter 4 to assess the dataset's quality. These metrics provided a standardized and comprehensive way to measure the various aspects of data quality and evaluate the effectiveness of our anomaly correction framework in enhancing the dataset's overall quality. Figures 25 and 26 show the measured quality metrics before and after applying the anomaly correction framework, the improvement achieved for each quality dimension, and the global quality score for the first and the second dataset, respectively.

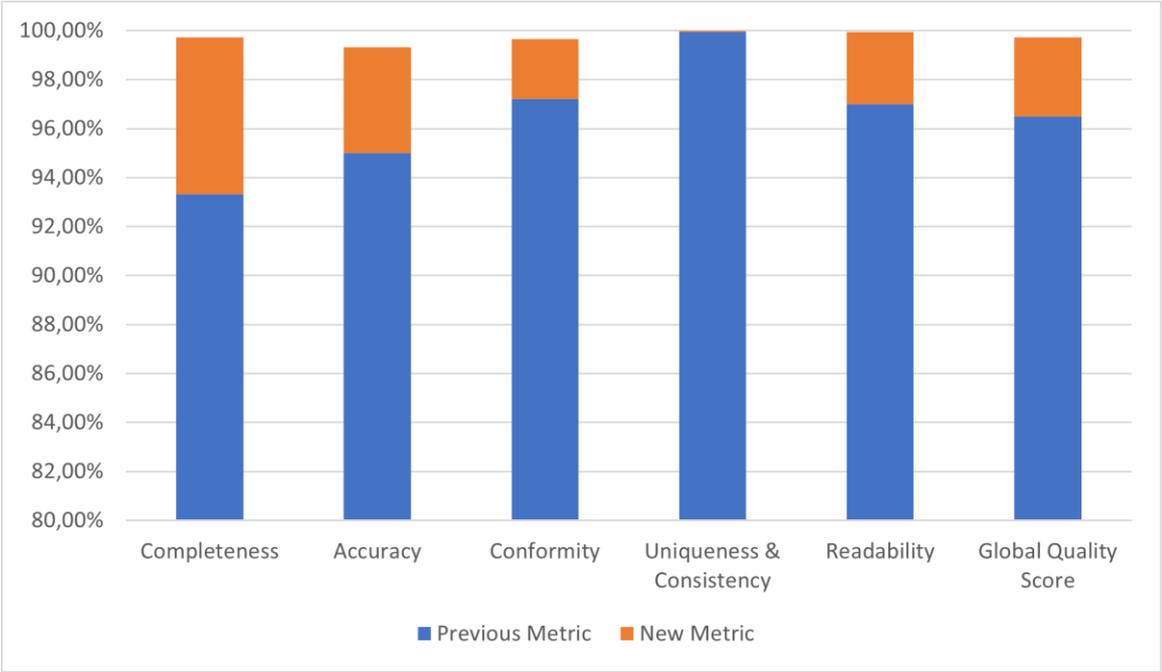

*Figure 25: The Quality Improvement Rate for the First Dataset*

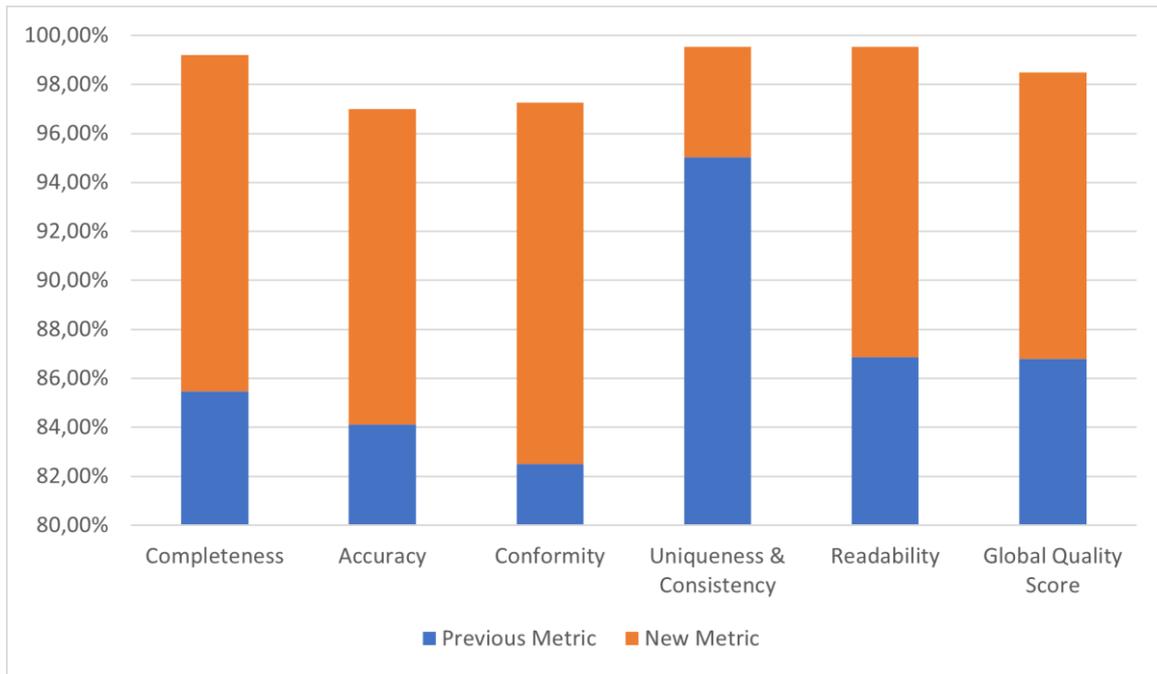

*Figure 26: The Quality Improvement Rate for the Second Dataset*

## 4.4. Discussion

Based on the above-obtained results, the framework has achieved good performance for correction quality anomalies for both datasets, with an average accuracy of 92.71% for the first dataset and 89.45% for the second dataset. The proposed framework has achieved a lower accuracy score for the second dataset than the first one because real-world datasets often have more significant variability, noise, and inconsistencies and contain more complex and diverse quality anomalies. With the minor difference in the achieved accuracy scores for both datasets, the framework has maintained relatively consistent performance, affirming its effectiveness for real-world scenarios. The accuracy score was higher for the uniqueness, consistency, and readability dimensions than the others. Indeed, conformity and accuracy anomalies are the most challenging to correct. They often require understanding and capturing more complex relationships between different data elements, which are not always available in the training datasets. Despite these challenges, the framework has still demonstrated its effectiveness by achieving a good accuracy score of not less than 81%, indicating its ability to capture and correct a significant portion of the anomalies in the dataset. For the second dataset, the framework could complete missing values in most features, such as class and gender. The framework could set an approximation range for continuous features' missing values. For instance, the missing values in the "age" feature were approximated by assigning them to an age group.

Similarly, the framework estimated missing values in the "cabin number" and "fare" features by assigning them to respective ranges. Using word embedding techniques, the framework successfully corrected non-conforming values. For instance, in the "class" feature, the value "1" was corrected to "1st class" based on its semantic similarity to the correct value. Also, in the "gender" feature, the non-conform value "M" was replaced by "Male," ensuring the conformity of the gender values. Overall, the framework corrected a significant portion of conformity anomalies with an accuracy rate of 84.36%. The framework also corrected most misspelled values with an accuracy rate of 96.51% using similarity measures. The model has also achieved a good accuracy score of 90.76% when addressing the Uniqueness and Consistency dimensions. Most of the duplicates were consolidated and merged correctly. To assess the real impact of the proposed framework on the dataset quality, we have measured the addressed quality dimensions before and after applying the anomaly correction framework to the dataset. The results obtained from using the framework demonstrate a significant enhancement in data quality, with an improvement rate of 18.98%. The corrections made by the framework have boosted the dataset's global quality score to 98.5%, with a critical quality improvement rate of 11.71%. For the first dataset, the framework achieved a better result in quality improvement. The framework addressed most of the quality anomalies and boosted all the quality dimensions to a score exceeding 99%. The improvement rate in the first dataset was comparatively lower than in the second dataset. This disparity can be attributed to the initial quality of the first dataset, which was relatively higher. The first dataset's larger size, exceeding 2 million records, contributed to a relatively lower proportion of anomalies than the second dataset. Nonetheless, the framework successfully addressed the majority of quality anomalies, resulting in a substantial enhancement in the dataset's overall quality with an improvement rate of 3.23%.

It is worth noting that our suggested framework has a distinct and unique focus on data quality anomalies. Indeed, it addresses many data quality dimensions and does not focus on a specific dimension, making it inappropriate to compare its performance to existing frameworks with different scopes and objectives. However, the framework outperforms existing approaches in scope as it effectively addresses a wide range of data quality anomalies, encompassing six key data quality dimensions and genericity without any field correlation, making it a unique contribution. Moreover, the framework has achieved promising accuracy and improvement rate results, further validating its effectiveness in correcting data quality anomalies. Furthermore, the framework has demonstrated excellent scalability with reasonable execution times on both datasets while maintaining a linear O(n) complexity. This scalability results from the designed architecture incorporating distributed storage and computing capabilities. Also, using technologies such as Hadoop Distributed File System (HDFS) for distributed storage and processing, along with libraries like H2O for machine learning models, further contributes to the framework's scalability. The framework has shown excellent accuracy and achieved significant data quality improvement rates when applied to real-world and large-scale datasets. However, it is essential to acknowledge the limitations discussed earlier, which should be considered for future work to enhance the framework's performance further. In summary, employing our proposed framework across both datasets has led to the following significant outcomes:

- The enhancement of the quality score of both datasets with an improvement rate of 3.23% for the first dataset and 11.71% for the second dataset.

- The predictive model made an accurate correction with an accuracy score of 92.71% for the first and 89.45% for the second datasets.

- Great framework scalability based on a distributed storage and computing platform appropriate for big datasets.

## Conclusion

In conclusion, increasing reliance on big data for decision-making has heightened the importance of data quality. However, existing studies in the field often have limited scopes and lack comprehensive solutions applicable across different domains. This research paper addresses these gaps by proposing a sophisticated framework that automatically corrects big data quality anomalies using an intelligent predictive model. The framework encompasses six key quality dimensions and is designed to be applicable in various fields, offering a generic approach to address data quality issues. Implementation of the framework on two datasets demonstrated its effectiveness, achieving a high accuracy of up to 98.22%. Additionally, the results showcased significant improvement in data quality, with a boost in quality scores of up to 99% and an impressive improvement rate of up to 14.76%. This chapter contributes to the field by providing a comprehensive and adaptable solution for addressing data quality anomalies in the context of big data.

# Conclusion and Future Research

## 1. Conclusion

In this thesis, we have comprehensively examined the critical issue of big data quality by delving into three principal axes of big data quality management. Initially, we have highlighted the importance of data quality within the realm of big data, elucidating fundamental concepts about the era of big data and its intrinsic relationship with data quality. To gain a more profound knowledge of data quality, we explored the current state-of-the-art by conducting a comprehensive survey of existing approaches that have addressed data quality, encompassing aspects of quality assessment, anomaly detection, and anomaly correction. Identifying notable gaps in the existing body of knowledge, we have made substantial contributions to each of these critical axes as follows:

1. To enhance the first axis, which consists of big data quality assessment, we proposed a novel comprehensive Big Data Quality Assessment Framework based on 12 quality metrics: Completeness, Timeliness, Volatility, Uniqueness, Conformity, Consistency, Ease of manipulation, Relevancy, Readability, Security, Accessibility, and Integrity. Moreover, we introduced the weighted data quality assessment concept, allowing us to enhance the accuracy and precision of the performed assessment.
2. Then, to address the second axis consisting of detecting data quality anomalies, we have defined a novel Data Quality Anomaly Detection Framework based on a machine learning model that allows identifying potential generic data quality anomalies for Big Data related to six quality dimensions: Accuracy, Consistency, Completeness, Conformity, Uniqueness, and Readability. Moreover, as a foundational step for detecting quality anomalies, we have suggested an End-to-End Big Data Entity Resolution Framework that outperforms the existing methods and provides the best results based on a Semi-Supervised learning approach. The proposed framework uses an online learning strategy that allows for addressing the decreasing performance of the model and maintaining a high accuracy during the serving. Furthermore, we introduced and measured a new metric called "Quality Anomaly Score." This metric refers to the degree of anomalousness and the poor quality of the detected anomalies of each quality dimension and the whole dataset.

3. Finally, to address the third axis of enhancing big data quality, we have proposed an advanced framework for Big Data Quality Anomaly Correction that adopts an intelligent and sophisticated methodology based on a predictive model allowing to correct anomalies related to six critical dimensions of data quality: Accuracy, Completeness, Conformity, Uniqueness, Consistency, and Readability. The proposed framework is not restricted to a specific field and offers a generic approach to correcting big data quality anomalies.

All the proposed frameworks were meticulously implemented and rigorously tested, demonstrating remarkable effectiveness in addressing and improving the three principal axes of big data quality management.

## 2. Future research

As part of our ongoing research and future endeavors, we aspire to expand and enrich our data quality assessment framework. Our current framework is based on a set of 12 essential quality metrics. However, the data landscape constantly evolves, and new data quality dimensions may emerge. We intend to identify and include additional metrics that capture emerging data quality concerns to ensure our framework remains relevant and robust. This inclusivity will enable us to adapt to evolving data challenges effectively.

Enhancing the effectiveness of anomaly detection is a critical aspect of our ongoing research in the second axis of our work. Our objectives in this area encompass improving the accuracy and precision of anomaly detection and extending our capabilities to identify and address more challenging anomalies. Indeed, anomalies in real-world data can vary widely in terms of complexity and characteristics. In our research, we are committed to expanding our capabilities to detect more challenging types of anomalies, including anomalies that are difficult to discern from normal behavior, context-dependent anomalies, and anomalies that evolve. We aim to develop specialized techniques for these scenarios.

Regarding the third and last addressed axis, we have identified an essential enhancement for the current anomaly correction framework. We plan to incorporate real-time data quality monitoring into the framework to address the growing need for real-time data processing. This includes exploring techniques and methodologies to monitor data quality in real time and provide timely anomaly correction. By incorporating real-time data quality monitoring, organizations can detect and address data quality issues as they occur, enabling them to make timely and accurate decisions based on reliable data.